%% file: root.tex
\documentclass[letterpaper, 10 pt, conference]{ieeeconf}  

\IEEEoverridecommandlockouts                              

\makeatletter
\let\NAT@parse\undefined
\makeatother
\input{macros/packages.tex}
\input{macros/notations.tex}

\title{\LARGE \bf
LTLDoG: Satisfying Temporally-Extended Symbolic Constraints for Safe Diffusion-based Planning
}

\author{Zeyu~Feng$^{\dag 1}$, Hao~Luan$^{\dag 1}$, Pranav~Goyal$^{1}$, and Harold~Soh$^{1,2}$
\thanks{\textsuperscript{\textdagger}Equal contribution.}
\thanks{$^{1}$All authors are with the Department of Computer Science, School of Computing, National University of Singapore, Singapore. {\tt\small \{zeyu, haoluan, pgoyal, harold\}@comp.nus.edu.sg}}%
\thanks{$^{2}$H.~S. is also with the Smart Systems Institute, National University of Singapore.}%
}

\begin{document}

\maketitle
\thispagestyle{empty}
\pagestyle{empty}

\begin{abstract}
  Operating effectively in complex environments while complying with specified constraints is crucial for the safe and successful deployment of robots that interact with and operate around people. In this work, we focus on generating long-horizon trajectories that adhere to static and temporally-extended constraints/instructions at test time. 
  We propose a data-driven diffusion-based framework, \method, that modifies the inference steps of the reverse process given an instruction specified using finite linear temporal logic ($\ltlf$). \method{} leverages a satisfaction value function on $\ltlf$ and guides the sampling steps using its gradient field. This value function can also be trained to generalize to new instructions not observed during training, enabling flexible test-time adaptability. Experiments in robot navigation and manipulation illustrate that the method is able to generate trajectories that satisfy formulae that specify obstacle avoidance and visitation sequences. Code and supplementary material are available online at \url{https://github.com/clear-nus/ltldog}. 
\end{abstract}

\input{sections/intro.tex}
\input{sections/preliminaries.tex}
\input{sections/method.tex}
\input{sections/relatedwork.tex}

\input{sections/experiments.tex}
\input{sections/conclusion.tex}

\input{sections/acks.tex}

\bibliographystyle{IEEEtran}
\bibliography{IEEEabrv,references}

\appendix
\input{sections/appendix}

\end{document}

%% file: macros/packages.tex
\usepackage{algorithm}
\usepackage{algpseudocode}

\usepackage{amsmath}
\usepackage{amstext}
\usepackage{amssymb}
\usepackage{amsfonts}
\usepackage{mathtools}
\usepackage{latexsym}
\usepackage{bm}
\usepackage{bbm}
\usepackage{dsfont}
\usepackage{nicefrac}
\usepackage[super]{nth}
\usepackage{wasysym}
\usepackage{pifont}

\usepackage{array}
\usepackage{multirow}
\usepackage{booktabs}
\usepackage{tikz}

\usepackage{xspace}
\usepackage{subfigure}  
\usepackage{balance}
\usepackage{threeparttable}
\usepackage[noadjust]{cite}
\usepackage[bookmarks=true]{hyperref}
\usepackage{xcolor}

%% file: macros/notations.tex
\usepackage{mathtools}
\usepackage{amssymb}
\usepackage{amsmath}
\usepackage{tikz}
\usepackage{xspace}

\newcommand{\btau}[1]{\bm{\tau}^{#1}}

\DeclareMathOperator*{\argmin}{arg\,min}
\newcommand*\diff{\mathop{}\!\mathrm{d}}

\newcommand{\expect}[2]{\mathbb{E}_{#1} \left[#2 \right]}
\newcommand{\prob}[2]{p_{#1} \left(#2\right)}
\newcommand{\cprob}[3]{p_{#1} \left(#2 |#3\right)}
\newcommand{\defeq}{\mathrel{\mathop:}=}

\newcommand{\tuple}[1]{\langle{#1}\rangle}

\newcommand{\ltlf}{\textnormal{LTL}_f}

\newcommand{\ltlalways}{\ensuremath{\Box}\xspace} 
\newcommand{\ltleventually}{\ensuremath{\Diamond}\xspace}
\newcommand{\ltlnext}{\ensuremath{\bigcirc}\xspace}

\newcommand{\true}{\ensuremath{\mathsf{true}}\xspace{}}
\newcommand{\false}{\ensuremath{\mathsf{false}}\xspace{}}
\newcommand{\ltluntil}{\ensuremath{\operatorname{\mathsf{U}}}\xspace{}}

\newcommand{\ltlimplies}{\ensuremath{\operatorname{\Rightarrow}}\xspace{}}

\newlength\squareheight
\definecolor{gold}{rgb}{1.0, 0.84, 0.0}

\newcommand{\ie}{\textit{i.e.}}
\newcommand{\eg}{\textit{e.g.}}
\newcommand{\etal}{\textit{et~al.}}

\newcommand{\figref}[1]{Fig.~\ref{#1}}
\newcommand{\tabref}[1]{TABLE~\ref{#1}}

\newcommand{\parabf}[1]{\vspace{0.1cm}\noindent\textbf{#1}}

\newcommand{\diffuser}{\textsc{Diffuser}\xspace}
\newcommand{\diffusionpolicy}{\textsc{Diffusion Policy}\xspace}
\newcommand{\safediffuser}{\textsc{SafeDiffuser}\xspace}
\newcommand{\method}{\textsc{LTLDoG}\xspace}
\newcommand{\methodrg}{\textsc{LTLDoG-R}\xspace}
\newcommand{\methodps}{\textsc{LTLDoG-S}\xspace}

%% file: sections/intro.tex
\section{Introduction}
\label{sec:intro}

Recent methodologies~\cite{janner2022diffuser,ajay2023is,chi2023diffusionpolicy} utilizing data-driven diffusion models~\cite{pmlr-v37-sohl-dickstein15,ho2020denoising,song2021scorebased} have shown remarkable performance in generating robot behaviors across a wide range of tasks. Thanks to their ability to model complex distributions, these methods have  surpassed several leading offline reinforcement learning techniques and classical model-based trajectory optimization methods, especially in long-horizon decision-making tasks~\cite{janner2022diffuser,chi2023diffusionpolicy}. However, while conventional diffusion models excel at learning from training datasets, they lack the ability to adapt to new objectives or comply with new constraints during deployment. This shortcoming can lead to unsafe behaviors, posing risks to humans, robots, and their surrounding environment.

In view of this limitation, there has been very recent work on diffusing \emph{safe} trajectories. Xiao~\etal~\cite{xiao2023safediffuser} integrated a dynamics model into the denoising diffusion process and incorporated a class of Control Barrier Functions (CBF)
to meet safety criteria. Botteghi~\etal~\cite{botteghi2023trajectory} approached the issue by embedding both safety and reward considerations into a constrained optimization framework, employing CBF constraints as labels for classifier guidance. However, these approaches primarily address \emph{static} environmental constraints. For example, while they can maneuver around obstacles on a \emph{local} scale, they fail to comply with more complex temporally-extended directives such as ``\emph{avoid the kitchen until you are clean}''.

\begin{figure}[t]
    \centering
    \includegraphics[width=0.9\columnwidth]{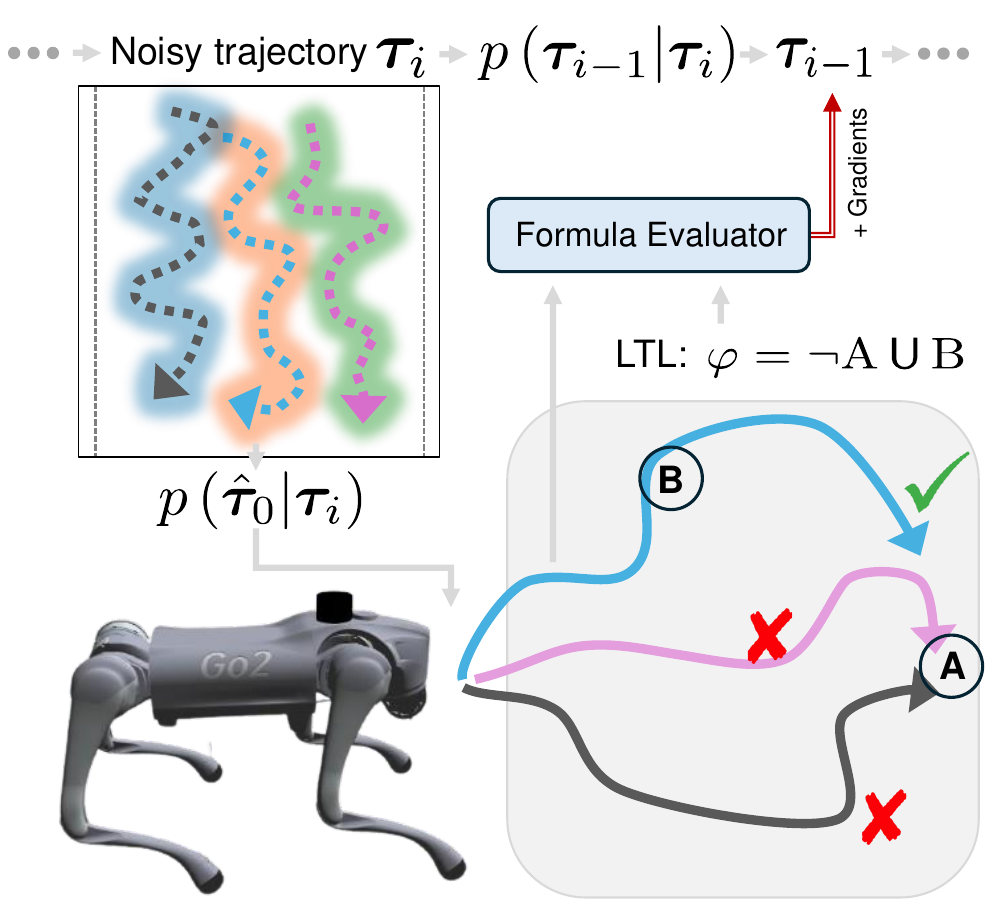}
    \caption{We present \method{}, a diffusion-based planning framework for generating trajectories that comply with specified $\ltlf$ formulae. In the example above, a robot dog is tasked to arrive at the goal position (A), but first has to visit B and avoid obstacles (crosses).}
    \label{fig:illustration}
    \vspace*{-10pt}
\end{figure}

In this work, we propose an alternative approach to flexible trajectory planning with diffusion models, designed to satisfy both static safety requirements and temporal constraints. The core idea is to plan with diffusion models to satisfy finite linear temporal logic ($\ltlf$) formulae~\cite{4567924}. $\ltlf$ offers the ability to define a broad spectrum of instructions/constraints that might emerge during deployment. For example, $\ltlf$ can describe a visitation order of different objects and locations.
The use of propositional logic operators, such as \textit{not}, facilitates the delineation of safe regions within the state space.

We develop \method{} (\textbf{LTL} \textbf{D}iffusion-\textbf{o}rienting \textbf{G}uidance, pronounced ``Little Dog''), a posterior-sampling based diffusion framework that accommodates  finite LTL formulae at test time. We present two variants of \method{}: our main method, \methodps, can be applied to the generation of finite-length trajectory in robot tasks where the labeling function for propositional events is differentiable. \methodps employs a differentiable formula checker in conditional sampling --- specifically, we modify the reverse process to condition upon the criteria that the final (predicted) trajectory satisfies a given $\ltlf$ formula. 
For when a differentiable labeling function is unavailable, we propose \methodrg, which uses a trained $\ltlf$ neural-symbolic evaluator for posterior sampling. 
Notably, both variants do not require collecting expert demonstrations for every potential $\ltlf$ instruction. They retain the temporal compositionality and local consistency properties associated with diffusion models~\cite{janner2022diffuser} --- as long as the dataset contains a diverse set of paths, they can potentially ``stitch together'' snippets of trajectories from the training data to generate plans for unseen $\ltlf$ formulae.

Experiments on two benchmark environments (long-horizon planning for navigation and policy learning for manipulation) demonstrate that \method is able to generate trajectories that satisfy feasible safety and temporal constraints. We find that our methods possess the ability to re-plan alternative paths at a high-level based on given instruction. Moreover, real robot experiments show that the generated trajectories can be successfully transferred to a quadruped robot. In summary, this paper makes three key contributions:
\begin{itemize}
    \item A conditional trajectory sampling approach designed for $\ltlf$ instructions that leverages pre-trained diffusion models;
    \item A regressor-guidance neural network for diffusion that generalizes to novel $\ltlf$ formulae within a given template structure; 
    \item Experimental results on benchmark problems and real world demonstrations that validate the effectiveness of planning with safety and temporal constraints.
\end{itemize}
From a broader perspective, \method{} is the first method that fuses symbolic model checking (using $\ltlf$) with expressive diffusion-based generative models. We hope our results lays the groundwork towards performant, yet safer and more trustworthy robots. 

%% file: sections/preliminaries.tex
\section{Preliminaries and Notation}
In this work, our focus is to extend diffusion-based planning methods towards generating trajectories that comply with specified $\ltlf$ formulae. 
Here, we provide a concise introduction to diffusion methods in the context of planning and finite linear temporal logic.

\subsection{Planning with Diffusion}

Many tasks in planning, reinforcement learning, and imitation learning require generating trajectories under some specific objective. Let $\mathcal{S}$ and $\mathcal{A}$ denote the state and action space, respectively. We use $\btau{} = \left(\mathbf{s}_0, \mathbf{a}_0, \mathbf{s}_1, \mathbf{a}_1, \ldots, \mathbf{s}_T, \mathbf{a}_T\right)$ to refer to a trajectory, where $T$ is the planning horizon. The environment transitions to a new state $\mathbf{s}_{t+1}$ when an agent executes action $\mathbf{a}_t$ at state $\mathbf{s}_t$. Let the abbreviation $\mathcal{J}(\btau{}|g)$ denote the objective value function conditioned on a goal state where the trajectory must terminate at, for example, (discounted) cumulative rewards in reinforcement learning, cumulative error of actions in imitation learning, or cost for safety constraints.

Diffusion-based planning methods directly generate partial or entire trajectories by using diffusion models pre-trained on a dataset of trajectories. Let $\prob{0}{\btau{0}}$ denote the distribution of trajectories in dataset, where $\btau{0}$ represents a noiseless trajectory. Given an $N$-step discrete approximation of forward diffusion process $\cprob{}{\btau{i}}{\btau{i-1}}$ that slowly corrupts data by adding prespecified noise, diffusion models learn an iterative denoising procedure by approximating the score function $\nabla_{\btau{i}} \log \prob{i}{\btau{i}}$ using a step-dependent neural network $s_\theta$ trained with denoising score matching~\cite{vincent2011connection}:
\begin{equation}
\label{eq:dsm}
    \theta^* = \argmin_\theta \expect{i,\btau{i},\btau{0}}{\left\lVert s_\theta\left(\btau{i},i\right) - \nabla_{\btau{i}}\log\cprob{}{\btau{i}}{\btau{0}}\right\rVert^2},
\end{equation}
in which $i\sim\mathcal{U}\{1,2,\ldots,N\}$ is the diffusion timestep, and $\btau{i}\sim \cprob{}{\btau{i}}{\btau{0}}$ is the trajectory $\btau{0}$ corrupted with noise. Throughout the paper, we adopt Denoising Diffusion Probabilistic Models (DDPM)~\cite{ho2020denoising} as the sampling method, where $\cprob{}{\btau{i}}{\btau{0}} = \mathcal{N}\left(\sqrt{\bar\alpha_i}\btau{0},\left(1-\bar\alpha_i\right)\bm{I}\right)$, $\bar\alpha_i \defeq \prod_{j=1}^i \alpha_i$, $\alpha_i \defeq 1-\beta_i$ and $\{\beta_i\}$ is a sequence of positive noise scales $0<\beta_1,\beta_2,\dots,\beta_N<1$.

\subsection{Linear Temporal Logic (LTL)}
\label{sec:ltl}

Given a finite set of propositional symbols $\mathcal{P}$, the formula set $\Psi$ of $\ltlf$ contains formulas recursively defined in Backus-Naur form as follows~\cite{baier0020348,belta2017formal}:
\begin{equation*}
    \varphi \defeq p \;|\; \neg \varphi \;|\; \varphi \wedge \psi \;|\; \ltlnext \varphi \;|\; \varphi \ltluntil \psi,  
\end{equation*}
where $p \in \mathcal{P}$ and $\varphi,\psi\in\Psi$.
Intuitively, the formula $\ltlnext \varphi$ (\textit{next} $\varphi$) is satisfied if $\varphi$ is satisfied at the next time step.
$\varphi \ltluntil \psi$ ($\varphi$ \textit{until} $\psi$) is satisfied if $\varphi$ is satisfied until $\psi$ is satisfied, and $\psi$ is satisfied by the end of the sequence.
From these, other commonly used logical connectives and temporal operators can be defined according to the following equivalences:
$\varphi \vee \psi = \neg \left(\neg\varphi \wedge \neg\psi\right)$,
$\ltleventually \varphi = \true \ltluntil \varphi$ (\textit{eventually} $\varphi$) and
$\ltlalways \varphi = \neg\ltleventually \left(\neg\varphi\right)$ (\textit{always} $\varphi$).
The symbols $\true$ and $\false$ can also be in the formula set defined by
$\true = \varphi \vee \neg\varphi$ and
$\false = \neg\true$.

In contrast to propositional logic, these formulas are evaluated over finite sequences of observations $\bm{\sigma} =\langle \sigma_0, \sigma_1, \sigma_2, \ldots \sigma_T \rangle$ (\ie,~\emph{truth assignments} to the propositional symbols in $\mathcal{P}$), where $\sigma_t\in \{0,1\}^{\vert\mathcal{P}\rvert}$ and $\sigma_{t,p}=1$ iff proposition $p \in \mathcal{P}$ is satisfied at time step $t$. $\true$ ($\false$) is always satisfied (not satisfied) by any assignment. Formally, $\bm{\sigma}$ \emph{satisfies} $\varphi$ at time $t \geq 0$, denoted by $\tuple{\bm{\sigma},t}\models\varphi$, as follows:
\begin{itemize}
    \item $\tuple{\bm{\sigma},t}\models p$ iff $\sigma_{t,p}=1$, where $p \in \mathcal{P}$
    \item $\tuple{\bm{\sigma},t}\models \neg \varphi$ iff $\tuple{\bm{\sigma},t}\not\models\varphi$
    \item $\tuple{\bm{\sigma},t}\models (\varphi\wedge\psi)$ iff $\tuple{\bm{\sigma},t}\models\varphi$ and $\tuple{\bm{\sigma},t}\models\psi$
    \item $\tuple{\bm{\sigma},t}\models\ltlnext\varphi$ iff $\tuple{\bm{\sigma},t+1}\models\varphi$
    \item $\tuple{\bm{\sigma},t}\models\varphi\ltluntil\psi$ iff $\exists t_2 \in [t,T]$ s.t. $\tuple{\bm{\sigma},t_2}\models\psi$ and $\forall t_1 \in [t,t_2)$, $\tuple{\bm{\sigma},t_1}\models\varphi$
\end{itemize}
A sequence $\bm{\sigma}$ is then said to \emph{satisfy} $\varphi$,~\ie, $\bm{\sigma}\models\varphi$, iff $\tuple{\bm{\sigma},0}\models\varphi$.

%% file: sections/method.tex
\section{Method}
In this section, we describe our primary contribution, \method, a diffusion-based framework for generating trajectories that satisfy $\ltlf$ formulae. We first discuss how to conditionally sample using diffusion models, followed by how $\ltlf$ formulae can be used to guide the diffusion process.

\subsection{Conditional Sampling in Diffusion Models}

Given a trained score function from \eqref{eq:dsm} such that $s_{\theta} \approx \nabla_{\btau{i}}\log\prob{i}{\btau{i}}$, a diffusion model denoises 
samples 
according to the distribution 
$\cprob{\theta}{\btau{i-1}}{\btau{i}} = \mathcal{N}\left(\frac{1}{\sqrt{\alpha_i}}\left( \btau{i} + \left(1-\alpha_i\right) s_{\theta}\left(\btau{i},i\right) \right),\left(1-\alpha_i\right)\bm{I}\right)$ 
starting from a Gaussian prior $\btau{N}\sim\mathcal{N}\left(\bm{0},\bm{I}\right)$. For example, \diffuser~\cite{janner2022diffuser} samples a trajectory $\btau{0}$ from a diffusion model, which an agent then executes. However, this original sampling process is unable to control detailed properties of generated context. 

Here, we are interested in sampling trajectories that satisfy both the final goal and the specified instructions encoded as an $\ltlf$ formula $\varphi$ that is provided during deployment. In other words, we aim to sample trajectories under an objective function $\mathcal{J}_{\varphi}\left(\btau{0}|g\right)$. For example, $\mathcal{J}$ can have a high value if the events induced by $\btau{0}$ satisfy $\varphi$ and have a low value otherwise.

Formally, given a set of atomic propositions $\mathcal{P}$, the assignments for $\btau{}$ are given by a labeling function $L: \mathcal{S}\times \mathcal{A}\rightarrow 2^{\lvert\mathcal{P}\rvert}$, where each timestep of $\btau{}$ induces an assignment to the propositional symbols in $\mathcal{P}$. For example, in navigation task, $\mathcal{P}$ can represent multiple regions to avoid and $L$ is a function indicating whether $(\mathbf{s}_t, \mathbf{a}_t)$ in $\btau{}$ are inside these regions or not. We will slightly abuse notation and write $\btau{}\models\varphi$ to indicate that $\btau{}$'s assignments $\bm{\sigma}$ satisfy a $\ltlf$ formula. 

We aim to sample from the posterior $\cprob{0}{\btau{0}}{\btau{0}\models\varphi,g}$ with the diffusion model as the prior. In this work, the constraint of goal state conditioning $g$ can be either achieved by inpainting (similarly on the start state $\mathbf{s}_0$) or implicitly encoded in the dataset, which does not require separate modeling in conditional sampling. Therefore, we mainly target the posterior $\cprob{0}{\btau{0}}{\btau{0}\models\varphi}$. Let $y_0$ denote a binary random variable indicating the likelihood of $\btau{0}$ satisfying $\varphi$. Hence, the denoising process requires a score function conditioned on $y_0=1$ and by Bayes’ rule: $\nabla_{\btau{i}} \log \cprob{i}{\btau{i}}{y_0=1} = \nabla_{\btau{i}} \log \prob{i}{\btau{i}} + \nabla_{\btau{i}} \log \cprob{i}{y_0=1}{\btau{i}}$. The first term on the right-hand side has been learned by a neural network in diffusion. However, the latter term requires an integration over all possible values of $\btau{0}$: $\cprob{i}{y_0}{\btau{i}} = \int\cprob{}{y_0}{\btau{0}}\cprob{i}{\btau{0}}{\btau{i}}\diff\btau{0}$.
We consider the plug-and-play conditional generation setting and approximating this integration with sample estimation~\cite{chung2023diffusion},~\eg, point estimation with $\cprob{i}{y_0}{\btau{i}} \approx \cprob{}{y_0}{\hat{\bm{\tau}}^0}$ where the noiseless trajectory $\hat{\bm{\tau}}^0$ is estimated via Tweedie’s formula~\cite{efron2011tweedie} $\hat{\bm{\tau}}^0 = \frac{1}{\sqrt{\bar\alpha_i}} \left(\btau{i} + \left(1-\bar\alpha_i\right) \nabla_{\btau{i}}\log\prob{i}{\btau{i}}\right)$. We model the likelihood term as $\cprob{}{y}{\btau{}} = Ze^{\mathds{1}\left[y=\mathds{1}\left[\btau{}\models\varphi\right]\right]}$, where $Z$ is a normalizing constant and $\mathds{1}\left[\cdot\right]$ is the indicator function. Putting the above elements together, the conditional score function can be computed as follows,
\begin{equation}
\begin{split}
    \nabla_{\btau{i}} \log \cprob{i}{y_0}{\btau{i}} & \approx \nabla_{\btau{i}} \log \cprob{}{y_0}{\hat{\bm{\tau}}^0} \\
    & = \nabla_{\btau{i}} \log \left(Ze^{\mathds{1}\left[y=\mathds{1}\left[\hat{\bm{\tau}}^0\models\varphi\right]\right]}\right) \\
    & = \nabla_{\btau{i}} \mathds{1}\left[y=\mathds{1}\left[\hat{\bm{\tau}}^0\models\varphi\right]\right].
\end{split}
\label{eq:score_ps_1}
\end{equation}
Unfortunately, both the indicator function and the satisfaction evaluation (performed by techniques like model checking with finite automata) are non-differentiable --- this prohibits application in the gradient-based sampling process of diffusion models. We address this problem in the next subsection. 

\subsection{Differentiable Evaluation of $\ltlf$}
\label{sec:dtl_ps}

Our key approach is to ``soften'' the satisfaction evaluation. 
To make the evaluation differentiable, we modify our formula evaluator to output positive real values if a trajectory satisfies the $\ltlf$ formula, and negative real values otherwise. Instead of using a binary labeling functions, we assume real valued assignments of atomic propositions can be obtained through computation on the generated trajectories. As a specific example, consider a navigation task where a robot has to avoid obstacles; one can determine the assignment for proposition $p$ at timestep $t$ using the Euclidean distance between $\mathbf{s}_t$ and the centers of the region $\mathbf{c}_p$ (assuming a circular shape). A positive value of $\ell(p,t) = r_p - \lVert\mathbf{s}_t-\mathbf{c}_p\rVert_2$, where $r_p$ is the radius of the circle, indicates a true assignment. Consequently, the labeling function for the entire trajectory $L: (\mathcal{S}\times\mathcal{A})^{T+1}\rightarrow \mathbb{R}^{\lvert\mathcal{P}\rvert\times(T+1)}$ is differentiable and the assignments are $\bm{\sigma}=L(\btau{})$.

A binary version of $\bm{\sigma}$ through a sign function $\text{sgn}(\bm{\sigma})$ satisfies the definition of~\emph{satisfy} defined in Section~\ref{sec:ltl}, but the sign operation breaks differentiability. As such, we employ a formula evaluator $f_t\left(\varphi,\bm{\sigma}_{t:T}\right):\Psi\times\mathbb{R}^{\lvert\mathcal{P}\rvert\times (T-t+1)} \rightarrow \mathbb{R}$ to check satisfaction~\cite{xu2022dont}, with positive values implying $\tuple{\bm{\sigma},t}\models\varphi$, similar to signal temporal logic~\cite{2023stl}. As such, $f$ is differentiable with the evaluation process defined as follows:
\begin{itemize}
    \item $f_t(\true,\bm{\sigma}) = +\infty$
    \item $f_t(\false,\bm{\sigma}) = -\infty$
    \item $f_t(p,\bm{\sigma}) = \bm{\sigma}_{t,p}$
    \item $f_t(\neg\varphi,\bm{\sigma}) = -f_t(\varphi,\bm{\sigma})$
    \item $f_t(\varphi\wedge\psi,\bm{\sigma}) = \min^{\gamma}\left\{f_t(\varphi,\bm{\sigma}),f_t(\psi,\bm{\sigma})\right\}$
    \item $f_t(\varphi\vee\psi,\bm{\sigma}) = \max^{\gamma}\{f_t(\varphi,\bm{\sigma}),f_t(\psi,\bm{\sigma})\}$
    \item $f_t(\ltlnext\varphi,\bm{\sigma}) = f_{t+1}(\varphi,\bm{\sigma})$
    \item $f_t(\ltlalways\varphi,\bm{\sigma}) = \min^{\gamma}\{f_{t:T}(\varphi,\bm{\sigma})\}$
    \item $f_t(\ltleventually\varphi,\bm{\sigma}) = \max^{\gamma}\{f_{t:T}(\varphi,\bm{\sigma})\}$
    \item $f_t(\varphi\ltluntil\psi,\bm{\sigma}) = \min^{\gamma}\{f_{t:k}(\varphi,\bm{\sigma}),f_t\left(\ltleventually\psi,\bm{\sigma}\right)\}$, where $k\geq t$ is the smallest integer s.t. $f_k(\psi,\bm{\sigma})>0$
\end{itemize}
Note that the $\min$ and $\max$ functions are likewise ``soft'' to maintain differentiability. To reduce clutter, we have omitted the subscript $t$ when $t=0$. With these operations, $f$ maintains  \emph{quantitative semantics}, which preserves the relative values between different $\bm{\sigma}$ such that trajectories with larger margin to a satisfying assignment have larger values.

Given $L$ and $f$, we can obtain a differentiable score function by replacing the likelihood term in \eqref{eq:score_ps_1} with $\cprob{}{y}{\btau{}} = Ze^{(2y-1)f\left(\varphi,L\left(\btau{}\right)\right)}$. The conditional score is then
\begin{equation}
\begin{split}
    \nabla_{\btau{i}} \log \cprob{i}{y_0=1}{\btau{i}} & \approx \nabla_{\btau{i}} \log \left(Ze^{f\left(\varphi,L\left(\hat{\bm{\tau}}^0\right)\right)}\right) \\
    & = \nabla_{\btau{i}} f\left(\varphi,L\left(\hat{\bm{\tau}}^0\right)\right).
\end{split}
\label{eq:score_ps_dtl}
\end{equation}
This approximation of the conditional score function can be directly used with a pre-trained diffusion model to sample trajectories conditioned on an $\ltlf$ formula. We call this method \methodps, since it performs the above posterior sampling in the reverse process (Algorithm~\ref{alg:ltl_diffuser_ps}). The gradient ascent step is controlled by a stepsize $\{\zeta_i\}_{i=1}^N$; in practice, the stepsize for each denoise step can be adaptive such that $\btau{i-1}$ remains valid according to the formula evaluator $f\left(\varphi,L\left(\hat{\bm{\tau}}^0\right)\right)$.

\begin{algorithm}[H]
    \caption{$\ltlf$ Planning with Posterior Sampling}
    \label{alg:ltl_diffuser_ps}
    \begin{algorithmic}[1]
        \Require $\varphi$, $N$, $s_{\theta}$, $\{\zeta_i\}_{i=1}^N$
        \State $\btau{N} \sim \mathcal{N}\left(\bm{0}, \bm{I}\right)$
        \For{$i = N-1$ {\bfseries to} $0$}
            \State{{$\hat s \gets s_{\theta}\left(\btau{i},i\right)$}}
            \State{$\hat{\bm{\tau}}^0 \gets \frac{1}{\sqrt{\bar\alpha_i}} \left(\btau{i} + \left(1-\bar\alpha_i\right) \hat s\right)$}
            \State{$\bm{\epsilon} \sim \mathcal{N}\left(\bm{0}, \bm{I}\right)$}
            \State{$\btau{i-1} \gets \frac{\sqrt{\alpha_i}\left(1-\bar\alpha_{i-1}\right)}{1 - \bar\alpha_i}\btau{i} + \frac{\sqrt{\bar\alpha_{i-1}}(1-\alpha_i)}{1 - \bar\alpha_i}\hat{\bm{\tau}}^0 + \sqrt{1-\alpha_i}\bm{\epsilon}$}
            \State{$\btau{i-1} \gets \btau{i-1} + {\zeta_i} \nabla_{\btau{i}} f\left(\varphi,L\left(\hat{\bm{\tau}}^0\right)\right)$}
        \EndFor
        \State {\bfseries return} $\hat{\bm{\tau}}^0$
    \end{algorithmic}
\end{algorithm}

\subsection{Classifier Guidance over $\ltlf$}
\label{sec:dtl_rg}

One limitation of \methodps is that it necessitates a known (and differentiable) expression for the formula evaluator $f\left(\varphi, L\left(\hat{\bm{\tau}}^0\right)\right)$. This requirement can be challenging to meet in scenarios where the truth assignments of propositions are uncertain, for example, when the ground-truth physical dynamics are unknown. Here, we circumvent this problem by employing classifier guidance using a trained formula evaluator.

We propose a variant of \method with \emph{regressor guidance}, which we abbreviate as \methodrg. Using an $\ell_2$ loss, we train a neural network  to predict the satisfaction values from noisy trajectories conditioned on $\ltlf$ instructions. In other words, our neural network approximates the conditional score function $\nabla_{\btau{i}} \log \cprob{i}{y_0=1}{\btau{i}}$. Note that in contrast to learning a binary classifier for trajectory satisfaction, we apply the labeling function in Section~\ref{sec:dtl_ps} on the dataset and associate with each trajectory its objective value $\mathcal{J}_{\varphi}\left(\btau{0}\right)$. 
We conduct ablation study in Section~\ref{sec:ablation} to show that using real values performs better than using binary labels. 

To generalize over different $\ltlf$, the neural network takes both noisy trajectory and $\ltlf$ formula embedding as input. 
Multiple methods exist for embedding $\ltlf$ formulae. For instance, one can use Graph Neural Networks (GNNs)~\cite{1555942,4700287} to embed the tree representation of an $\ltlf$ formula directly. Alternatively, the deterministic finite-state automaton (DFA)~\cite{ijcai2017p189,Camacho_Baier_Muise_McIlraith_2018} associated with the formulae can be embedded~\cite{Xie2021}.
In this work we employ $\ltlf$ tasks from~\cite{pmlr-v139-vaezipoor21a} and embed the directed graph of an $\ltlf$ formula using the Relational Graph Convolutional Network (R-GCN)~\cite{10.1007/978-3-319-93417-4_38}, which can generalize to $\ltlf$ formulae with same template structure. The model that approximates the score function $s_{\phi}(\phi,\btau{i},i) \approx \nabla_{\btau{i}} \log \cprob{i}{y_0}{\btau{i}}$ after training can be plugged into the conditional reverse process using regressor guidance as summarized in Algorithm~\ref{alg:ltl_diffuser_rg}.

\begin{algorithm}
    \caption{$\ltlf$ Planning with Regressor Guidance}
    \label{alg:ltl_diffuser_rg}
    \begin{algorithmic}[1]
        \Require $\varphi$, $N$, $s_{\theta}$, $s_{\phi}$, $\{\zeta_i\}_{i=1}^N$
        \State $\btau{N} \sim \mathcal{N}\left(\bm{0}, \bm{I}\right)$
        \For{$i = N-1$ {\bfseries to} $0$}
            \State{$\bm{\epsilon} \sim \mathcal{N}\left(\bm{0}, \bm{I}\right)$}
            \State{$\btau{i-1} \gets \frac{1}{\sqrt{\alpha_i}}\left(\btau{i} + (1-\alpha_i)s_{\theta}\left(\btau{i},i\right)\right) + \sqrt{1-\alpha_i}\bm{\epsilon}$}
            \State{$\btau{i-1} \gets \btau{i-1} + \zeta_i s_{\phi}(\phi,\btau{i},i)$}
        \EndFor
        \State {\bfseries return} $\hat{\bm{\tau}}^0$
    \end{algorithmic}
\end{algorithm}

%% file: sections/relatedwork.tex
\section{Related Work}

\method{} builds upon prior work in diffusion-based planning and symbolic reasoning using $\ltlf$ for robotics. In the following, we give a brief overview of related work. 

\parabf{Learning and Planning under $\ltlf$.}
As an expressive language for specifying high-level planning requirements~\cite{BACCHUS2000123, 10.5555/3037104.3037150, Camacho_Triantafillou_Muise_Baier_McIlraith_2017}, $\ltlf$ has been extensively used in various robotic tasks to express temporally extended goals~\cite{1570410, 10189357}.
These methods usually require the information about the environment's dynamics, \eg, a model or an abstraction, to effectively plan under a given formula. Reinforcement learning agent learn in a model-free way under $\ltlf$ objectives or constraints~\cite{ijcai2022p507, 10.5555/3237383.3237452, voloshin2022policy} with the ability to generalize over different formulae~\cite{pmlr-v139-vaezipoor21a}.
However, these methods operate agent in an online manner via trial and error, which can lead to expensive or even unsafe interactions.

\parabf{Planning and Policy Learning with Diffusion.}
Recent diffusion-based planning methods are flexible that only rely on offline datasets without  access or interaction to environments. They have been successfully applied to long-horizon planning problems by generating states or actions for control~\cite{janner2022diffuser, chi2023diffusionpolicy, ajay2023is}, but not tasks with test-time temporal requirements. 
Recent work has looked into safety critical tasks~\eg, the aforementioned CBF-based methods~\cite{xiao2023safediffuser,botteghi2023trajectory}. As discussed above, these methods were designed for static safety criteria, and the lack the ability to satisfy temporally extended behaviors. Our work inherits the advantages of diffusion based methods and can fulfill $\ltlf$ requirements.

\parabf{Inverse Problems in Diffusion.} Our proposed method formulates conditional measurement under $\ltlf$ with differentiable loss function using the unnormalized likelihood~\cite{chung2023diffusion, pmlr-v202-song23k} for posterior sampling and can guide the sampling process in a plug-and-play fashion. 
The most popular methods to guide diffusion models during inference is classifier guidance~\cite{dhariwal2021diffusion} and classifier-free guidance~\cite{ho2021classifierfree}. However, these methods cannot be applied in a plug-and-play fashion for new conditioning factors. Our work is also related to inverse task that infers a posterior distribution over data given a measurement.
Inverse methods~\cite{jalal2021robust, chung2022score, chung2022come, chung2022improving} do not require training a conditional model in diffusion and can directly leverage pre-trained neural networks.

%% file: sections/experiments.tex
\begin{figure}[t]
    \centering
    \subfigure[Studio-like Lab.]{
        \includegraphics[width=.29\columnwidth]{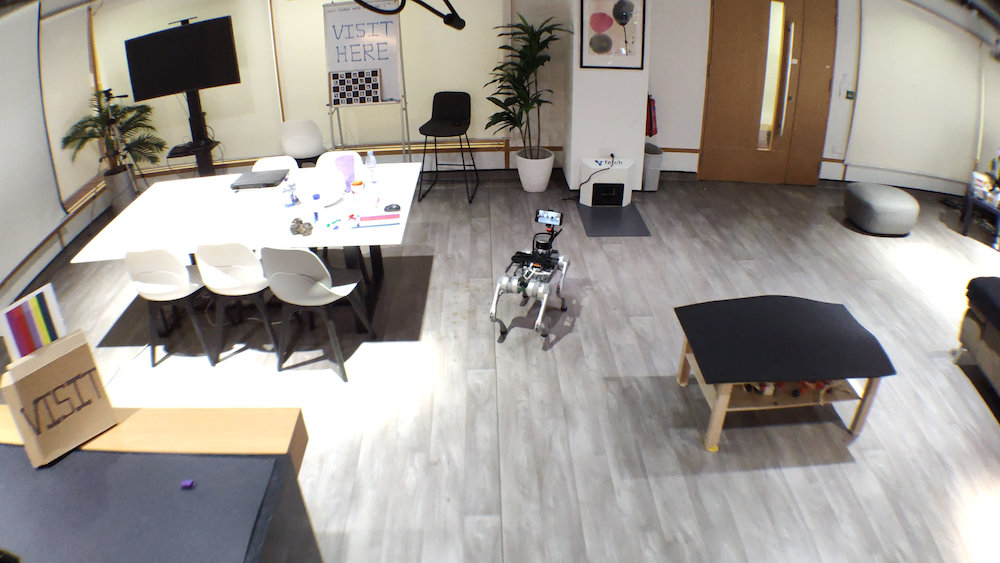}
        \label{fig:illustration_rls}
    }
    \hfill
    \subfigure[Office room.]{
        \includegraphics[width=.29\columnwidth]{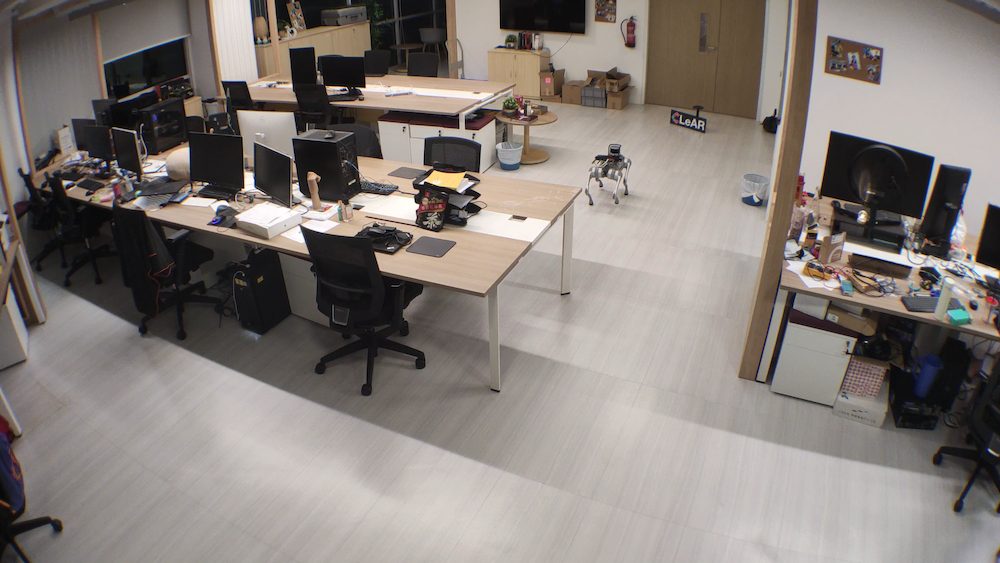}
        \label{fig:illustration_ssi}
    }
    \hfill
    \subfigure[Unitree Go2.]{
        \includegraphics[width=.29\columnwidth]{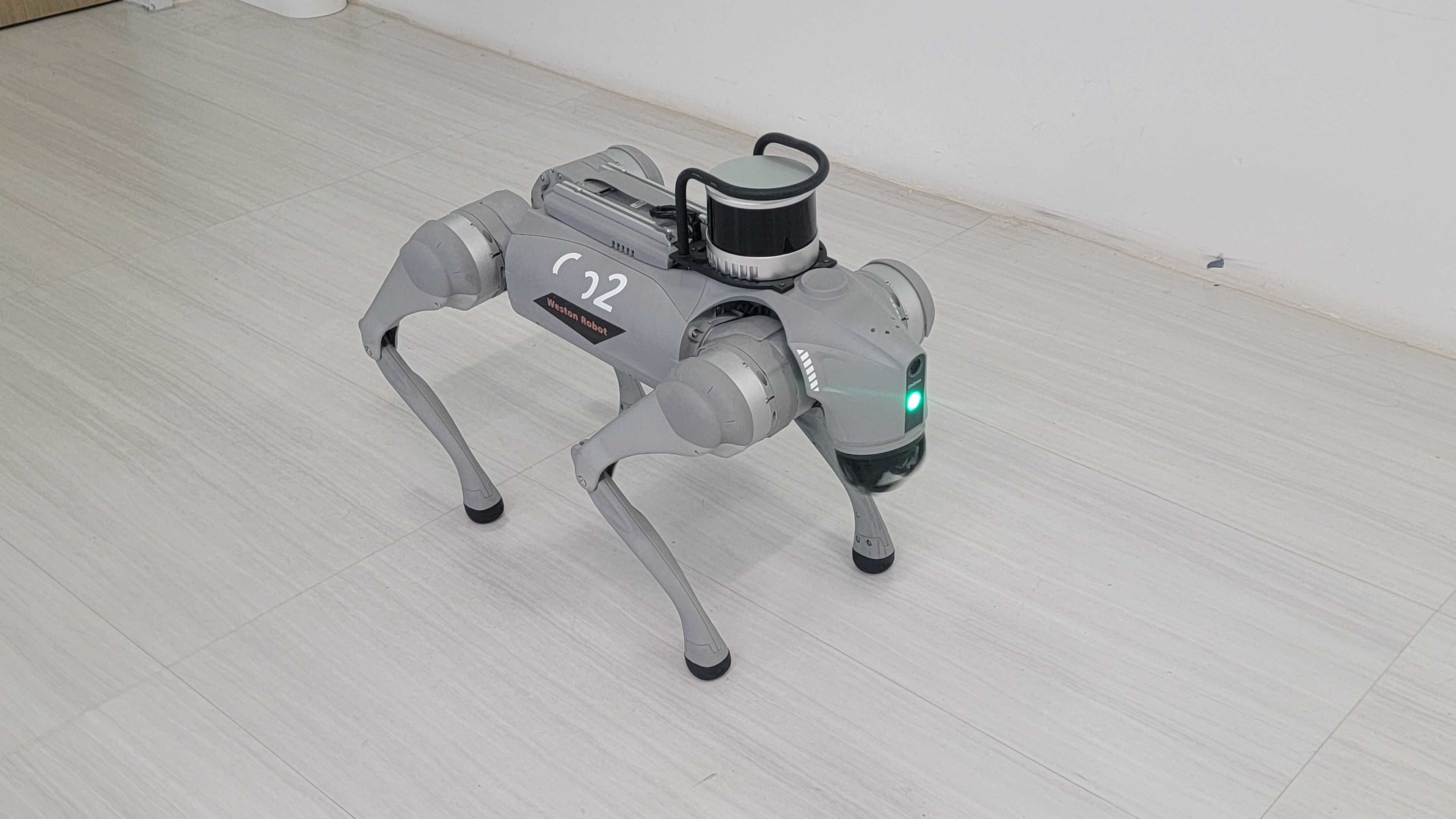}
        \label{fig:illustration_dog}
    }
    \vspace*{-5pt}
    \caption{Real world environments for quadruped robot navigation.}
    \label{fig:illustration_real}
\end{figure}

\section{Experiments}

\begin{table}[!t]
    \centering
    \begin{threeparttable}
    \caption{Performance on Avoidance Tasks in Maze2d}
    \label{table:maze2d_large_unsafe}
    \renewcommand{\tabcolsep}{9pt}
    \begin{tabular}{lccccccc} \toprule
        \multirow{2}{*}{\textbf{Method\textbackslash Perf.}} && \multicolumn{2}{c}{Satisfaction rate\tnote{1} ($\%$) $\uparrow$} & \multirow{2}{*}{\shortstack{Reward \\ (UnCon)}\tnote{2} $\uparrow$} \\ \cmidrule{3-4}
        && Planning & Rollout \\ \midrule
        \diffuser         && 9.5$\pm$3.1    & 11.0$\pm$1.7        & 142.2$\pm$5.1 \\
        \safediffuser     && 99.4$\pm$0.9   & 12.3$\pm$3.5        & 135.8$\pm$5.2 \\
        \methodps         && 99.0$\pm$0.8   & 73.0$\pm$3.0        &  97.3$\pm$2.8 \\
        \methodrg         && 98.8$\pm$0.8   & \bf{92.0$\pm$1.4}   & 127.1$\pm$5.1 \\ \bottomrule
    \end{tabular}
    \begin{tablenotes}
        \item[1] Mean and standard deviation calculated from $10$ groups of tests. Each test contains $100$ trials, where a trajectory is labeled as either satisfied or not satisfied in each trial. Best result during rollout is highlighted. Same for other tables of Maze2d.
        \item[2] Unconstrained rewards do \emph{not} take unsafe penalties into account.
    \end{tablenotes}
    \end{threeparttable}
\end{table}

\begin{table}[!t]
    \centering
    \begin{threeparttable}
    \caption{Results of Avoidance Task in PushT}
    \label{table:pusht_unsafe}
    \renewcommand{\tabcolsep}{3pt}
    \begin{tabular}{lccccc} \toprule
        \textbf{Method\textbackslash Perf.} && Satisfaction rate($\%$) $\uparrow$ & Overlap Score\tnote{1} $\uparrow$ \\ \midrule
        \diffusionpolicy  && 34.8$\pm$18.0      & 0.941$\pm$0.0584 & \\
        \methodps         && \bf{85.6$\pm$13.1} & 0.890$\pm$0.0647 & \\ 
        \methodrg         && \bf{85.6$\pm$12.5} & 0.842$\pm$0.0985 & \\ \bottomrule
    \end{tabular}
    \begin{tablenotes}
        \item[1] measures the final overlap of the \texttt{T} block and target area (min: $0.0$, max: $1.0$).
    \end{tablenotes}
    \end{threeparttable}
    \vspace*{-5pt}
\end{table}

\begin{figure}[t]
    \centering
    \subfigure[\diffuser.]{
        \includegraphics[width=.29\columnwidth]{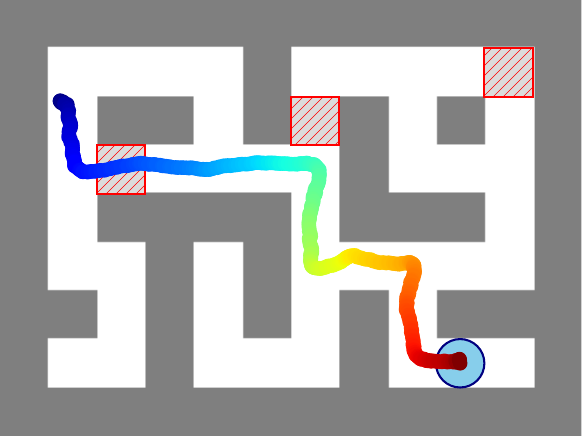}
        \label{fig:result_maze2d_large_unsafe_diffuser}
    }
    \hfill
    \subfigure[\footnotesize \safediffuser.]{
        \includegraphics[width=.29\columnwidth]{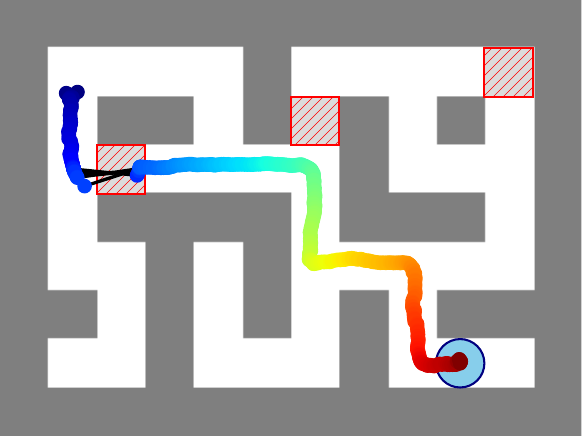}
        \label{fig:result_maze2d_large_unsafe_cbf}
    }
    \hfill
    \subfigure[\scriptsize \methodps.]{
        \includegraphics[width=.29\columnwidth]{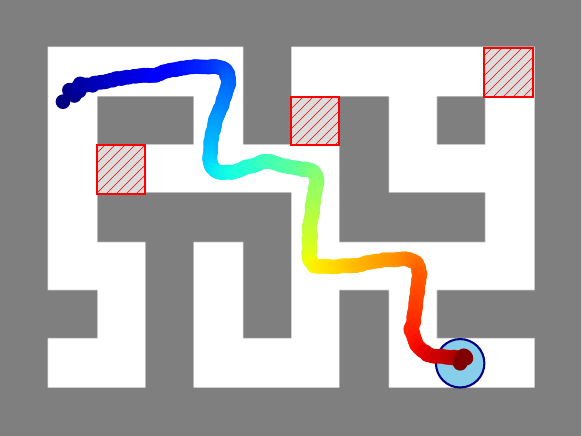}
        \label{fig:result_maze2d_large_unsafe_dps}
    }
    \vspace*{-5pt}
    \caption{Examples of safe planning in Maze2d-Large. There are three unsafe blocks (red squares, labeled $p_L$, $p_M$, $p_R$ from left to right) that need to be avoided during navigation to the goal (shaded circle). The $\ltlf$ constraint for this task is $\varphi = \ltlalways\neg(p_L \wedge p_M \wedge p_R)$. (a) Trajectories from \diffuser ignore safety and can violate the specified constraints. (b) \safediffuser{} produces discontinuous trajectories. (c) Our \method{} is able to plan trajectories that detours around the obstacles to successfully arrive at the goal.}
    \label{fig:result_maze2d_large_unsafe}
\end{figure}

\begin{figure}
    \centering
    \subfigure[\footnotesize PushT environment.]{
        \includegraphics[width=.29\columnwidth]{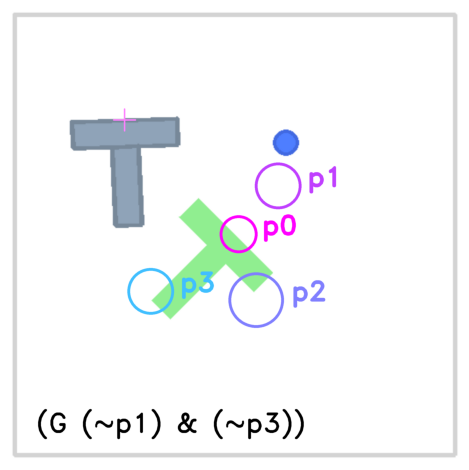}
        \label{fig:illustration_pusht_regions}
    }
    \hfill
    \subfigure[\scriptsize \diffusionpolicy.]{
        \includegraphics[width=.29\columnwidth]{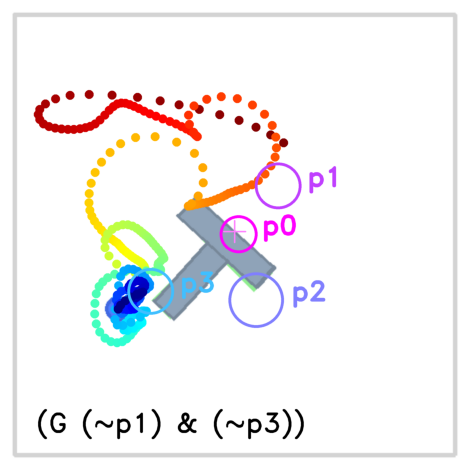}
        \label{fig:result_pusht_diffusionpolicy}
    }
    \hfill
    \subfigure[\scriptsize \methodps.]{
        \includegraphics[width=.29\columnwidth]{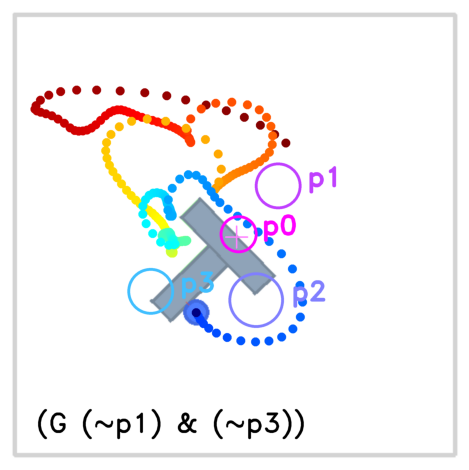}
        \label{fig:result_pusht_dps}
    }
    \vspace*{-5pt}
    \caption{
        Results of safe control in PushT. 
        (a) A robot arm's end effector (circles filled in blue) should manipulate the \texttt{T} block (gray) to a goal pose (green), and avoid entering unsafe regions (hollow circles marked with $pX$),  specified by an $\ltlf$ formula (text in black). 
        In this example, the $\ltlf$ specifies the end effector should never enter regions $p1$ (purple) and $p3$ (cyan). (b) The actions generated and executed by \diffusionpolicy do not satisfy the $\ltlf$ formula. (c) In contrast, \methodps guides the diffusion to avoid $p1$ and $p3$, yet still completes the manipulation task. 
    }
    \label{fig:illustration_pusht}
    \vspace*{-10pt}
\end{figure}

Our experiments focus on testing \method's ability to handle static  and temporal safety constraints. We first briefly describe the simulated and real environments, and baseline data-driven methods. Then we report empirical results on benchmark environments and demonstrate \method's applicability in real world tasks through a case study on a quadruped robot (\figref{fig:illustration_real}). We conclude with a brief ablation study and analysis. Due to space restrictions, we focus on conveying our main results. More details of the environments, implementations and analysis can be found in the appendix.

\subsection{Experimental Setup}

\parabf{Environments.} We evaluate methods in two simulation benchmark  environments (Maze2d~\cite{janner2022diffuser} and PushT~\cite{chi2023diffusionpolicy}) and demonstrate in two real indoor rooms. Maze2d (Fig.~\ref{fig:result_maze2d_large_unsafe} and \ref{fig:result_maze2d_LTL}) presents challenging long-horizon navigation tasks, where state-of-the-art offline algorithms fail to learn a reliable goal-reaching policy. The atomic propositions are determined by the occurrence of events when the agent is inside key regions in the maze (for avoidance and visitation). 
The PushT task (Fig.~\ref{fig:illustration_pusht}) requires manipulation of a \texttt{T} block through interaction with a controllable mover. In our experiments, the mover is constrained to visit specific regions and avoid others.
Our real-world experiments involve two indoor environments: a lab designed to mimic a studio apartment, and an office environment (\figref{fig:illustration_real}). 

\parabf{Compared methods.} Our work involves trajectory generation by learning from an offline dataset and as such, we compare against data-driven planning methods. \diffuser and \diffusionpolicy are state-of-the-art methods for sampling viable plans but without any guarantees over external constraints. To evaluate how well \method{} enforces safety specifications, we compare with \safediffuser, a safe planning diffusion model using CBFs. However, note that \safediffuser cannot handle temporal constraints or instructions; to our knowledge, our work is the first data-driven method to handle both static and temporal constraints. As such, there is no direct comparison baseline. 
We analyze the differences between the two variants of our method, \methodps and \methodrg.

\subsection{Comparative Analysis of Methods}

\parabf{Can \method achieve safe planning for static constraints?}
Our results indicate that yes, \method{} is better able to generate  trajectories that satisfy given region-avoidance constraints relative to existing methods. In both the Maze2D and PushT benchmarks, \method{} achieves high success rates (in both planning and rollout), without severely compromising reward accumulation (Tables~\ref{table:maze2d_large_unsafe} and ~\ref{table:pusht_unsafe}). Lower total rewards are expected since safe paths are typically longer and rollout scores are generally lower as the low-level controller may not exactly follow the diffused trajectory. 
Qualitatively, Fig.~\ref{fig:result_maze2d_large_unsafe} shows that \method{} is able to generate safe trajectories in Maze2D, whilst \safediffuser{} is limited to ``local'' deviations and fails to find paths that detour  around unsafe regions. Results in PushT are consistent with Maze2D; Fig. \ref{fig:illustration_pusht} shows that \method{} performs the orientation task without entering unsafe regions.

\begin{figure}
    \centering
    \subfigure[Regions for evaluation of atomic propositions in $\mathcal{P}$.]{
        \includegraphics[width=0.28\columnwidth]{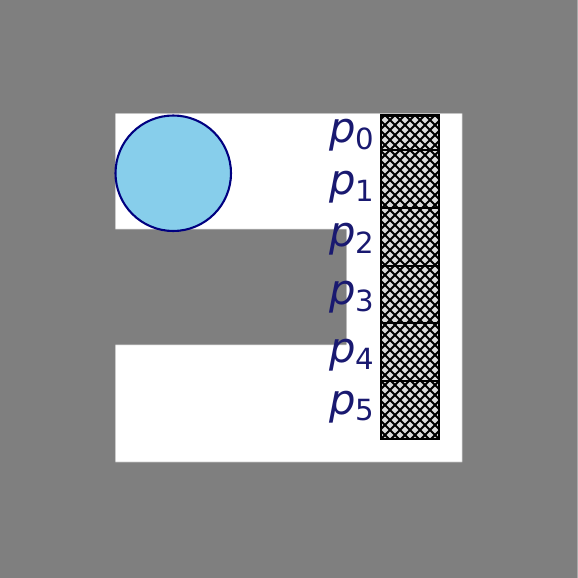}
        \includegraphics[width=0.28\columnwidth]{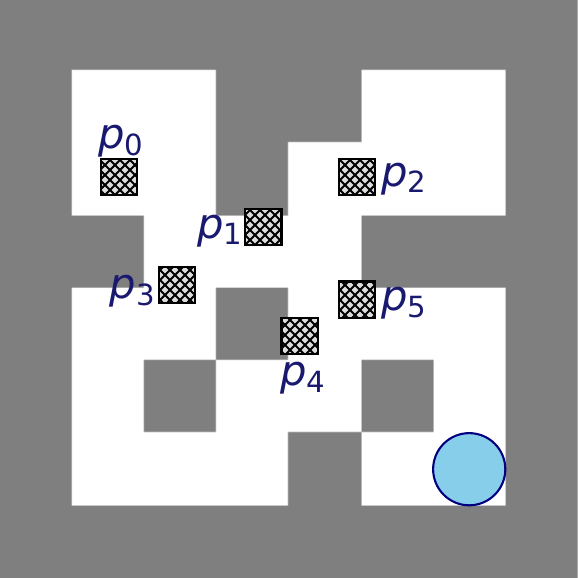}
        \includegraphics[width=0.38\columnwidth]{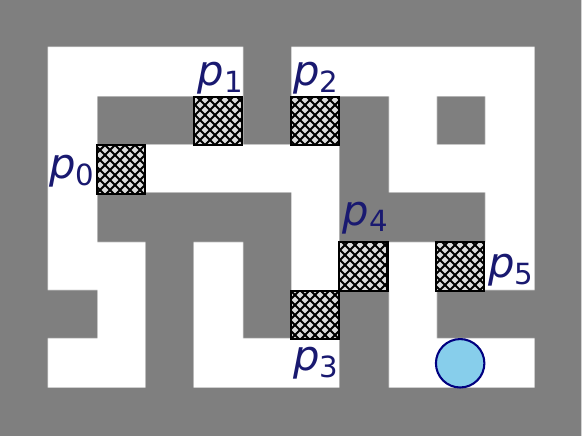}
        \label{fig:illustration_maze2d_atomic_locations}
    }
    \hfill
    \subfigure[\diffuser.]{
        \includegraphics[width=0.28\columnwidth]{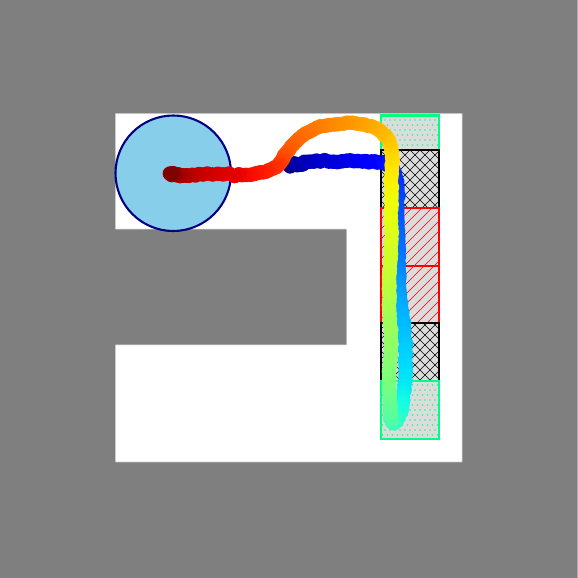}
        \includegraphics[width=0.28\columnwidth]{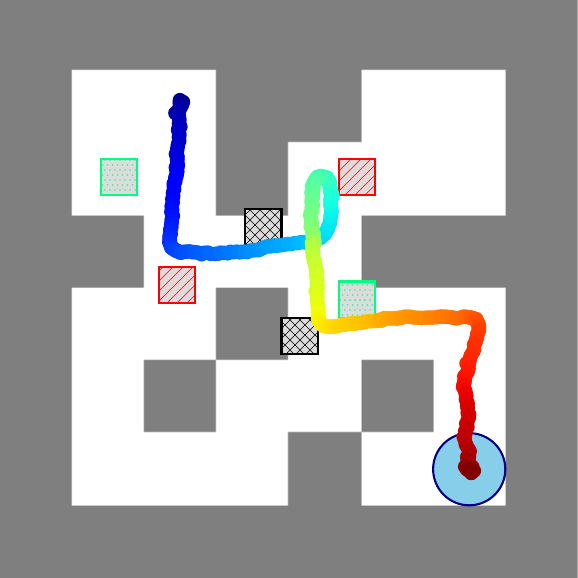}
        \includegraphics[width=0.38\columnwidth]{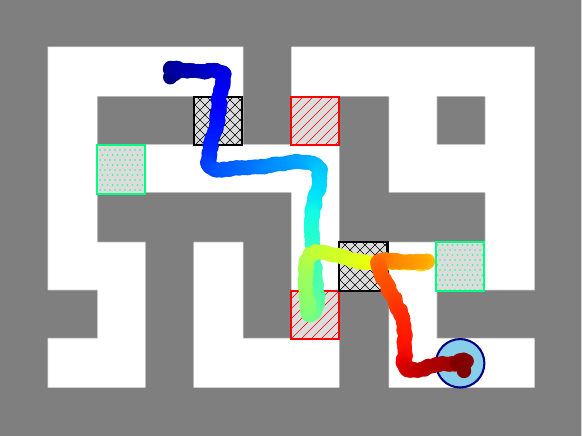}
        \label{fig:result_maze2d_LTL_diffuser}
    }
    \hfill
    \subfigure[Ours (\methodps).]{
        \includegraphics[width=0.28\columnwidth]{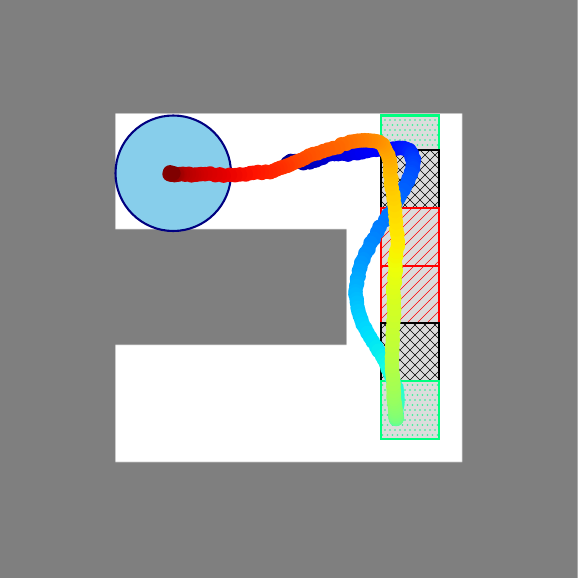}
        \includegraphics[width=0.28\columnwidth]{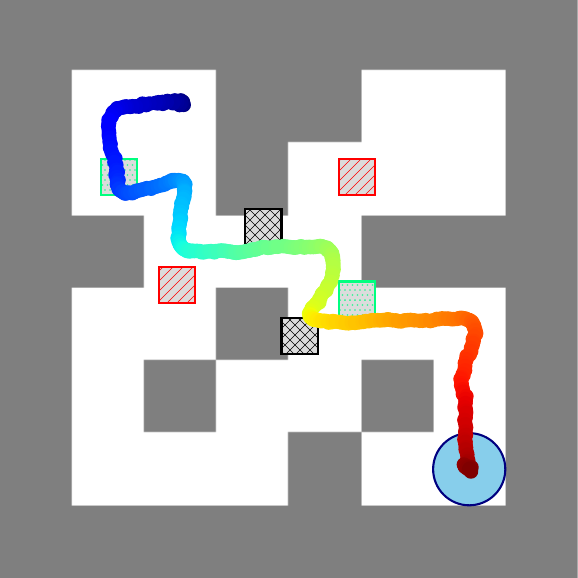}
        \includegraphics[width=0.38\columnwidth]{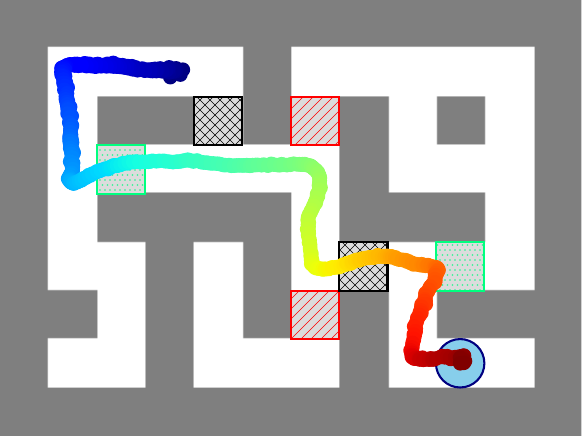}
        \label{fig:result_maze2d_LTL_ps}
    }
    \vspace*{-5pt}
    \caption{Temporal Constraints in Maze2D. (a) Each maze has 6 non-overlapping regions. Agents are requested to visit some of these blocks under different temporally-extended orders. (b) and (c) show generated trajectories under  $\varphi=\neg p_3\ltluntil(p_5 \wedge(\neg p_2\ltluntil p_0))$. Our method can satisfy $\neg$ propositions (red zones) before reaching the green regions.}
    \label{fig:result_maze2d_LTL}
    \vspace*{-10pt}
\end{figure}

\begin{table*}
    \centering
    \begin{threeparttable}
    \caption{Performance on Different $\ltlf$s in Maze2d. }
    \label{table:maze2d_ltl}
    \renewcommand{\arraystretch}{0.9}
    \begin{tabular}{llccccccc} \toprule
        \multirow{3}{*}{\textbf{Environment}} & \multirow{3}{*}{\textbf{Method\textbackslash Performance}} && \multicolumn{3}{c}{Training $\ltlf$s} & \multicolumn{3}{c}{Testing $\ltlf$s} \\ \cmidrule{4-9}
        &&& \multicolumn{2}{c}{Satisfaction rate ($\%$) $\uparrow$} & \multirow{2}{*}{Reward (UnCon) $\uparrow$} & \multicolumn{2}{c}{Satisfaction rate ($\%$) $\uparrow$} & \multirow{2}{*}{Reward (UnCon) $\uparrow$} \\ \cmidrule{4-5} \cmidrule{7-8}
        &&& Planning & Rollout && Planning & Rollout & \\ \midrule
        \multirow{3}{*}{\textbf{U-Maze (Horizon 256)}}  & \diffuser     && 31.1$\pm$0.5 & 31.0$\pm$0.5 & 33.5$\pm$2.7           & 33.9$\pm$0.5 & 34.1$\pm$0.6 & 35.6$\pm$0.3 \\
                                                        & \methodps     && 83.8$\pm$0.2 & \bf{57.6$\pm$1.3} & 31.3$\pm$1.2           & 82.7$\pm$0.3 & \bf{56.6$\pm$0.9} & 32.8$\pm$0.5 \\
                                                        & \methodrg     && 56.3$\pm$0.4 & 51.3$\pm$0.9 & 31.5$\pm$0.2           & 57.7$\pm$0.4 & 52.1$\pm$0.3 & 32.3$\pm$0.7 \\ \midrule
        \multirow{3}{*}{\textbf{Medium (Horizon 384)}}  & \diffuser     && 15.0$\pm$0.7 & 13.4$\pm$0.6 & 84.8$\pm$0.3           & 11.6$\pm$1.4 & 10.1$\pm$1.2 & 84.8$\pm$0.5 \\
                                                        & \methodps     && 77.9$\pm$5.7 & 31.8$\pm$2.6 & 53.1$\pm$5.2           & 68.4$\pm$6.7 & 28.7$\pm$3.5 & 50.5$\pm$4.7 \\
                                                        & \methodrg     && 51.8$\pm$1.8 & \bf{39.5$\pm$1.6} & 57.3$\pm$0.2           & 43.3$\pm$4.4 & \bf{30.6$\pm$1.9} & 57.7$\pm$0.1 \\ \midrule                                                        
        \multirow{3}{*}{\textbf{Large (Horizon 512)}}   & \diffuser     && 13.5$\pm$0.4 & 12.8$\pm$0.1 & 76.3$\pm$0.1         & 11.6$\pm$2.3 & 11.5$\pm$1.7 & 77.8$\pm$3.9 \\
                                                        & \methodps     && 73.8$\pm$2.4 & 32.6$\pm$1.4 & 42.3$\pm$5.0        & 66.6$\pm$2.7 & 24.9$\pm$1.7 & 40.9$\pm$4.4 \\
                                                        & \methodrg     && 66.9$\pm$0.6 & \bf{47.4$\pm$0.8} & 54.6$\pm$1.3        & 57.5$\pm$2.3 & \bf{39.0$\pm$2.9} & 54.5$\pm$3.9 \\ \bottomrule
    \end{tabular}
    \end{threeparttable}
    \vspace*{-10pt}
\end{table*}
\begin{table}
    \centering
    \begin{threeparttable}
    \caption{Generalization to Different $\ltlf$ formulae in PushT}
    \label{table:pusht_ltl}
    \renewcommand{\tabcolsep}{3pt}
    \begin{tabular}{llcccc} \toprule
        \textbf{Method\textbackslash Performance} & \textbf{$\ltlf$ Set} && Satisf. rate ($\%$) $\uparrow$ & Score $\uparrow$ \\ \midrule
        \multirow{2}{*}{\diffusionpolicy}    & Training && 22.9$\pm$8.0  & 0.354$\pm$0.153 \\
                                             & Test     && 30.7$\pm$13.9 & 0.371$\pm$0.177 \\
        \multirow{2}{*}{\methodps}           & Training && 28.2$\pm$8.33 & 0.290$\pm$0.115 \\ 
                                             & Test     && 43.0$\pm$17.0 & 0.299$\pm$0.145 \\ 
        \multirow{2}{*}{\methodrg}           & Training && \bf{69.3$\pm$9.90} & 0.292$\pm$0.121 \\ 
                                             & Test     && \bf{66.0$\pm$20.8} & 0.340$\pm$0.168 \\ 
                                             \bottomrule 
    \end{tabular}
    \end{threeparttable}
    \vspace*{-5pt}
\end{table}

\parabf{Can \method satisfy static and temporal constraints, and generalize to novel $\ltlf$ formulae?} 
Tables~\ref{table:maze2d_ltl} and \ref{table:pusht_ltl} show the performance of the compared methods on both training and test LTLs in Maze2d and PushT environments. To elaborate, we follow the $\ltlf$ specifications in~\cite{pmlr-v139-vaezipoor21a} and adopt the \texttt{Until} sampler to generate random $\ltlf$s ($200$ for Maze2d and $36$ for PushT) that contain different visitation sequences and unsafe regions. The training set has $80\%$ of all $\ltlf$s and the rests are used as test set. Atomic propositions consist of 6 regions in Maze2d (Fig.~\ref{fig:illustration_maze2d_atomic_locations}, \ie, $p_0, p_1,\ldots,p_5$) and 4 regions in PushT (Fig.~\ref{fig:illustration_pusht_regions}).

Results reveal that both \methodps and \methodrg achieve significantly higher success rates than \diffuser/\diffusionpolicy. The baselines have a non-zero performance as some generated $\ltlf$ formulae are trivial to satisfy at some start locations. Also note that some specifications may be impossible to satisfy given the physical locations of the agent, walls, and propositional regions in the maze. In these cases, planned trajectories will violate constraints during rollout, causing a drop in the satisfaction rate. Fig.~\ref{fig:result_maze2d_LTL} shows examples of trajectories generated in Maze2d. 

\begin{figure}
    \centering
    \subfigure[Lab map and dataset coverage.]{
        \includegraphics[width=.295\columnwidth]{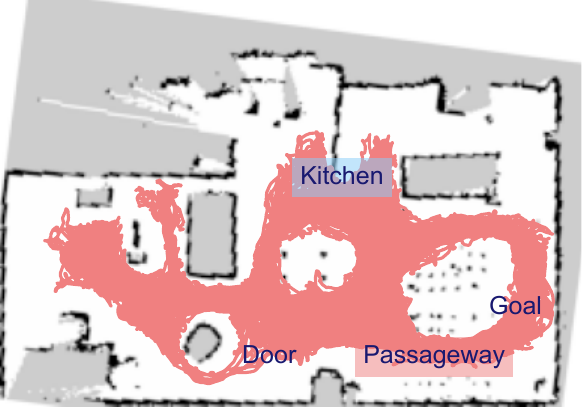}
        \label{fig:illustration_map_data_rls}
    }
    \hfill
    \subfigure[\diffuser.]{
        \includegraphics[width=.295\columnwidth]{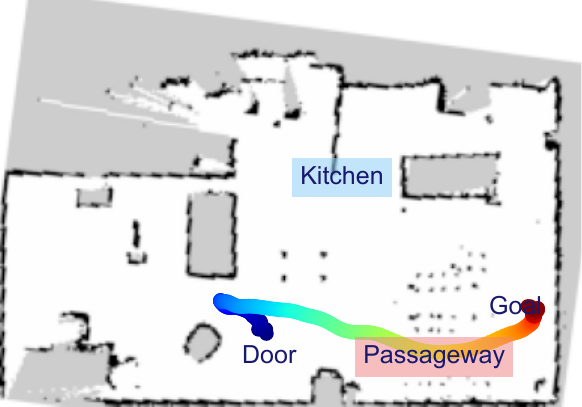}
        \label{fig:result_rls_ltl_diffuser}
    }
    \hfill
    \subfigure[Ours.]{
        \includegraphics[width=.2951\columnwidth]{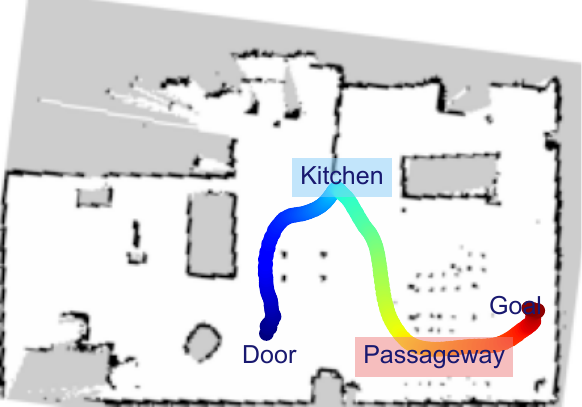}
        \label{fig:result_rls_ltl_ps}
    }
    \hfill
    \subfigure[Office map and dataset coverage.]{
        \includegraphics[bb = 15 0 150 215,clip=true,width=.295\columnwidth]{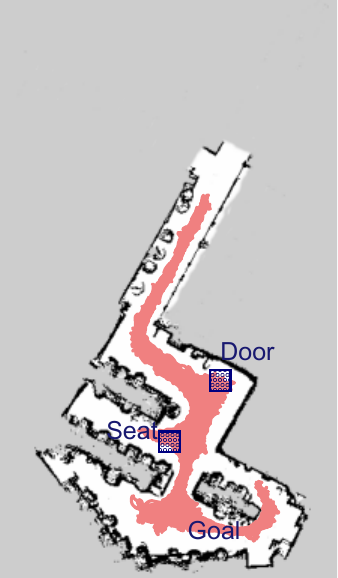}
        \label{fig:illustration_map_data_ssi}
    }
    \hfill
    \subfigure[\diffuser.]{
        \includegraphics[bb = 15 0 150 215,clip=true,width=.295\columnwidth]{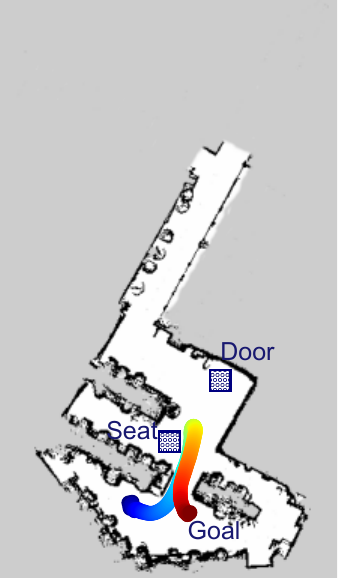}
        \label{fig:result_ssi_ltl_diffuser}
    }
    \hfill
    \subfigure[Ours.]{
        \includegraphics[bb = 15 0 150 215,clip=true,width=.2951\columnwidth]{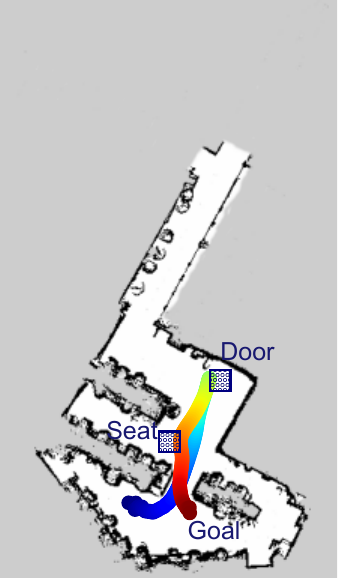}
        \label{fig:result_ssi_ltl_ps}
    }
    \vspace*{-10pt}
    \caption{Results in real world rooms. The instructed $\ltlf$ is $\varphi = \neg \text{Passageway} \ltluntil \text{Kitchen}$ for lab (first row) and $\varphi = \ltleventually\left( \text{Door} \wedge \ltleventually\text{Seat} \right)$ for office (second row). In the lab task the robot has to first unload in the kitchen area after entering the door before being allowed to go to the goal. When loaded, the robot is prohibited from going through the narrow passageway where people sit on the side. In the office task, the robot has to first visit the seat, followed by the door, before the goal.}
    \label{fig:result_rls_ltl}
    \vspace*{-10pt}
\end{figure}

\begin{table}
    \centering
    \begin{threeparttable}
    \caption{Results of achieving goals and $\ltlf$ in real-world tasks.}
    \label{table:result_sr_lab}
    \begin{tabular}{llccc} \toprule
        \multirow{2}{*}{\textbf{Environment}} & \multirow{2}{*}{\textbf{Method\textbackslash Performance}} && \multicolumn{2}{c}{Satisfaction rate ($\%$) $\uparrow$} \\ \cmidrule{4-5}
        &&& Goal & $\ltlf$ \\ \midrule
        \multirow{2}{*}{\textbf{Lab}}       & \diffuser     && 100.00 & 0.00 \\
                                            & \methodps     && 91.67 & \bf{91.67} \\ \midrule
        \multirow{2}{*}{\textbf{Office}}    & \diffuser     && 100.00 & 0.00 \\
                                            & \methodps     && 95.83 & \bf{95.83} \\ \bottomrule
    \end{tabular}
    \end{threeparttable}
    \vspace*{-10pt}
\end{table}

\subsection{Real Robot Case Study.}

We show that \method can plan for a robot dog (Unitree Go2 Edu) given $\ltlf$ instructions in two real-world navigation environments --- a lab that mimics a studio apartment and an office room. Training of diffusion models was performed in simulation using Gazebo and ROS1 using LIDAR scanned maps (Fig.~\ref{fig:illustration_map_data_rls} and \ref{fig:illustration_map_data_ssi}). Note that the training trajectories do not require running an oracle policy to satisfy many different $\ltlf$ formulae; we simply made the robot navigate to randomly sampled goals  using the global planner and TEB Local Planner~\cite{Rsmann2017IntegratedOT} from the ROS1 Navigation stack  (Fig.~\ref{fig:illustration_map_data_rls} and \ref{fig:illustration_map_data_ssi}).

To test on potential constraints, we queried GPT-4 for $\ltlf$ formulae representing meaningful robotic tasks including obstacle avoidance and sequential navigation. We used 4 kinds of $\ltlf$s from the generated results, \ie, $\ltlalways\neg p_0$, $\ltleventually p_0\wedge\ltleventually p_1$, $\ltleventually\left(p_0\wedge\ltleventually p_1\right)$ and $\neg p_1 \ltluntil p_0$. The first $\ltlf$ corresponds to an obstacle avoidance task where the robot should never visit a specific region. For example, $\ltlalways\neg \text{Seat}$ means the robot should never enter the Seat region (Fig~\ref{fig:illustration_map_data_ssi}). The remaining three $\ltlf$ formulae represent: 1) visiting all regions at least once; 2) visiting regions in a specific sequence; 3) avoiding a specific region until another has been visited. Some example regions designed in our real environments and start/goal locations are shown in Fig~\ref{fig:result_rls_ltl}.

In total, $96$ trials were executed on the real robot, involving $12$ trajectories ($6$ for baseline \diffuser and $6$ for our method \methodps) with varying regions and different $\ltlf$ formulae in each room. The overall satisfaction rate of all raw generated trajectories in \emph{simulation} is $85.8\pm14.0\%$ (\emph{c.f.} baseline $2.9\pm5.4\%$). 
For each specific formula, a sample was selected for real-world execution based on their feasibility.
The results in~\tabref{table:result_sr_lab} show that \method{} has a high satisfaction rate compared to \diffuser{};  Fig.~\ref{fig:result_rls_ltl_ps} and~\ref{fig:result_ssi_ltl_ps} illustrate trajectories for qualitative comparison.

\subsection{Ablation Study and Analysis}
\label{sec:ablation}

\begin{table}
    \centering
    \begin{threeparttable}
    \caption{Ablation Study -- Binary Classifier Guidance}
    \label{table:maze2d_umaze_ltl_cg_rg}
    \renewcommand{\tabcolsep}{3pt}
    \begin{tabular}{llcccc} \toprule
        \multirow{2}{*}{\textbf{Method\textbackslash Performance}} & \multirow{2}{*}{\textbf{LTL Set}} && \multicolumn{2}{c}{Satisfaction rate ($\%$) $\uparrow$} & \multirow{2}{*}{\shortstack{Reward \\ (UnCon)} \shortstack{$\uparrow$}} \\ \cmidrule{4-5}
        &&& Planning & Rollout & \\ \midrule
        \multirow{2}{*}{\diffuser}           & Training && 31.1$\pm$0.5 & 31.0$\pm$0.5 & 32.5$\pm$2.7 \\
                                             & Test     && 33.9$\pm$0.5 & 34.1$\pm$0.6 & 35.6$\pm$0.3 \\
        \multirow{2}{*}{Classifier guidance\tnote{1}} & Training && 41.1$\pm$0.6 & 40.6$\pm$0.7 & 33.9$\pm$0.9 \\
                                             & Test     && 40.0$\pm$0.8 & 41.8$\pm$1.1 & 35.3$\pm$0.3 \\
        \multirow{2}{*}{\methodrg}           & Training && 56.3$\pm$0.4 & \bf{51.3$\pm$0.9} & 31.5$\pm$0.2 \\ 
                                             & Test     && 57.7$\pm$0.4 & \bf{52.1$\pm$0.3} & 32.3$\pm$0.7 \\ \bottomrule 
    \end{tabular}
    \begin{tablenotes}
        \item[1] The classifier guidance method only leverages \emph{binary labels} for LTL satisfaction checking in Maze2d U-Maze.
    \end{tablenotes}
    \end{threeparttable}
    \vspace*{-5pt}
\end{table}

\parabf{Ablation study.} Unlike classifier guidance, where each trajectory is labelled as \emph{satisfy} or \emph{not satisfy}, we leverage the continuous values from our formula evaluator (as described in Section~\ref{sec:dtl_rg}) and train a regressor guidance network. Although classifier guidance improves over the vanilla \diffuser, it achieves lower performance than \methodrg ( Table~\ref{table:maze2d_umaze_ltl_cg_rg}). This comparison supports the notion that soft labels improves the guidance neural network; we posit using real values provides richer information in terms of how \emph{well} the trajectory satisfies a given $\ltlf$ formula. 

\begin{figure}[tb]
    \centering
    \subfigure[Planning.]{
        \includegraphics[width=.46\columnwidth]{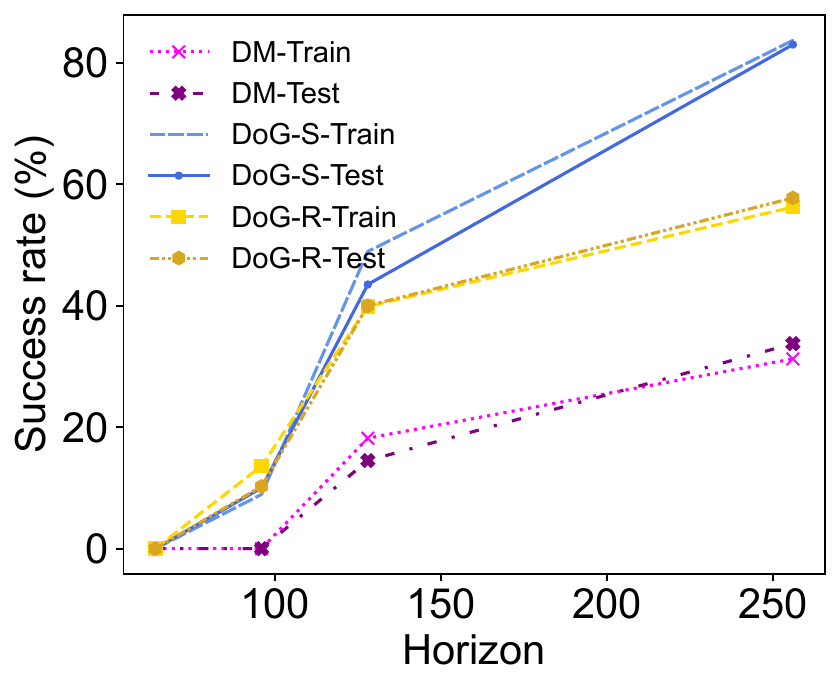}
        \label{fig:result_maze2d_umaze_ltl_horizon_train}
    }
    \hfill
    \subfigure[Rollout.]{
        \includegraphics[width=.46\columnwidth]{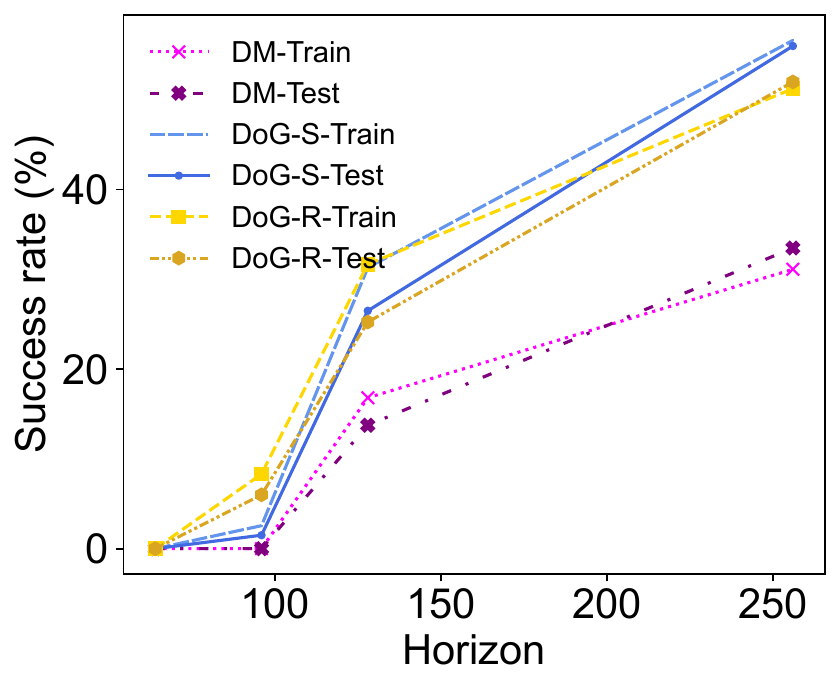}
        \label{fig:result_maze2d_umaze_ltl_horizon_test}
    }
    \vspace*{-5pt}
    \caption{Performance in Maze2d U-Maze with different lengths of trajectory. DM stands for \diffuser and DoG-S/DoG-R are our methods.}
    \label{fig:result_maze2d_umaze_ltl_horizon}
    \vspace*{-10pt}
\end{figure}

\parabf{Analysis on horizon.}  Fig.~\ref{fig:result_maze2d_umaze_ltl_horizon} demonstrates that planning with a longer horizon leads to improved performance in terms of $\ltlf$ satisfaction. This improvement is attributed to the fact that $\ltlf$ instructions often require a longer sequence of steps in a path compared to mere goal navigation, \eg, visiting a specific region before reaching the goal.

%% file: sections/conclusion.tex
\section{Conclusion, Discussion and Future Work} 

In this work, we presented \method{}, an approach towards generating safe trajectories that comply with $\ltlf$ specifications at test time. Within our overall scheme, we presented two methods: \methodps  guides the sampling process under any $\ltlf$ formula while \methodrg uses a trained model that we show generalizes to new  formulae with similar structure. To our knowledge, this work is the first that successfully incorporates model checking using a formal language with diffusion models for safe planning. Notably, \method{} does not require data collection for each potential $\ltlf$ instruction; rather, we control the sampling process during diffusion using ``soft'' model checking to generate new trajectories using existing information provided by the training dataset.

\parabf{Limitations and Future Work.} \method{} is a step towards trustworthy trajectory planning using generative models. There are several areas where \method{} can be improved. Similar to other diffusion models, \method{} is generally unable to generate realistic trajectories when the context (environment/goal) is far from the training distribution. As such, the dataset should preferably contain trajectories with a variety of complex behaviors that can potentially satisfy different test-time $\ltlf$ formulae. It would be interesting to develop methods to ensure the sampling process of \methodps adheres to the data manifold. Additionally, \method{} plans in an open-loop fashion, where the entire trajectory is generated conditioned on the constraints. 
This is mainly  because the evaluation of an $\ltlf$ formula depends on the entire trajectory. We plan to further explore planning using receding horizon control using partial evaluations on an incomplete trajectory. 
Finally, diffusion models usually require significant amounts of training data and many diffusion steps during inference. Recent work on interpolant diffusion methods~\cite{chen2024behavioral} leverages source policies to reduce data and computation costs. We aim to explore how integrating this approach with conditional sampling using $\ltlf$ can enhance the generation of safe trajectories.

%% file: sections/acks.tex
\section*{Acknowledgements}

This research is supported by A*STAR under its National Robotics Programme (NRP) (Award M23NBK0053).  The authors would also like to acknowledge partial support from a Google South Asia \& Southeast Asia Award and from the National Research Foundation, Singapore under its Medium Sized Center for Advanced Robotics Technology Innovation.

%% file: sections/appendix.tex
\subsection{Data}
\label{appsec:data}

    \subsubsection{\texttt{Maze2d}}
\begin{figure}[ht]
        \centering
        \subfigure[U-Maze.]{
            \includegraphics[width=.245\columnwidth]{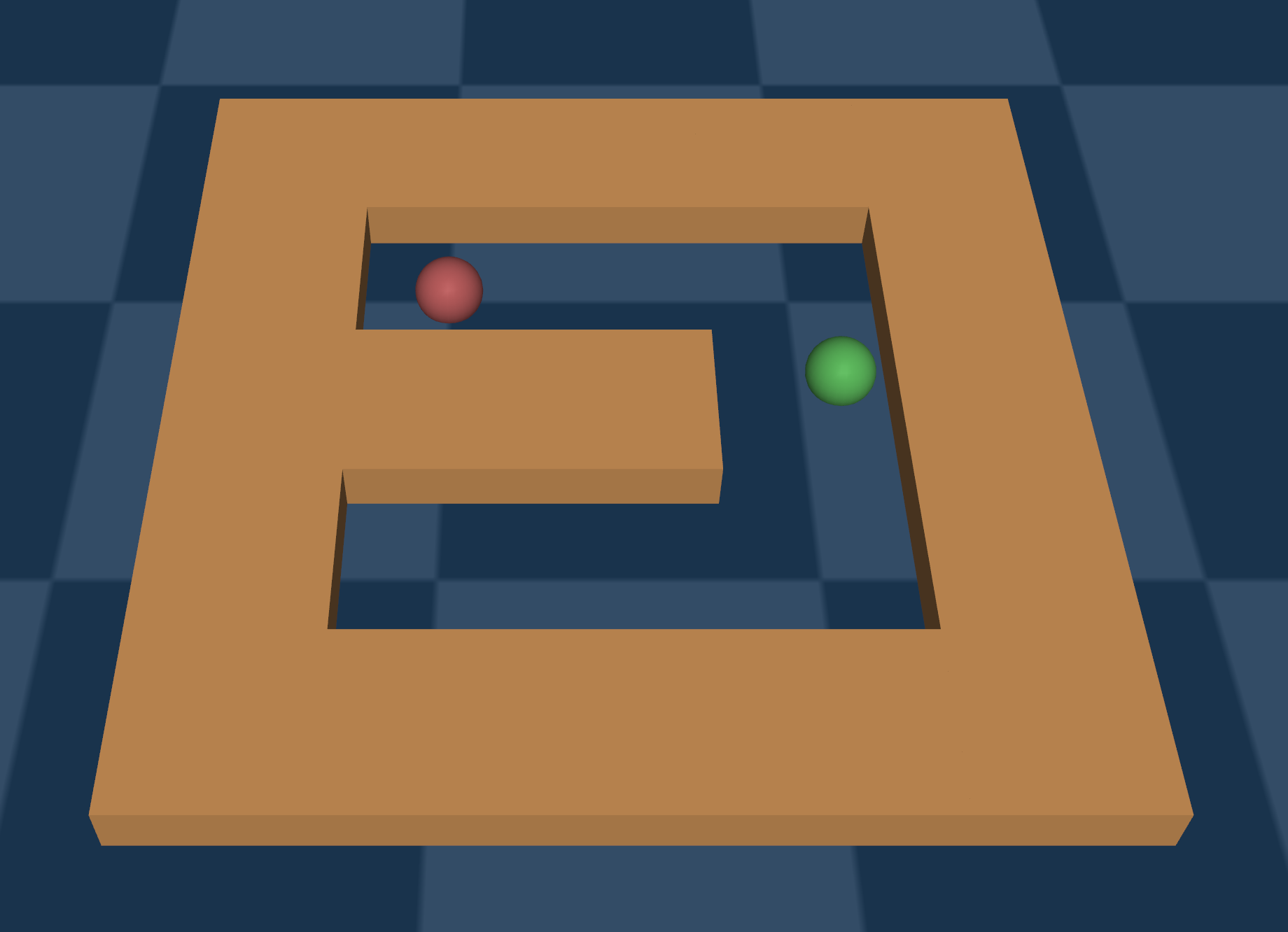}
            \label{fig:illustration_maze2d_umaze}
        }
        \hfill
        \subfigure[Medium.]{
            \includegraphics[width=.255\columnwidth]{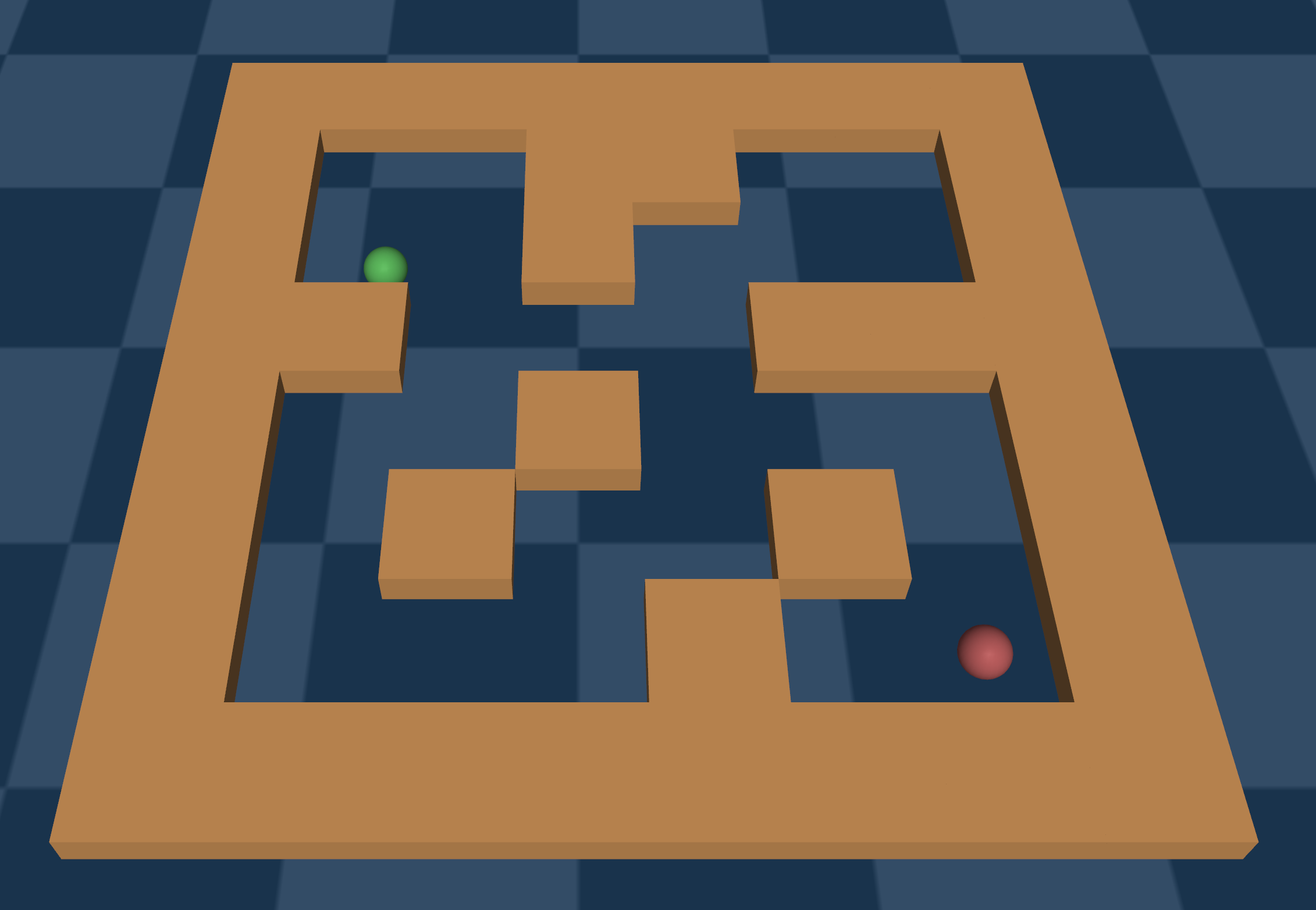}
            \label{fig:illustration_maze2d_medium}
        }
        \hfill
        \subfigure[Large.]{
            \includegraphics[width=.38\columnwidth]{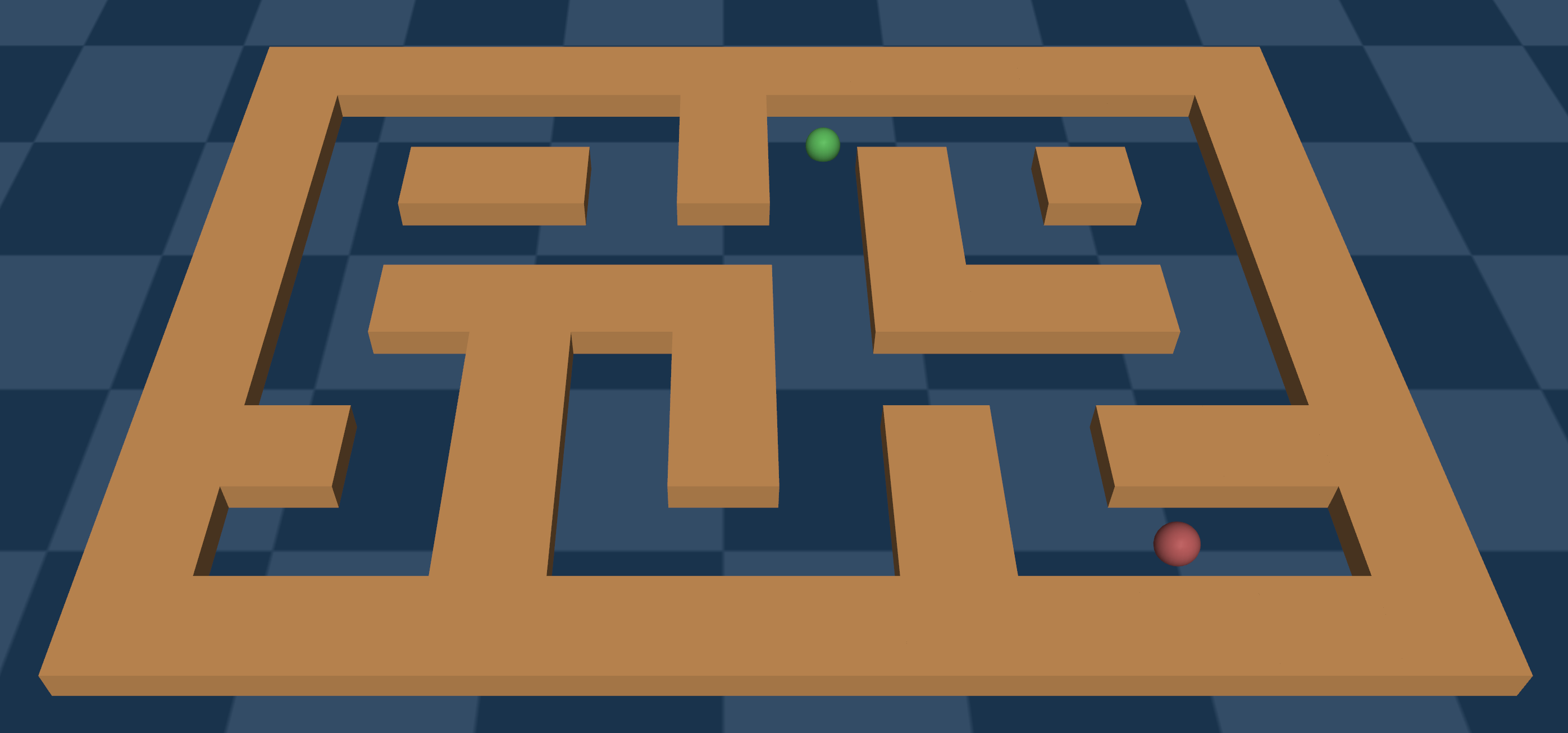}
            \label{fig:illustration_maze2d_large}
        }
        \caption{Maze2d environments. The agent is tasked to move force-actuated ball to goals (red circle) from random start locations (green ball). There are three kinds of maze configurations with increasing difficulties and all of them are long-horizon path planning tasks with continuous states and actions.
        }
        \label{fig:illustration_maze2d}
        \vspace*{-5pt}
    \end{figure}

    The Maze2d environment is part of a well-known open-source benchmark for offline reinforcement learning --- D4RL~\cite{fu2020d4rl}.  
    Datasets of trajectories are publicly available in the D4RL library. Each of the trajectories within is one continuous trajectory of an agent navigating to random goals without interruption. 
    See \figref{fig:illustration_maze2d} for an overview of the environments. 
    We utilize the original dataset in D4RL as is to train the baseline \diffuser, which also serves as the diffusion model of \method.  

    \subsubsection{\texttt{PushT}}
    PushT is a robot manipulation benchmark adopted in \diffusionpolicy~\cite{chi2023diffusionpolicy} to showcase diffusion models' planning capabilities. 
    A dataset of expert policies for pushing a \texttt{T} block to the same goal position starting from random poses are publicly available at~\cite{chi2023diffusionpolicy}. 
    Though \diffusionpolicy achieves good performance on the original manipulation task, trained with around only 200 expert demonstrations (\ie, 206 episodes), we find that these trajectories lack the modalities or abundant behaviors that potentially or partially satisfy different $\ltlf$ constraints. 
    As such, expanding the original dataset is necessary for our method to work. 

    \parabf{Data Augmentation} 
    Given that obtaining extra expert demonstrations is inefficient, we performed common data augmentation techniques to expand the original dataset. 
    Concretely, we first employed simple geometric augmentation, only rotation and flipping, for each trajectory in the original dataset, assuming basic physical symmetry in this simulation environment. 
    Then, we used a pretrained \diffusionpolicy to push the \texttt{T} block to the goal pose, aligning with the original dataset in which the goal of the manipulation task is implicitly encoded in trajectories. 
    In this way, we augmented the original set from 206 trajectories to 6686 trajectories, and elongated the average episode length from 125 to 289 steps. 
    The augmented dataset is publicly available online\footnote{\url{https://github.com/clear-nus/ltldog}}.

    In this task, the baseline (\diffusionpolicy) and our methods share the same diffusion model backbone, which is trained over the augmented dataset.

    \subsubsection{Real Robot Scenarios}
    Data for both of the real-world scenarios are collected in simulated environments. For the office we collected 4 long continuous trajectories in the simulation of step size of 11,000, 107,000, 60,000 and 130,000. Similarly for the lab map we collected 4 trajectories of steps 11,000, 19,000, 110,000 and 120,000.  
    First, we built by SLAM the maps of the two environments, a studio-like lab and an office room, and the maps are used to create two simulation scenarios. 
    For creating the maps, we teleoperated a quadruped robot (Unitree Go2) mounted with LiDAR in each of these environments. We used the \texttt{slam\_toolbox} along with \texttt{Nav2} for generating the maps in both environments. We created simple 3D world for each of the maps for simulation in gazebo. 
    After that, we enabled a Go2 robot simulated in Gazebo and driven by ROS Navigation Stack to navigate to random goal positions sampled in the maps for data collection, creating long trajectories in a similar fashion to Maze2d.

    \subsubsection{LTL Satisfaction Values}
    The soft labels (values indicating how well the $\ltlf$s are satisfied) for training the regressor guide (\methodrg) are calculated with an adapted version of the DTL toolbox~\cite{xu2022dont} and saved before each training session, and can be reused if possible, since each training session may sample and cut different segments of the trajectories in datasets based on different training configurations. 
    Values calculation for benchmarking test results are performed online simultaneously with all tests.

\subsection{$\ltlf$ Generation}
\label{appsec:ltlf_generation}
To generate practical $\ltlf$s meaningful for real robotic tasks, we query GPT-4 via prompting with some constraints on $\ltlf$s. The prompt and the response from ChatGPT is listed in \tabref{table:chatgpt}. In the real experiments, we tested the first four returned $\ltlf$s as these are the most representative $\ltlf$s among them and are also commonly used in robot learning literature~\cite{10.5555/3237383.3237452,ijcai2019p840,pmlr-v139-vaezipoor21a}. Some of the other generated $\ltlf$s shown in this table are either not logically correct or not meaningful for robotic tasks. For instance, the $5$th $\ltlf$ has a similar meaning of sequential visiting as the $2$nd $\ltlf$. The imply operator does not enforce any constraints if the condition is not met (\eg, the condition in the $6$th $\ltlf$ can never be satisfied for disjoint regions). The $7$th, $8$th and $9$th $\ltlf$s are not correct with their corresponding descriptions. The last $\ltlf$ can be considered as a combination of the $2$nd and $3$rd $\ltlf$s. More complex $\ltlf$s and combinations of them can be derived based on different visitation and avoidance patterns~\cite{8859226}. We plan to further explore these temporal specifications in environments like Robomimic, and adapting our method to complex combinations of $\ltlf$s in future work.

\begin{table*}
    \centering
    \begin{threeparttable}
    \caption{Prompts and Responses from ChatGPT (GPT-4)\tnote{1} for $\ltlf$ Generation.}
    \label{table:chatgpt}
    \begin{tabular}{llp{0.8\textwidth}} \toprule
        \textbf{Prompt} && In a robot navigation task, the robot is required to visit four key regions (referred to as p\_0, p\_1, p\_2 and p\_3) in specific temporal orders. The constraints on the robot's trajectory concerning the visitation (or non-visitation) of these regions can be specified using finite Linear Temporal Logic (LTLf). Please generate several LTLf formulas (e.g., 10) that are realistic in real-world environments and tasks involving robot navigation. The details of the task and additional requirements are as follows:\vspace{0.1cm}

        1. The LTLf operators that can be used are: $\neg$ (Negate), $\wedge$ (And), $\vee$ (Or), F (Eventually), G (Always), U (Until). Avoid using other LTL operators. Also, avoid using X (Next) since the number of timesteps needed to move from one region to another is unknown.\vspace{0.1cm}
        
        2. The atomic propositions are determined by whether the robot is inside p\_0, p\_1, p\_2 and p\_3. Therefore, the labelling function returns assignments with a 4-dimensional 0-1 vector at each timestep of the robot's trajectory.\vspace{0.1cm}
        
        3. Only one proposition can be true at each timestep because these regions are disjoint. The robot cannot be inside any two regions simultaneously. An assignment at a timestep can be a zero vector (0, 0, 0, 0), indicating that the robot is not currently in any of these regions. This occurs when the robot is initially not in these regions and all atomic propositions are False.\vspace{0.1cm}
        
        4. The robot cannot move directly from one region to another as the regions might not be connected. Therefore, LTLf formulas such as p\_0 U p\_1 are not valid since both propositions will be False when the robot is not inside these two regions.\vspace{0.1cm}
        
        5. The robot's trajectory is finite, so the robot cannot repeatedly visit different regions.\vspace{0.1cm}
        
        6. The LTLf formulas are supposed to apply from the first timestep of the assignment.\vspace{0.1cm}
        
        7. When generating an LTLf formula, please analyze the generated formula first to ensure that it is valid and that there exist assignments that can satisfy it. \\ \midrule
        \textbf{ChatGPT}  && To create realistic LTLf formulas for a robot navigation task considering the specified constraints and operators, we need to structure these formulas in a way that reflects possible navigation scenarios. Here are ten LTLf formulas that align with the given requirements:

        \begin{enumerate}
            \item   Visit all regions at least once:
                    \begin{itemize}
                        \item $F(p_0) \wedge F(p_1) \wedge F(p_2) \wedge F(p_3)$
                        
                        This formula states that eventually, the robot will visit each of the four regions at least once during its trajectory.
                    \end{itemize}
            \item   Visit regions in a specific sequence:
                    \begin{itemize}
                        \item $F(p_0 \wedge F(p_1\wedge F(p_2 \wedge F(p_3))))$
                        
                        This formula requires the robot to visit $p_0$, then $p_1$, followed by $p_2$, and finally $p_3$, in that order.
                    \end{itemize}
            \item   Avoid a specific region until another has been visited:     
                    \begin{itemize}
                        \item $\neg p_1 \ltluntil p_0$
                        
                        Here, the robot must avoid $p_1$ until $p_0$ has been visited.
                    \end{itemize}
            \item   Never visit a specific region:
                    \begin{itemize}
                        \item $G(\neg p_3)$
                        
                        This formula specifies that the robot should never enter region $p_3$.
                    \end{itemize}
            \item   Visit one region only after visiting two others:
                    \begin{itemize}
                        \item $F(p_0 \wedge F(p_1)) \ltlimplies F(p_2)$
                        
                        The robot can visit $p_2$ only after it has visited $p_0$ and then $p_1$.
                    \end{itemize}
            \item   Ensure two regions are visited before a third:
                    \begin{itemize}
                        \item $F(p_0\wedge p_1) \ltlimplies F(p_2)$
                        
                        This formula ensures that $p_2$ and $p_1$ are visited (in no particular order) before visiting $p_2$:
                    \end{itemize}
            \item   Visit at least three regions:
                    \begin{itemize}
                        \item $F(p_0) \vee F(p_1) \vee F(p_2) \vee F(p_3)$
                        
                        This ensures the robot visits at least three out of the four regions. It is a less strict requirement, allowing for flexibility in navigation.
                    \end{itemize}
            \item   Visit a specific region last:
                    \begin{itemize}
                        \item $F(p_0\wedge F(p_1 \wedge F(p_2))) \wedge G(\neg p_3\ltluntil(p_0\wedge F(p_1\wedge F(p_2))))$
                        
                        The robot must visit $p_3$ last, after visiting $p_0$, $p_1$, and $p_2$ in any order.
                    \end{itemize}
            \item   Avoid a region until all other regions have been visited:
                    \begin{itemize}
                        \item $\neg p_3\ltluntil (F(p_0) \wedge F(p_1) \wedge F(p_2))$
                        
                        The robot should avoid $p_3$ until it has visited $p_0$, $p_1$, and $p_2$ in any order.
                    \end{itemize}
            \item   Visit a pair of regions, avoiding others until then:
                    \begin{itemize}
                        \item $\neg(p_2\vee p_3)\ltluntil(p_0\wedge F(p_1))$
                        
                        The robot must avoid $p_2$ and $p_3$ until it has visited $p_0$ and then $p_1$:
                    \end{itemize}
        \end{enumerate}

        Each of these formulas considers the finite nature of the robot's trajectory, the impossibility of being in multiple regions simultaneously, and the requirement that the robot may need to travel through unspecified areas (not within $p_0$, $p_1$, $p_2$, or $p_3$) to move from one region to another. The formulas are structured to reflect realistic navigation tasks that could be encountered in real-world robot navigation scenarios, such as search and rescue, surveillance, or delivery tasks within a specified area. \\ \bottomrule 
    \end{tabular}
    \begin{tablenotes}
        \item[1] \url{https://chat.openai.com/share/43670685-3cb9-4496-ad3e-1ee70cd358b8}.
    \end{tablenotes}
\end{threeparttable}
\end{table*}

\subsection{Additional Illustrations}
\label{appsec:more_experiment_results}

\begin{figure}[t]
    \centering
    \subfigure[Trajectories from~\diffuser.]{
        \includegraphics[width=0.323\columnwidth]{results/maze2d/maze2d_large_unsafe_diffuser1.pdf}
        \includegraphics[width=0.323\columnwidth]{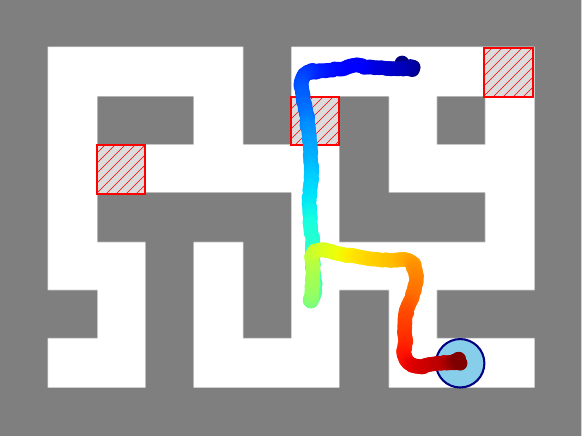}
        \includegraphics[width=0.323\columnwidth]{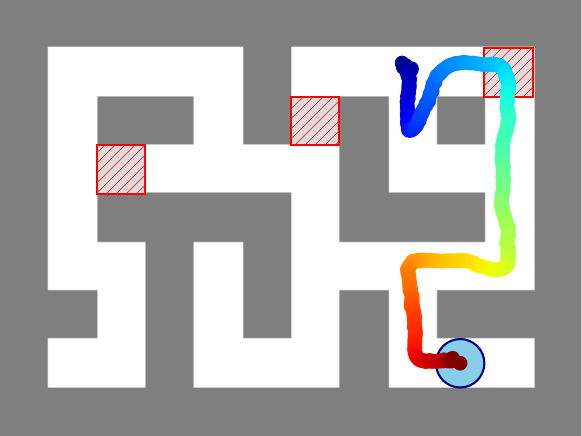}
        \label{fig:result_maze2d_large_unsafe_diffuser_appendix}
    }
    \hfill
    \subfigure[Trajectories from~\safediffuser.]{
        \includegraphics[width=0.33\columnwidth]{results/maze2d/maze2d_large_unsafe_cbf1.pdf}
        \includegraphics[width=0.33\columnwidth]{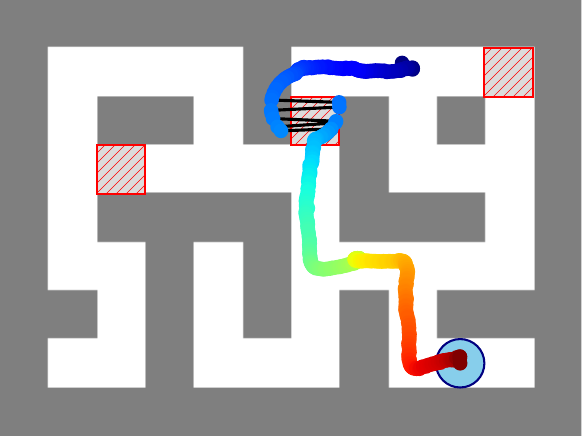}
        \label{fig:result_maze2d_large_unsafe_cbf_appendix}
    }
    \hfill
    \subfigure[Trajectories from~\methodps.]{
        \includegraphics[width=0.33\columnwidth]{results/maze2d/maze2d_large_unsafe_dps1.pdf}
        \includegraphics[width=0.33\columnwidth]{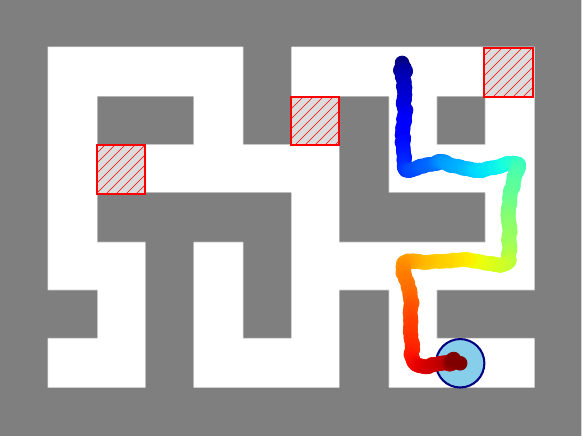}
        \label{fig:result_maze2d_large_unsafe_dps_appendix}
    }
    \hfill
    \subfigure[Trajectories from~\methodrg.]{
        \includegraphics[width=0.33\columnwidth]{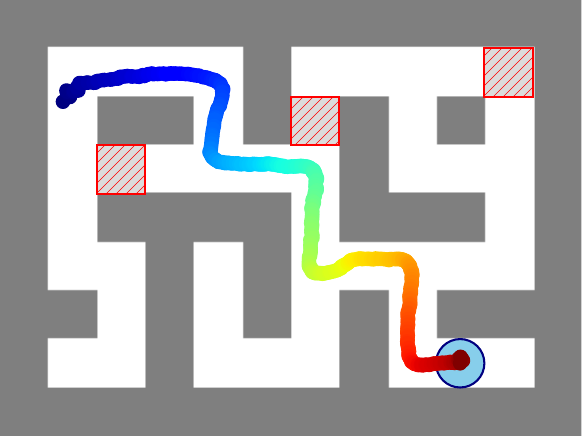}
        \includegraphics[width=0.33\columnwidth]{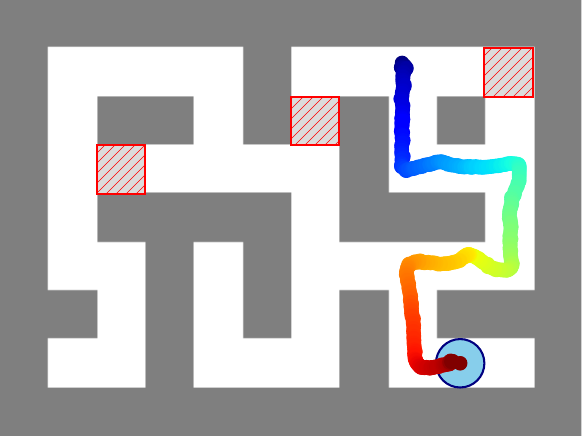}
        \label{fig:result_maze2d_large_unsafe_rg_appendix}
    }
    \caption{Results of safe planning in Maze2d-Large with unsafe regions. There are three unsafe blocks (red squares, labeled $p_L$, $p_M$, $p_R$ from left to right) that need to be avoided during navigation to the goal (shaded circle). The $\ltlf$ constraint for this task is $\varphi = \ltlalways\neg(p_L \wedge p_M \wedge p_R)$. (a) Trajectories from \diffuser ignore safety and can violate the specified constraints. (b) \safediffuser{} produces discontinuous trajectories (the constraints here are slightly reshaped to accommodate the CBF function). (c, d) Our \method{} is able to plan trajectories that detours around the obstacles to successfully arrive at the goal.}
    \label{fig:result_maze2d_large_unsafe_appendix}
\end{figure}

More illustrations and comparisons against the baseline for the obstacle avoidance task in Maze2d environment are in \figref{fig:result_maze2d_large_unsafe_appendix}. 

\begin{table}[h]
    \centering
    \caption{Results of generated $\ltlf$s in simulation rooms.}
    \label{table:result_rls_ltl}
    \renewcommand{\tabcolsep}{3pt}
    \begin{tabular}{llcccc} \toprule
        \multirow{2}{*}{\textbf{Task}} & \multirow{2}{*}{\textbf{Method\textbackslash Performance}} && \multicolumn{2}{c}{Success rate ($\%$) $\uparrow$} \\ \cmidrule{4-5}
        &&& Lab & Office \\ \midrule
        \multirow{2}{*}{Obstacle avoidance} & \diffuser     && 11.7$\pm$3.7     & 0.0$\pm$0.0 \\
                                            & \methodps     && 100.0$\pm$0.0    & 80.0$\pm$8.1 \\ \midrule
        \multirow{2}{*}{Temporal behaviors} & \diffuser     && 0.0$\pm$0.0      & 0.0$\pm$0.0 \\ 
                                            & \methodps     && 96.5$\pm$1.5     & 66.7$\pm$1.8 \\ \bottomrule 
    \end{tabular}
\end{table}

\begin{figure}[h]
    \centering
    \subfigure[Planning.]{
        \includegraphics[width=.293\columnwidth]{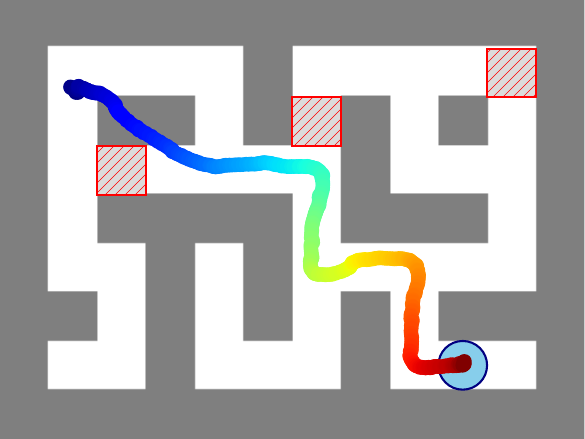}
        \label{fig:result_maze2d_large_unsafe_failure_plan}
    }
    \hfil
    \subfigure[Rollout C1.]{
        \includegraphics[width=.293\columnwidth]{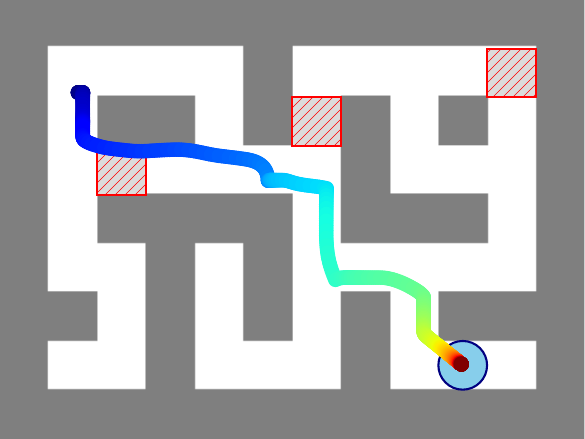}
        \label{fig:result_maze2d_large_unsafe_failure_rollout_c1}
    }
    \hfil
    \subfigure[Rollout C2.]{
        \includegraphics[width=.293\columnwidth]{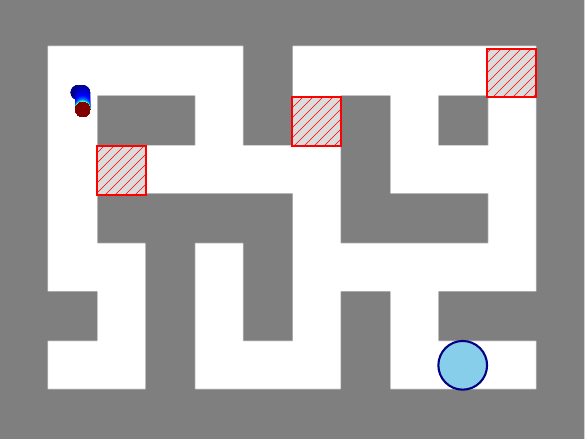}
        \label{fig:result_maze2d_large_unsafe_failure_rollout_c2}
    }
    \caption{Example of failure cases from~\methodps.}
    \label{fig:result_maze2d_large_unsafe_failure}
\end{figure}

\begin{table}[h]
    \centering
    \begin{threeparttable}
    \caption{Comparison of Controllers\tnote{1} in Avoidance Tasks in Maze2d}
    \label{table:maze2d_large_unsafe_apdx}
    \renewcommand{\tabcolsep}{2pt}
    \begin{tabular}{lccccccc} \toprule
        \multirow{3}{*}{\textbf{Method\textbackslash Perf.}} && \multicolumn{3}{c}{Success rate ($\%$) $\uparrow$} & \multicolumn{2}{c}{Reward (UnCon) $\uparrow$} \\ \cmidrule{3-7}
        && \multirow{2}{*}{Planning} & \multicolumn{2}{c}{Rollout} & \multirow{2}{*}{C1} & \multirow{2}{*}{C2} \\ \cmidrule{4-5}
        &&& C1 & C2 \\ \midrule
        \diffuser         && 9.5$\pm$3.1  & 11.0$\pm$1.7      & 15.0$\pm$2.6      & 142.2$\pm$5.1 & 92.1$\pm$3.0 \\
        \safediffuser     && 99.4$\pm$0.9 & 12.3$\pm$3.5      & 14.2$\pm$2.8      & 135.8$\pm$5.2 & 131.9$\pm$5.4 \\
        \methodps         && 99.0$\pm$0.8 & \bf{73.0$\pm$3.0} & \bf{90.6$\pm$0.6} & 97.3$\pm$2.8  & 36.2$\pm$11.7 \\
        \methodrg         && 98.8$\pm$0.8 & \bf{92.0$\pm$1.4} & \bf{99.6$\pm$0.6} & 127.1$\pm$5.1 & 76.4$\pm$13.4 \\ \bottomrule
    \end{tabular}
    \begin{tablenotes}
        \item[1] Planned trajectories are rolled out using two different low level control strategies, denoted as C1 and C2.
    \end{tablenotes}
    \end{threeparttable}
\end{table}

\begin{figure*}[t]
    \centering
    \subfigure[Regions.]{
        \includegraphics[width=.233\textwidth]{figures/pusht_regions.png}
        \label{fig:illustration_pusht_regions_app}
    }
    \hfil
    \subfigure[\diffusionpolicy.]{
        \includegraphics[width=.231\textwidth]{figures/pusht_dp_ef_oa.png}
        \label{fig:result_pusht_diffusionpolicy_app}
    }
    \hfil
    \subfigure[Ours (\methodps).]{
        \includegraphics[width=.231\textwidth]{figures/pusht_ps_ef_oa.png}
        \label{fig:result_pusht_dps_app}
    }
    \hfil
    \subfigure[Ours (\methodrg).]{
        \includegraphics[width=.231\textwidth]{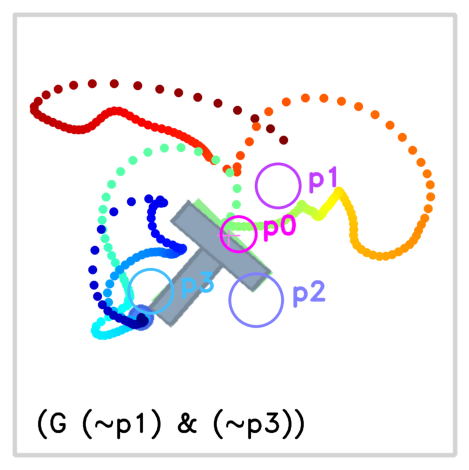}
        \label{fig:result_pusht_rg_app}
    }
    \caption{PushT task. Diffusion policy generates next locations of the mover to push the \texttt{T} block by following these locations. 
    Atomic propositions for LTLs are defined by constraints over actions.}
    \label{fig:illustration_pusht_app}
\end{figure*}

\begin{figure*}[t]
    \centering
    \subfigure[\methodps]{
        \includegraphics[width=.19\textwidth]{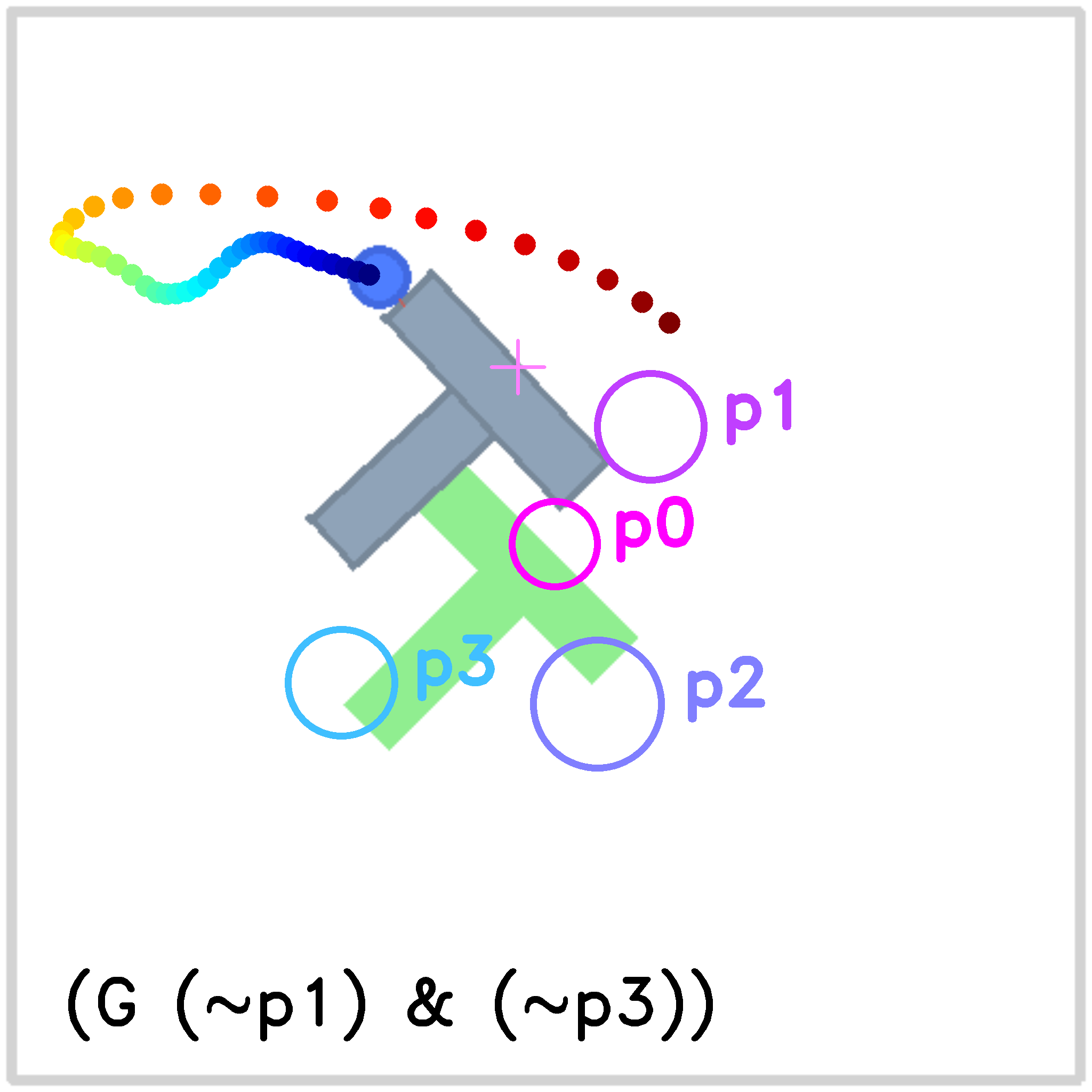}
        \includegraphics[width=.19\textwidth]{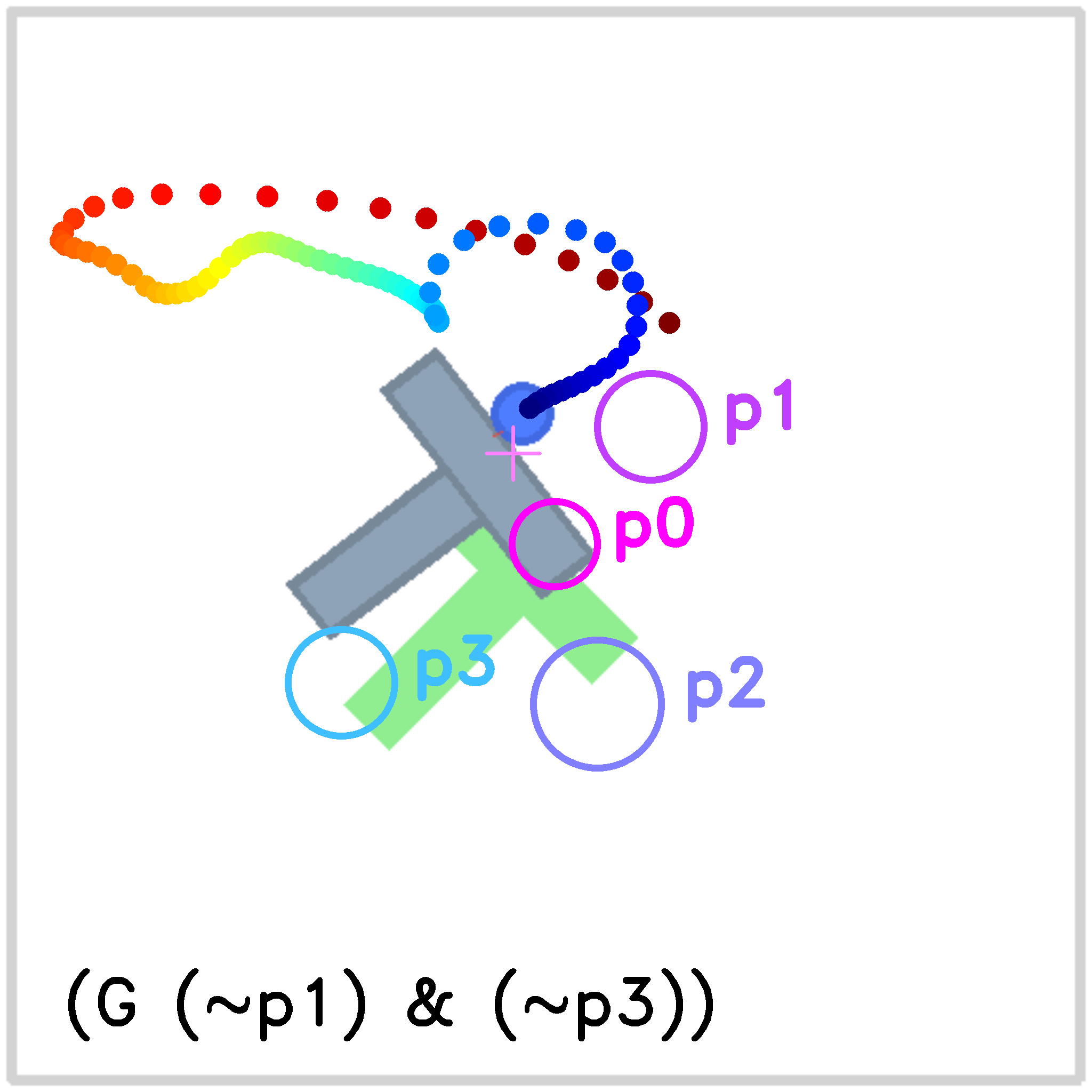}
        \includegraphics[width=.19\textwidth]{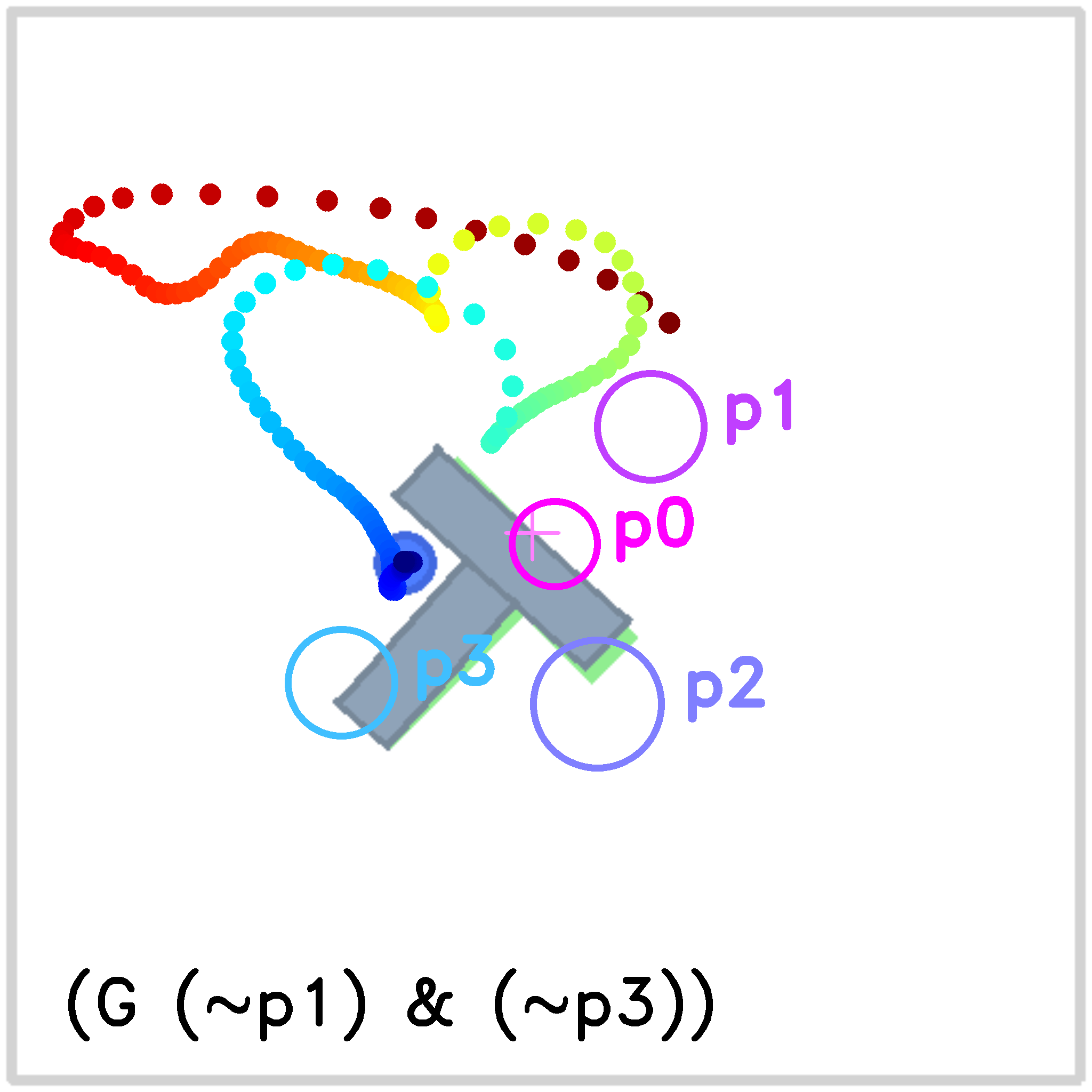}
        \includegraphics[width=.19\textwidth]{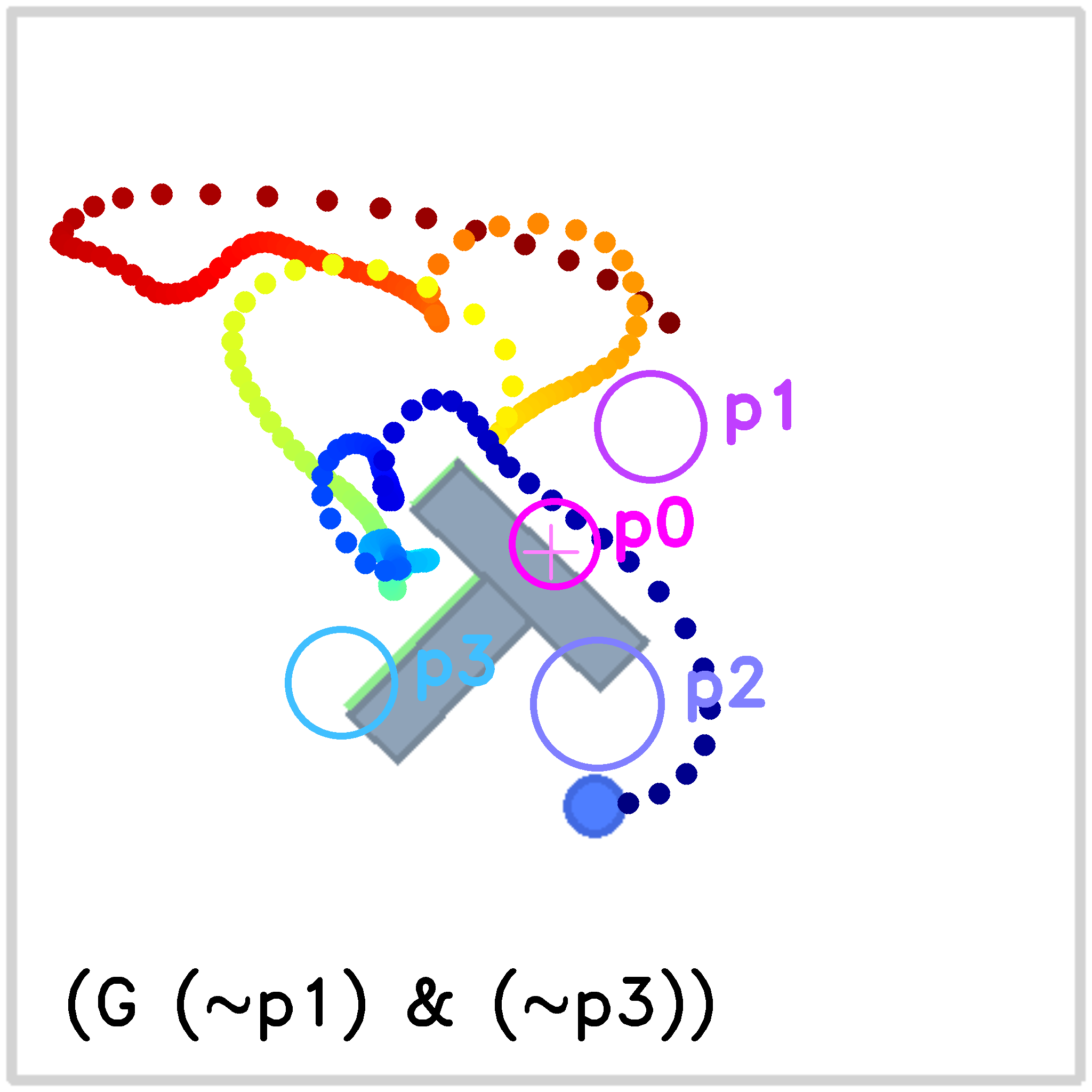}
        \includegraphics[width=.19\textwidth]{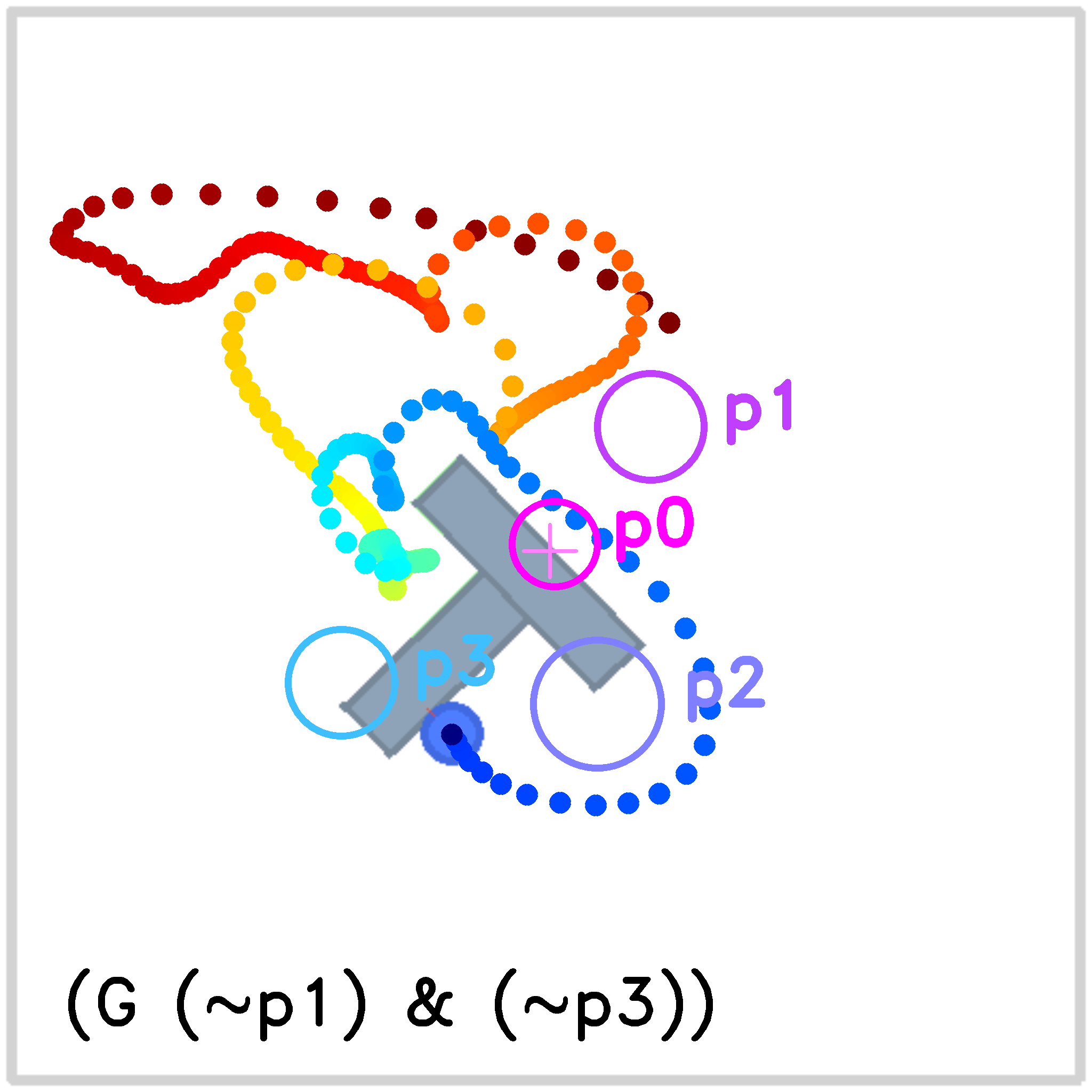}
        \label{fig:ps_frames_ltl8}
    }
    \subfigure[\methodrg]{
        \includegraphics[width=.19\textwidth]{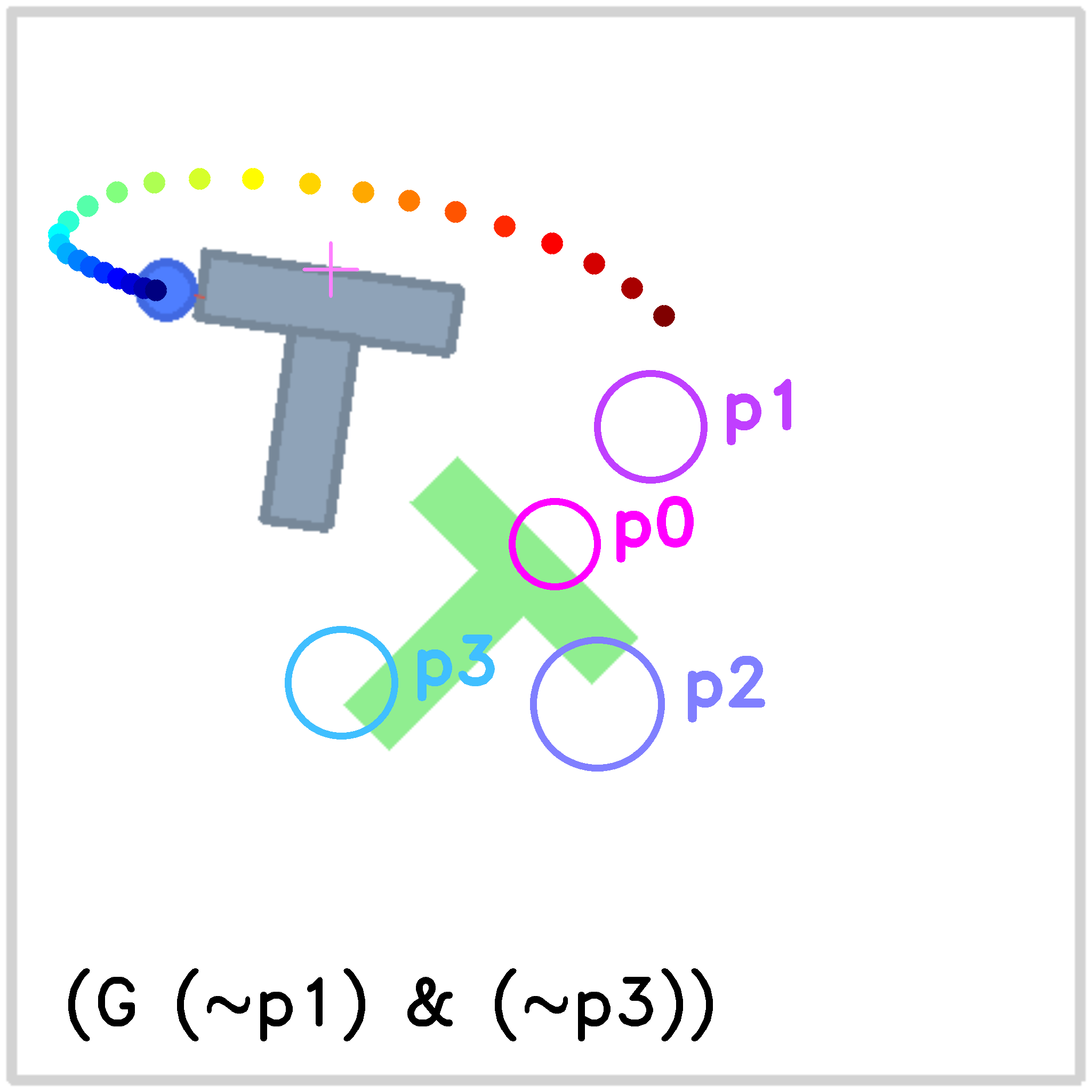}
        \includegraphics[width=.19\textwidth]{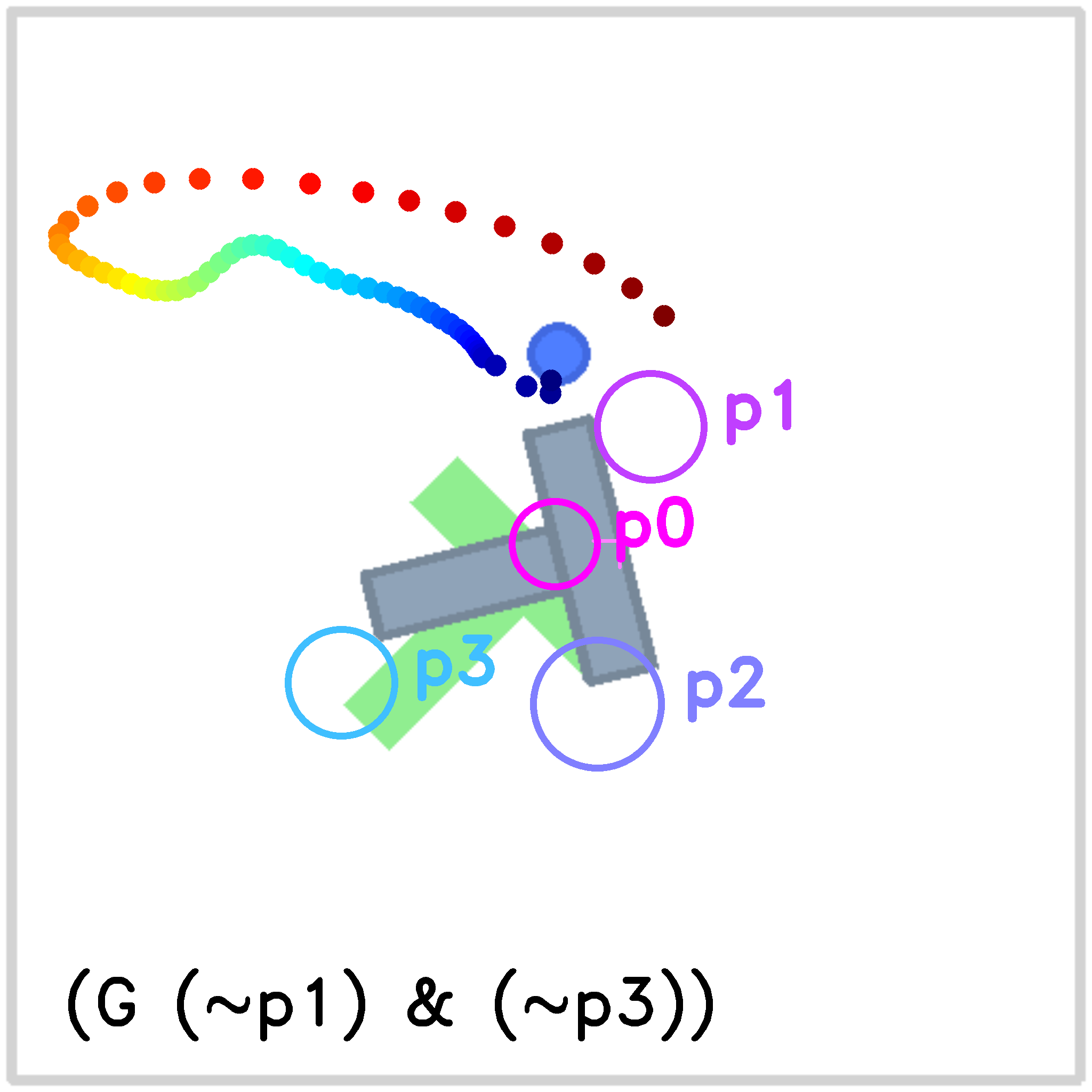}
        \includegraphics[width=.19\textwidth]{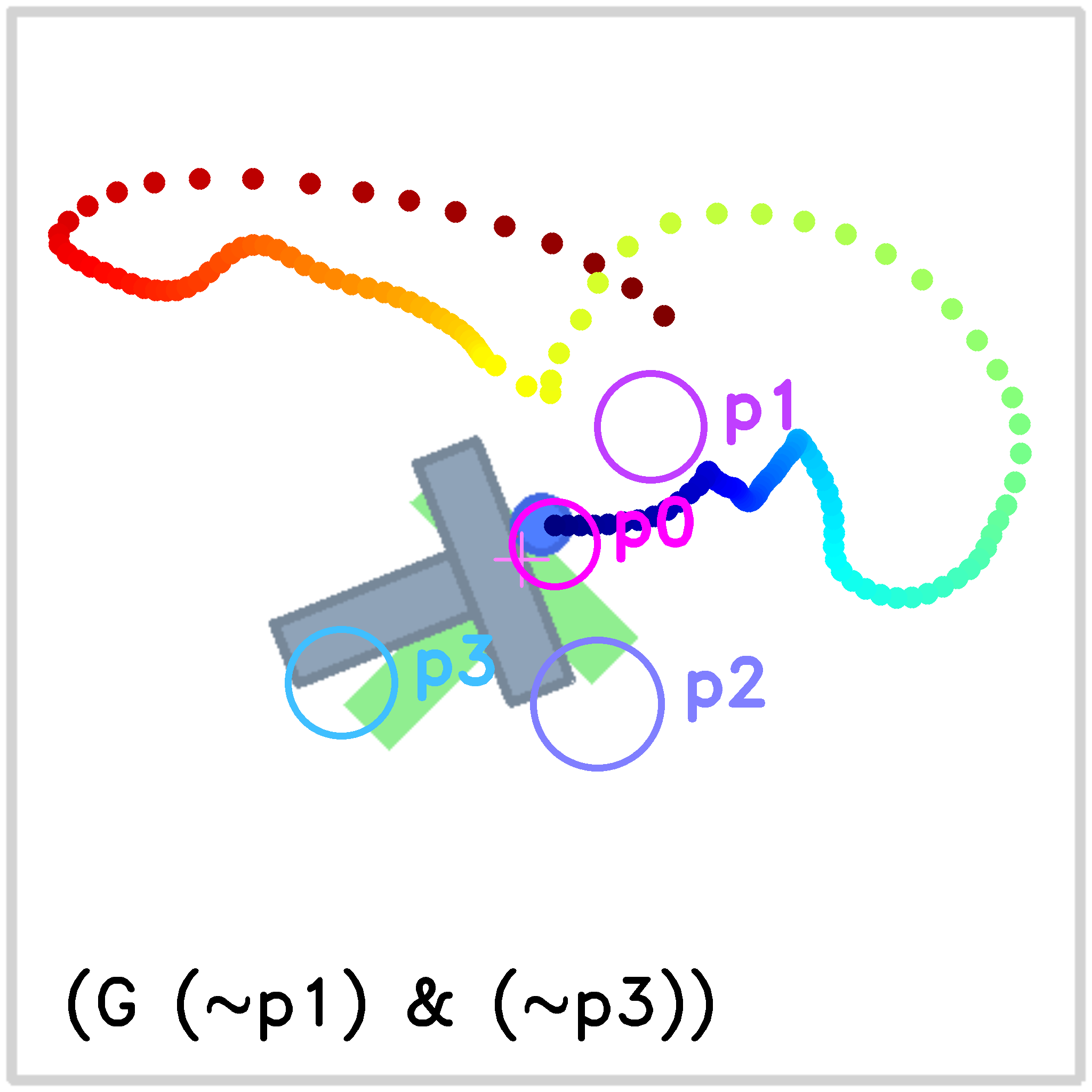}
        \includegraphics[width=.19\textwidth]{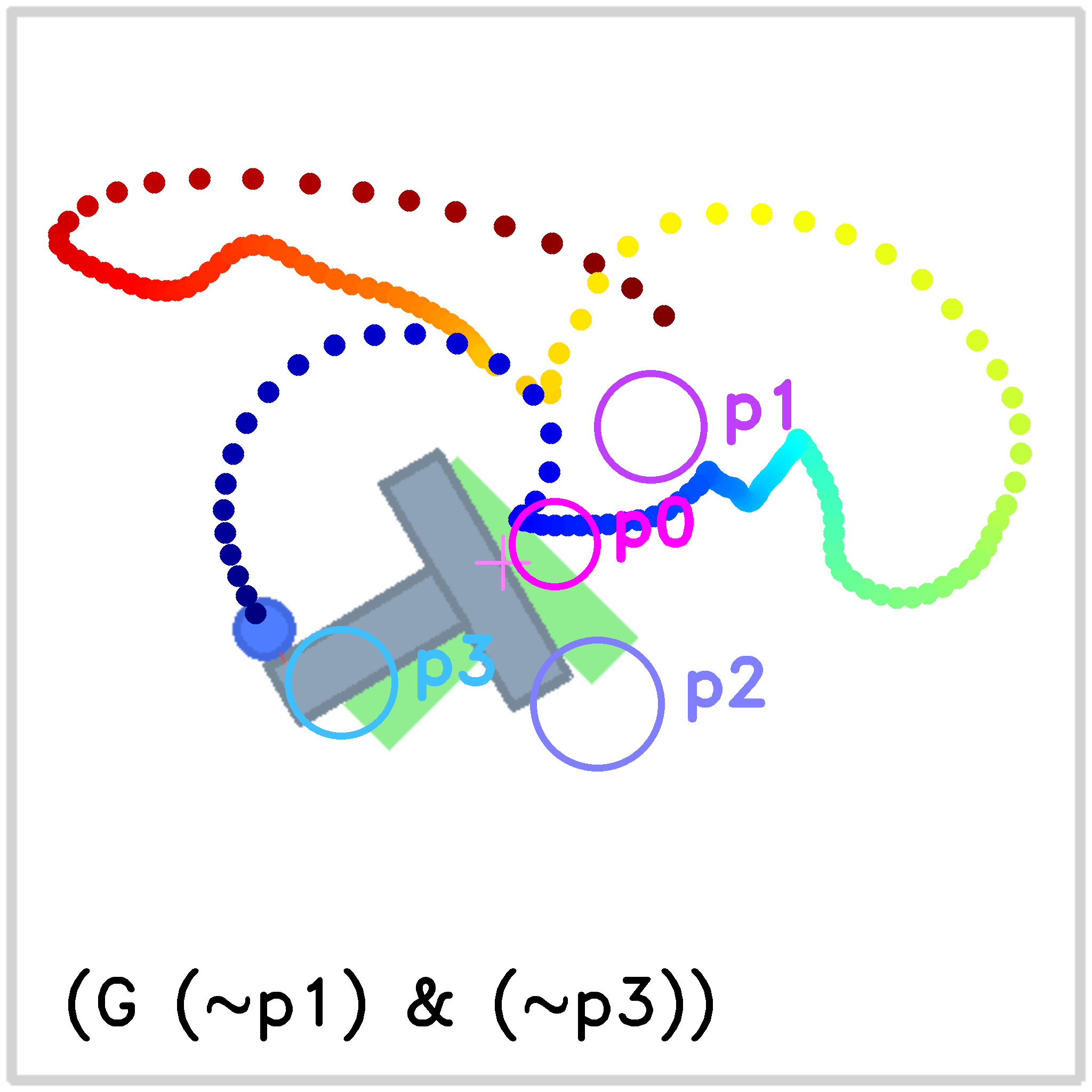}
        \includegraphics[width=.19\textwidth]{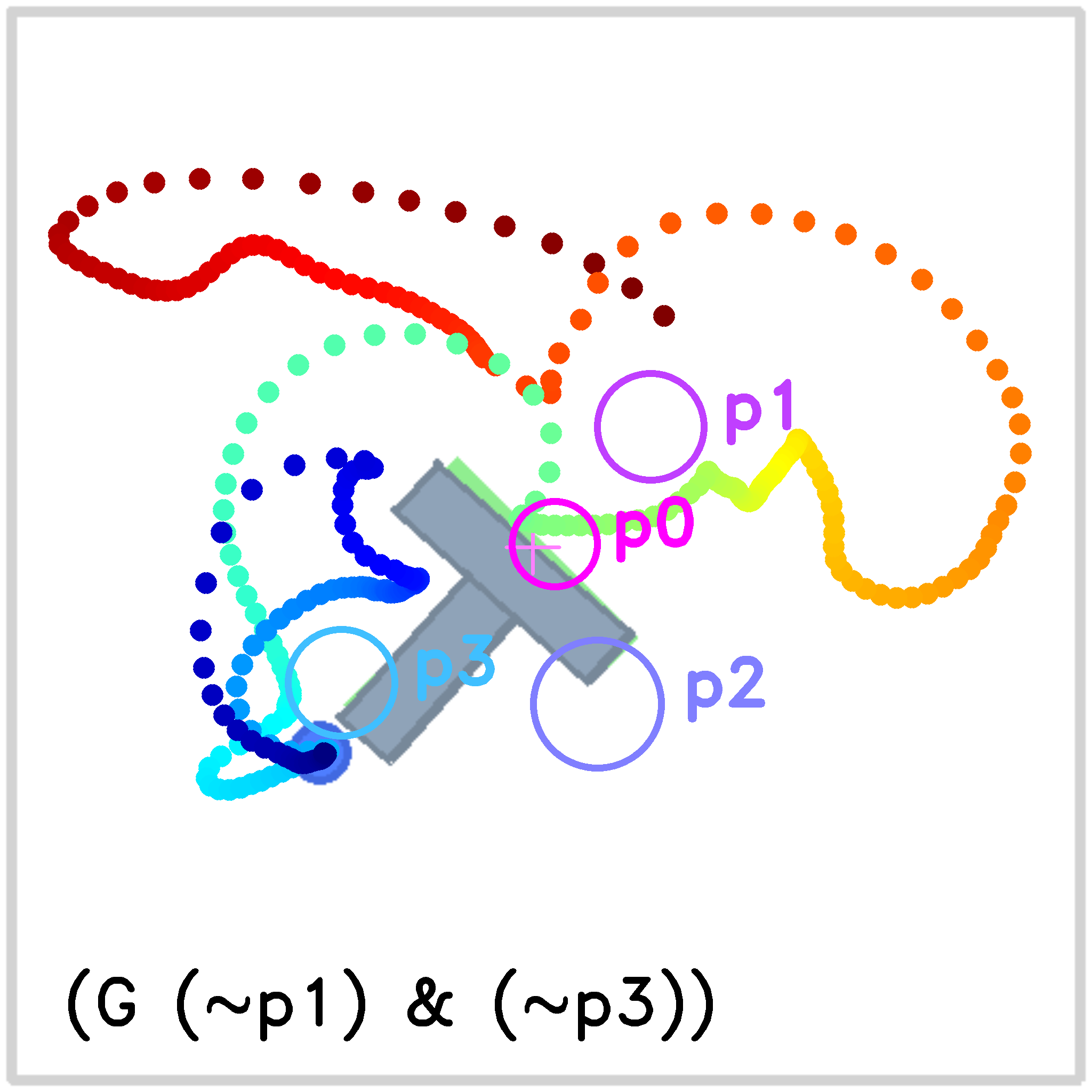}
        \label{fig:rg_frames_ltl8}
    }
    \caption{
        Snapshots of two successful rollouts of the two variants of our method accomplishing the same task in \figref{fig:illustration_pusht_app}. 
        The mover (circle filled in blue) should avoid region $p1$ and $p3$ at all time, as designated in the $\ltlf$ ``$\ltlalways (\neg p_1 \wedge \neg p_3)$.''
        \figref{fig:ps_frames_ltl8} and \figref{fig:rg_frames_ltl8} showcase that \methodps and \methodrg exploit different modalities to complete the original manipulation task while respecting the constraint specified by the $\ltlf$. 
    }
    \label{fig:pusht_frames_ef_oa_ltl8_app}
\end{figure*}

\begin{figure*}[t]
    \centering
    \subfigure[\diffusionpolicy (baseline)]{
        \includegraphics[width=.19\textwidth]{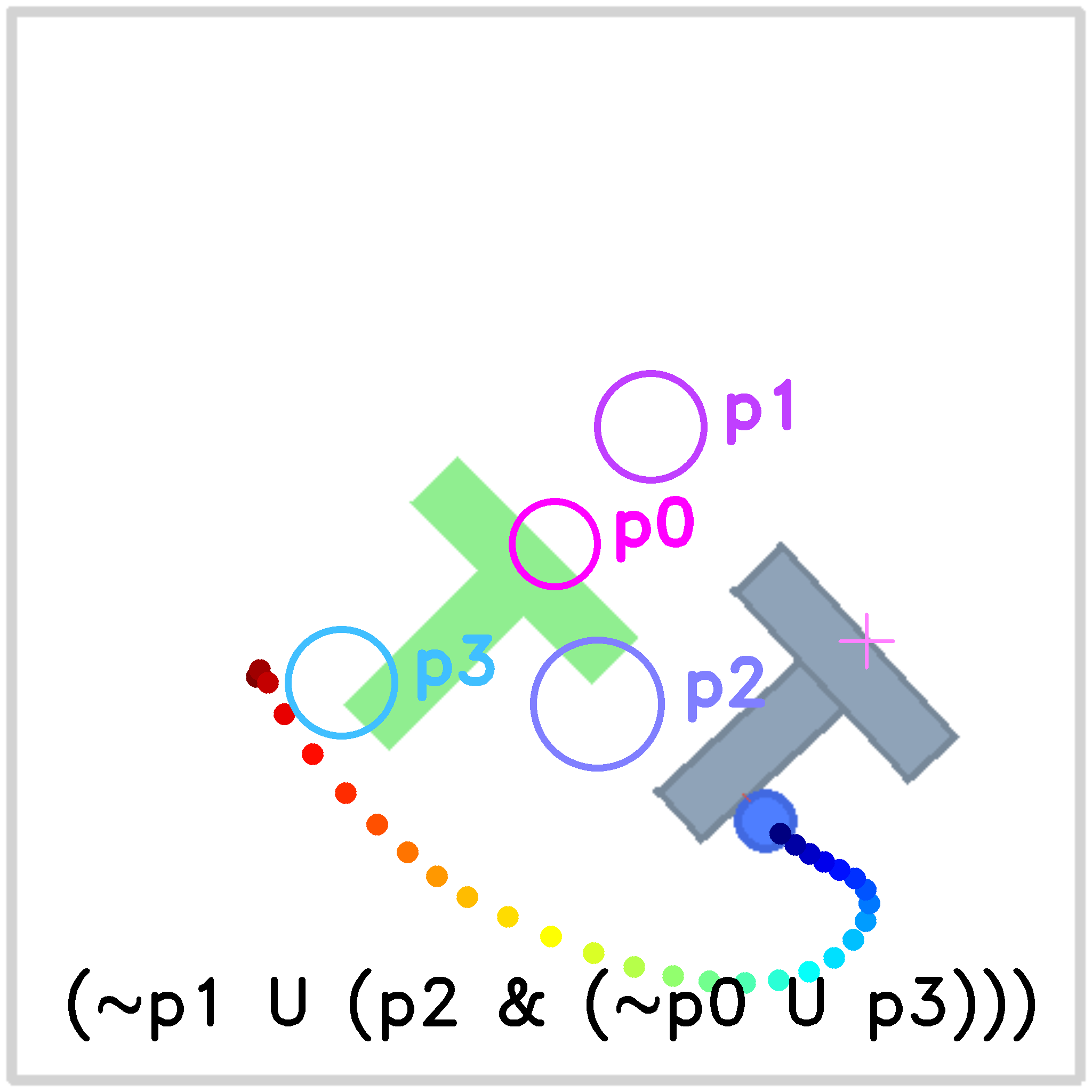}
        \includegraphics[width=.19\textwidth]{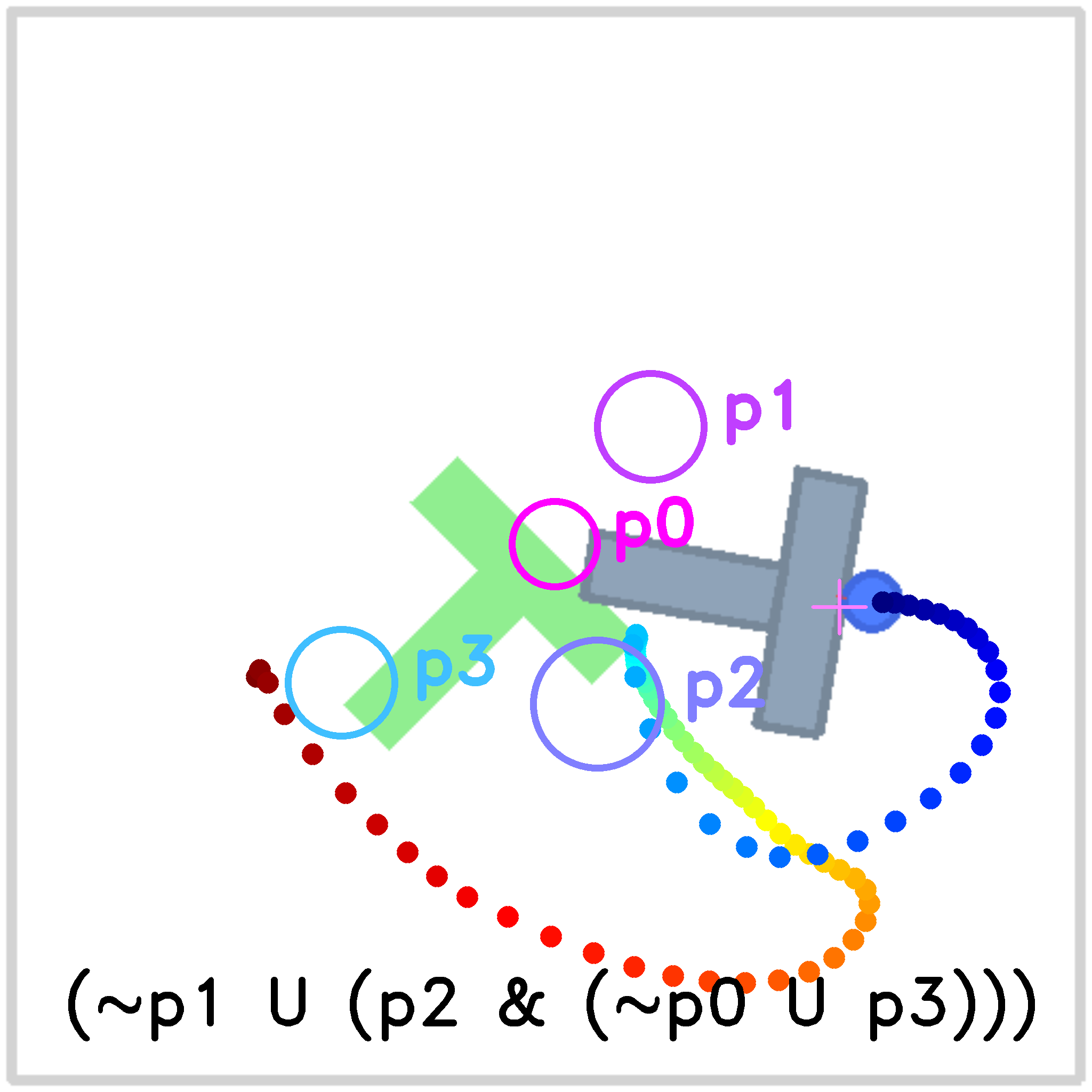}
        \includegraphics[width=.19\textwidth]{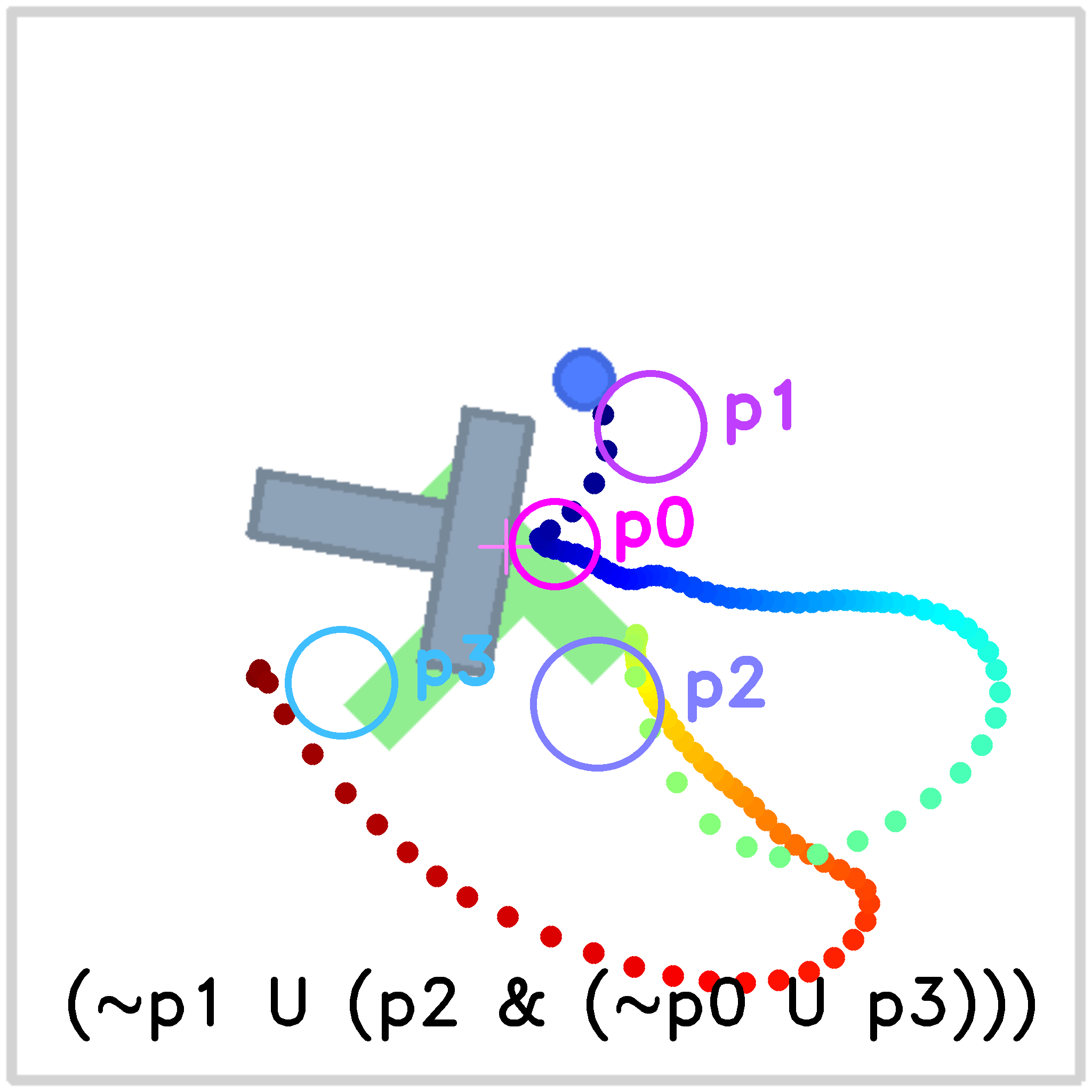}
        \includegraphics[width=.19\textwidth]{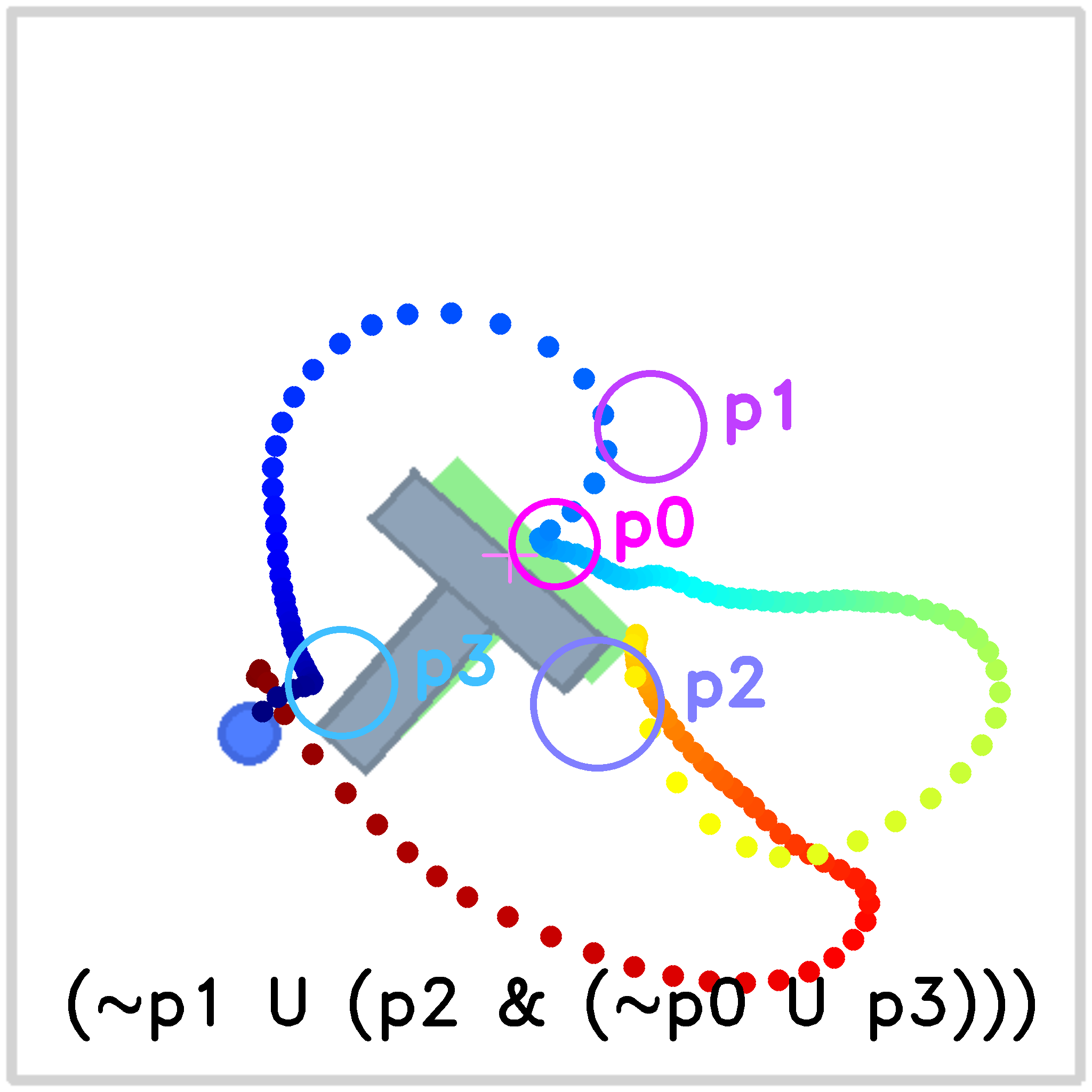}
        \includegraphics[width=.19\textwidth]{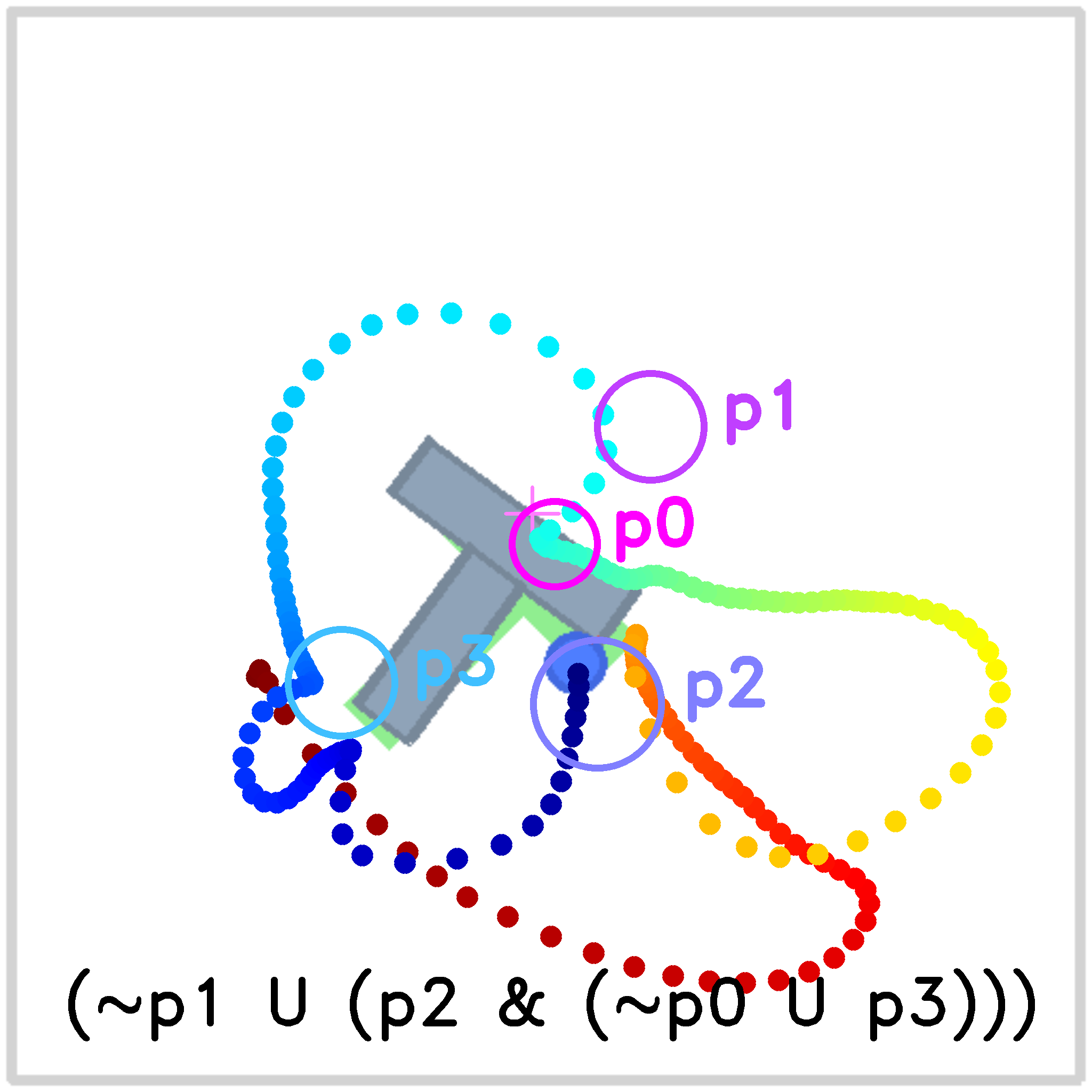}
        \label{fig:baseline_frames_ltl35}
    }
    \subfigure[\methodps (ours)]{
        \includegraphics[width=.19\textwidth]{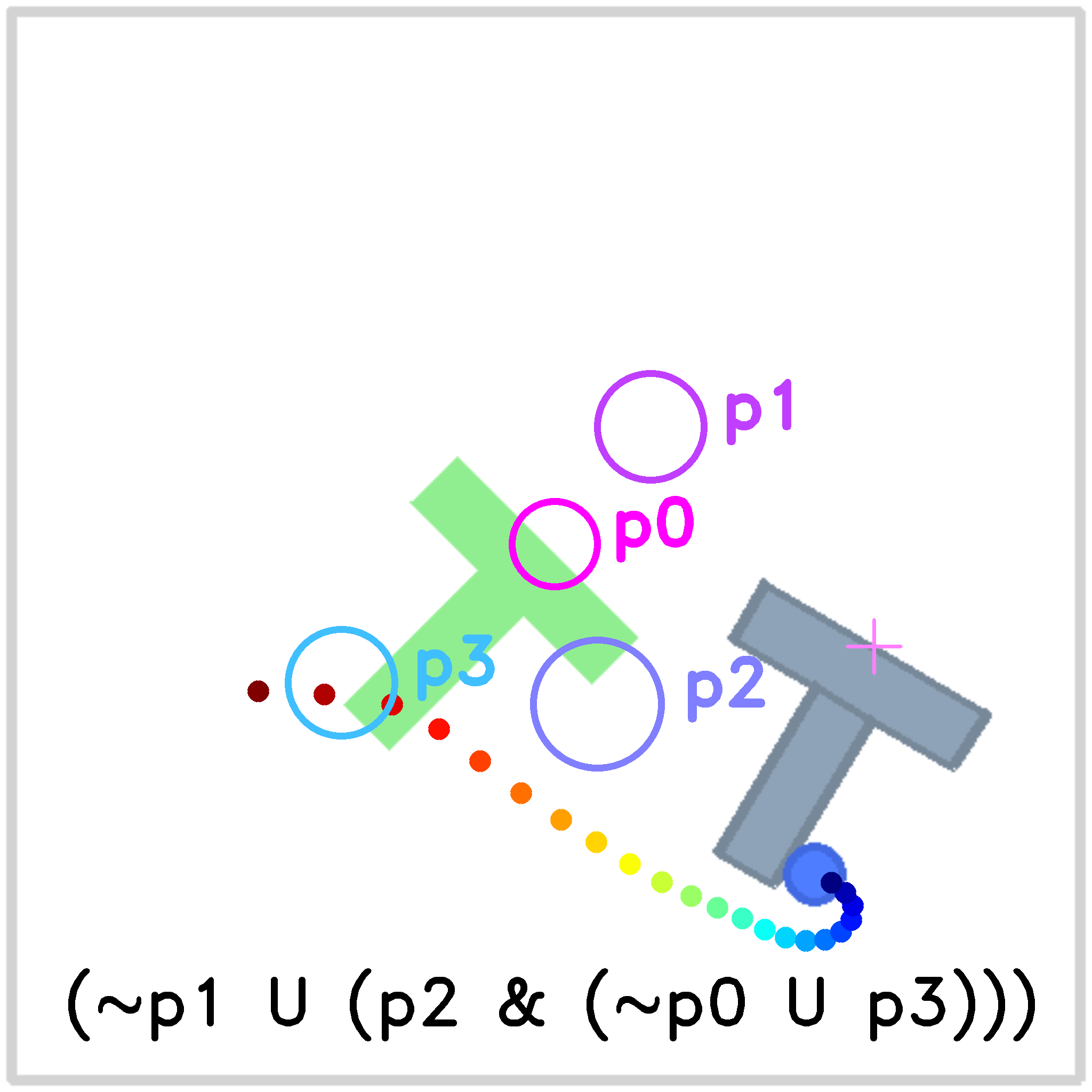}
        \includegraphics[width=.19\textwidth]{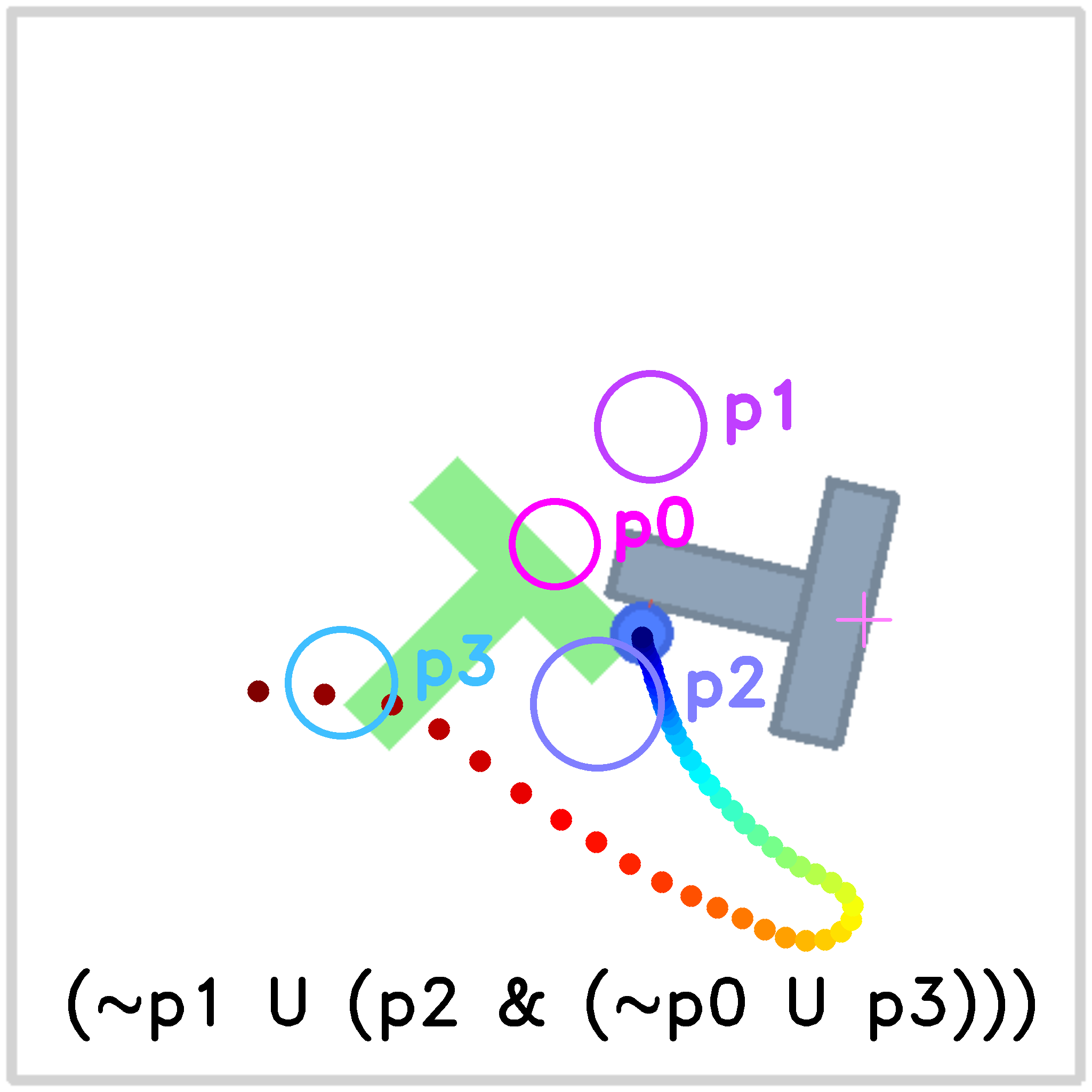}
        \includegraphics[width=.19\textwidth]{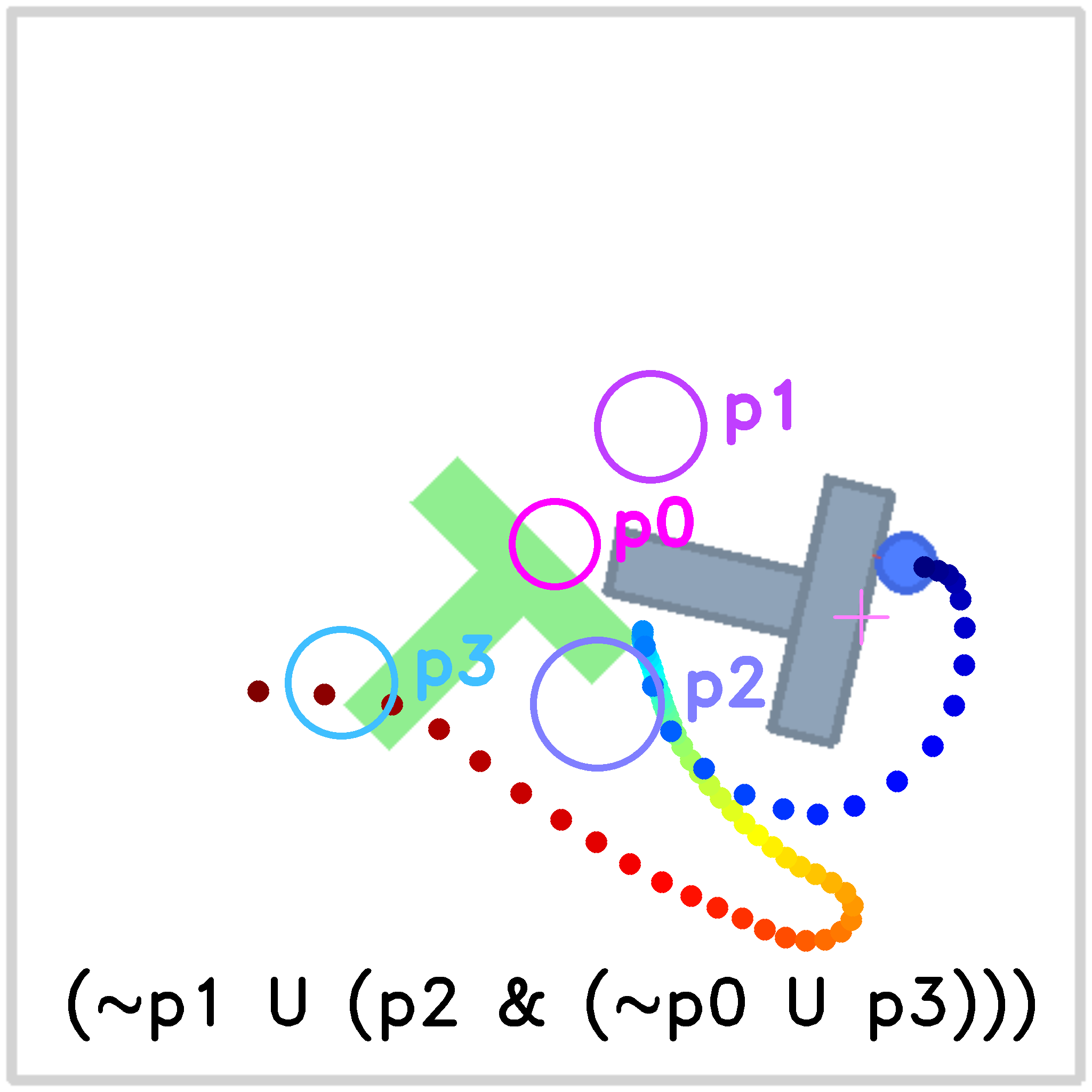}
        \includegraphics[width=.19\textwidth]{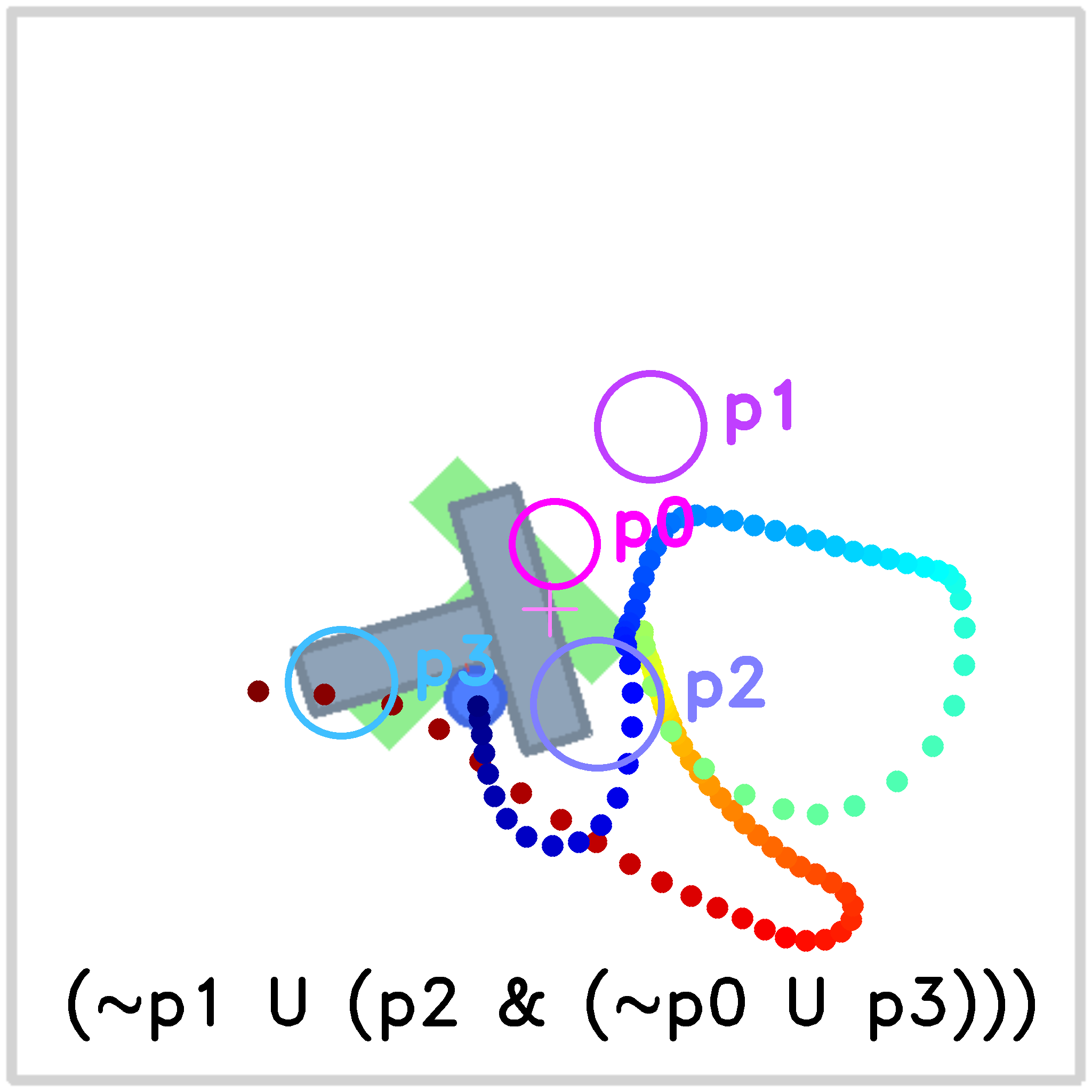}
        \includegraphics[width=.19\textwidth]{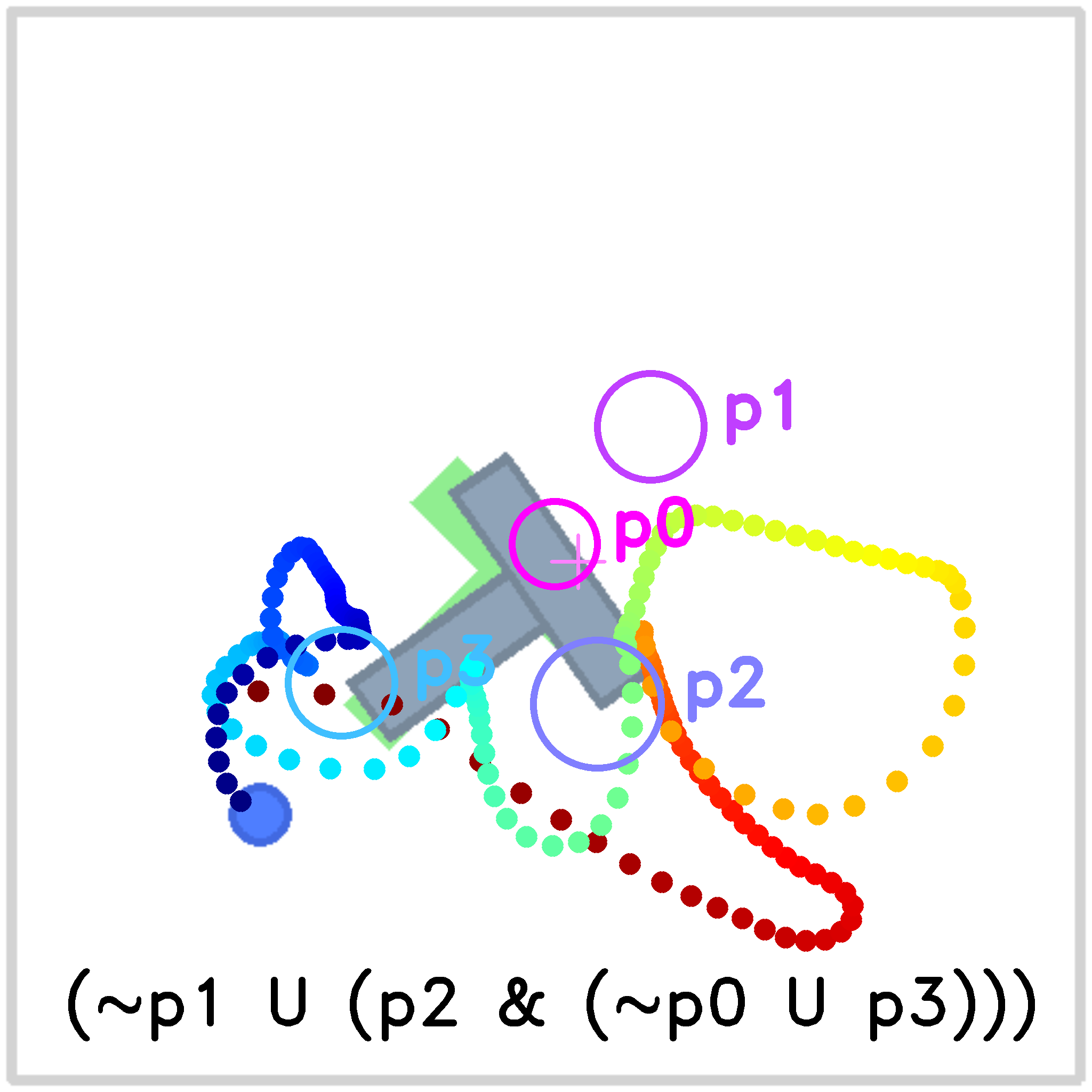}
        \label{fig:ps_frames_ltl35}
    }
    \subfigure[\methodrg (ours)]{
        \includegraphics[width=.19\textwidth]{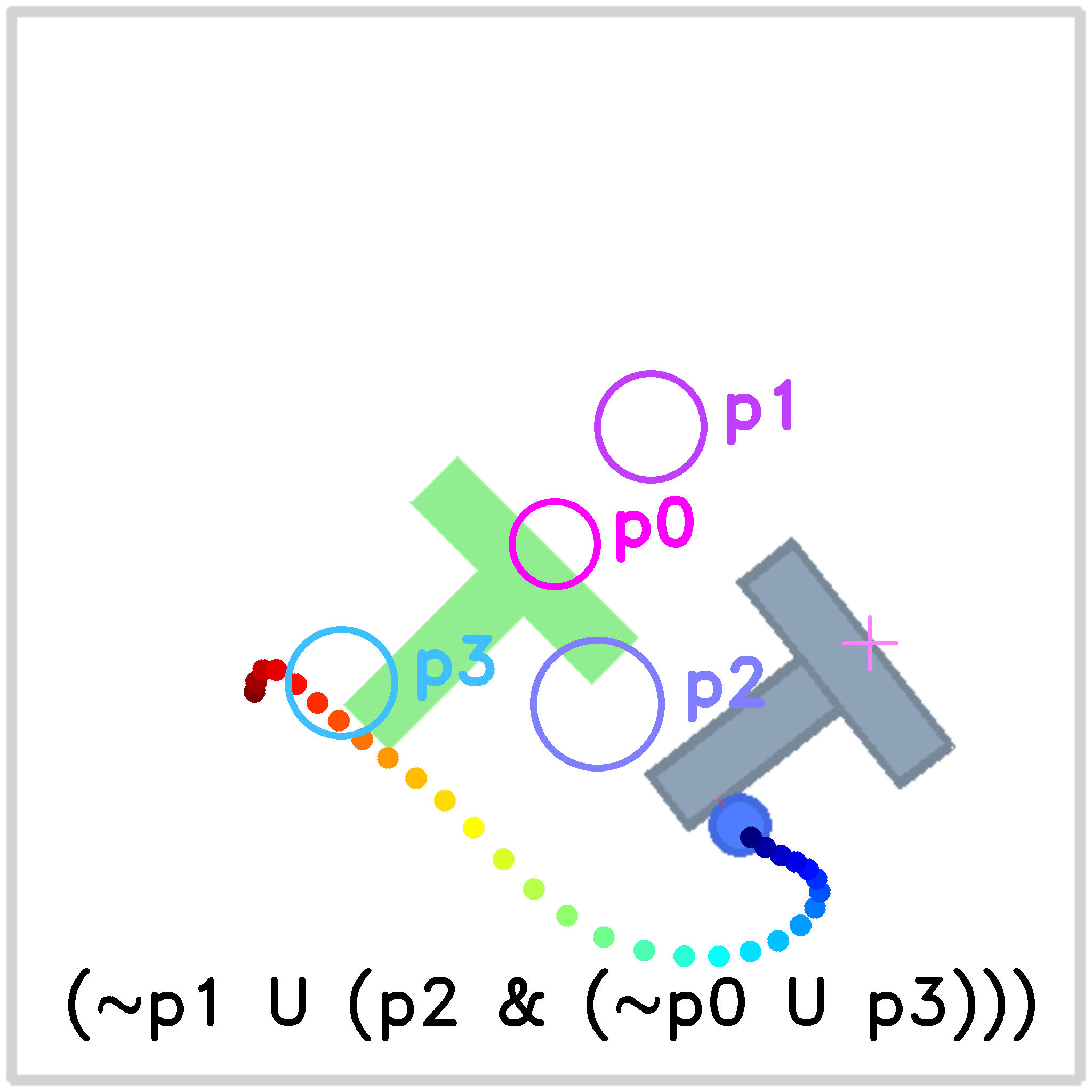}
        \includegraphics[width=.19\textwidth]{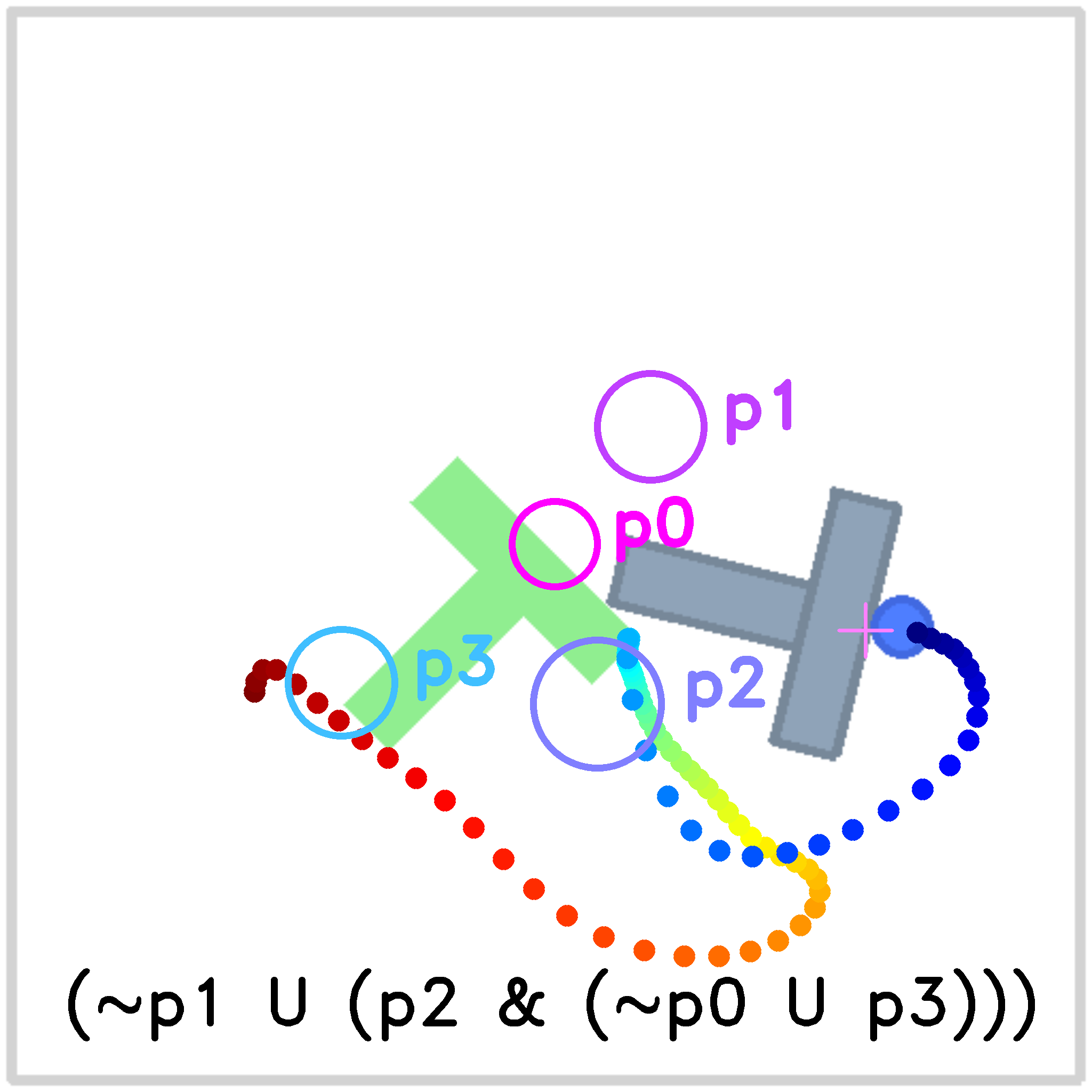}
        \includegraphics[width=.19\textwidth]{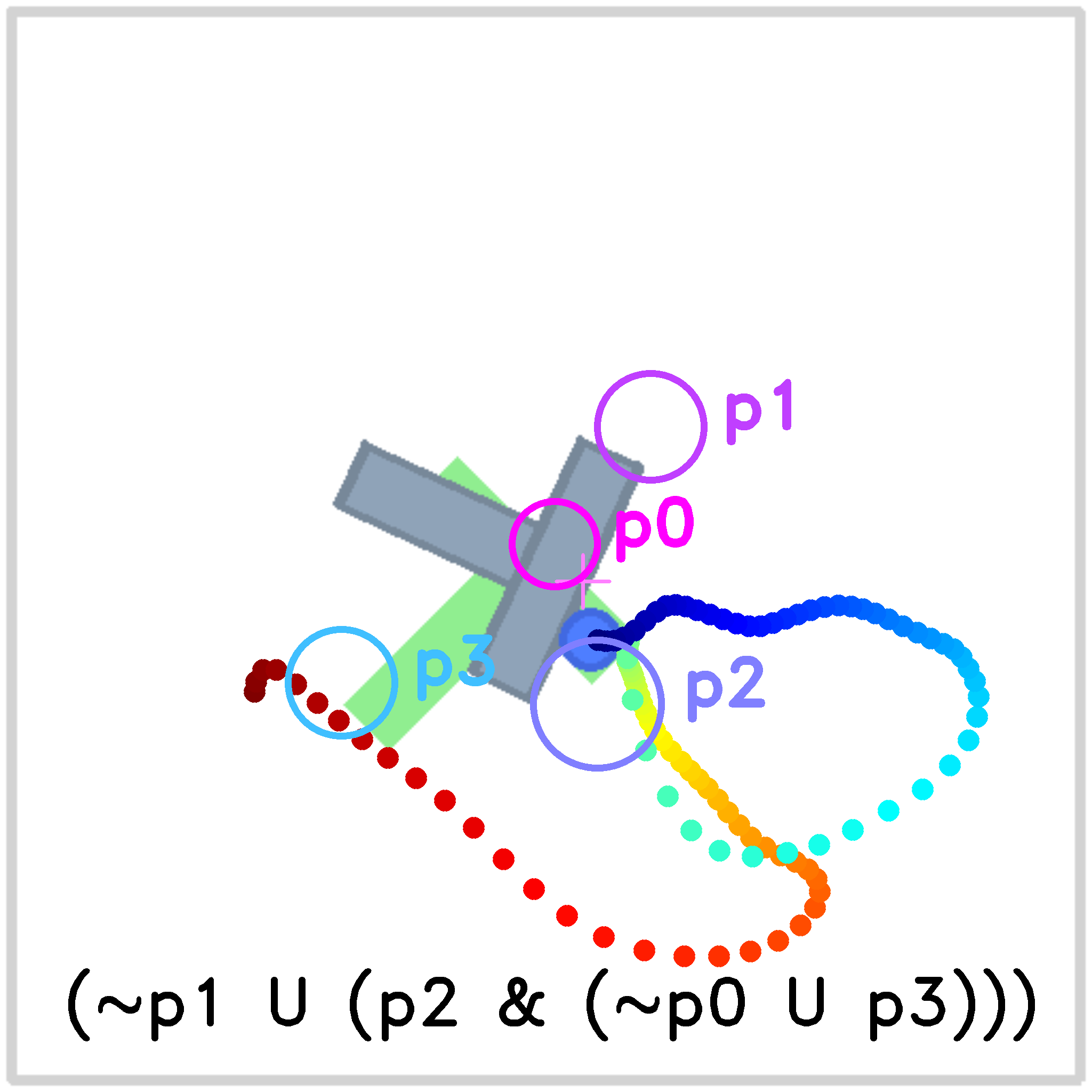}
        \includegraphics[width=.19\textwidth]{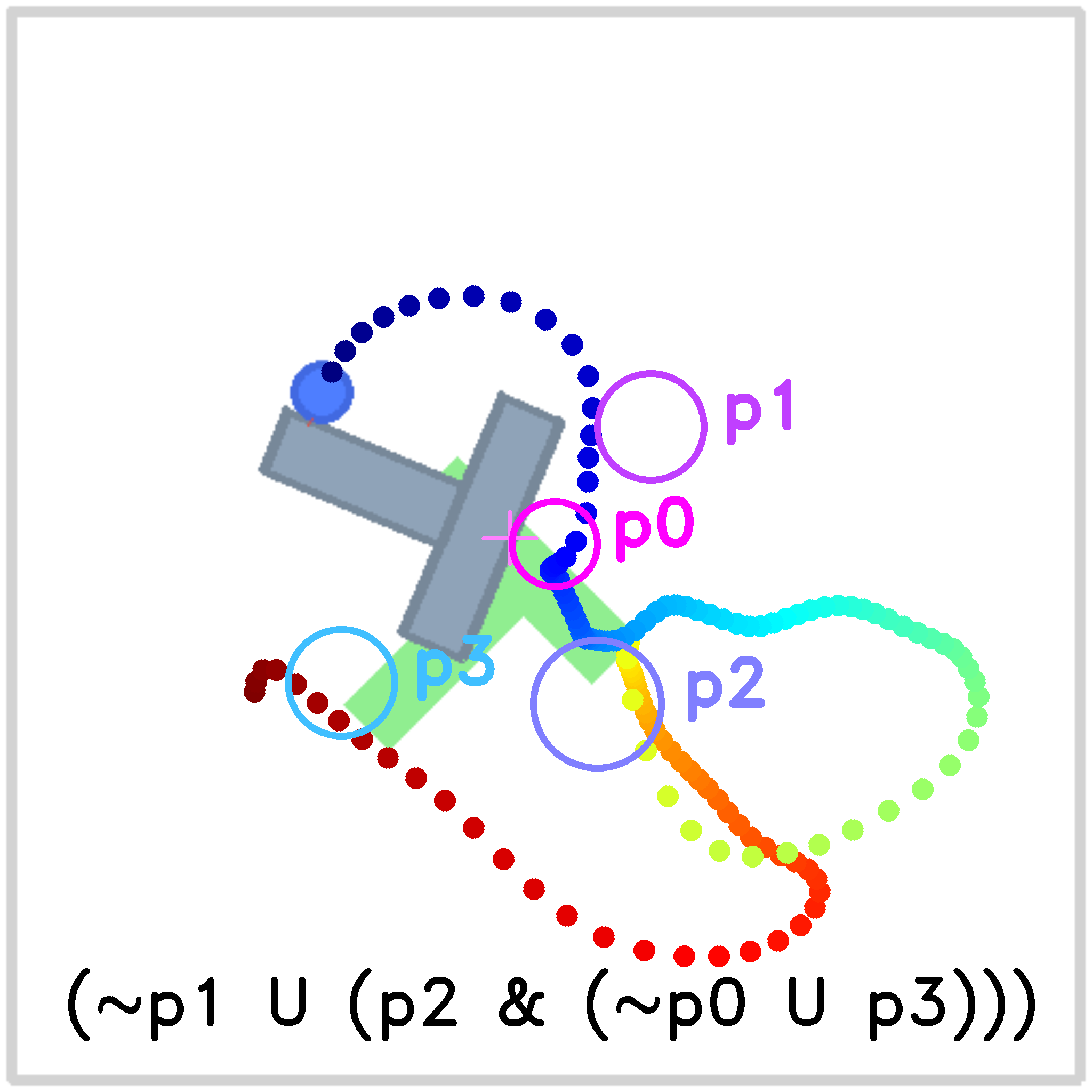}
        \includegraphics[width=.19\textwidth]{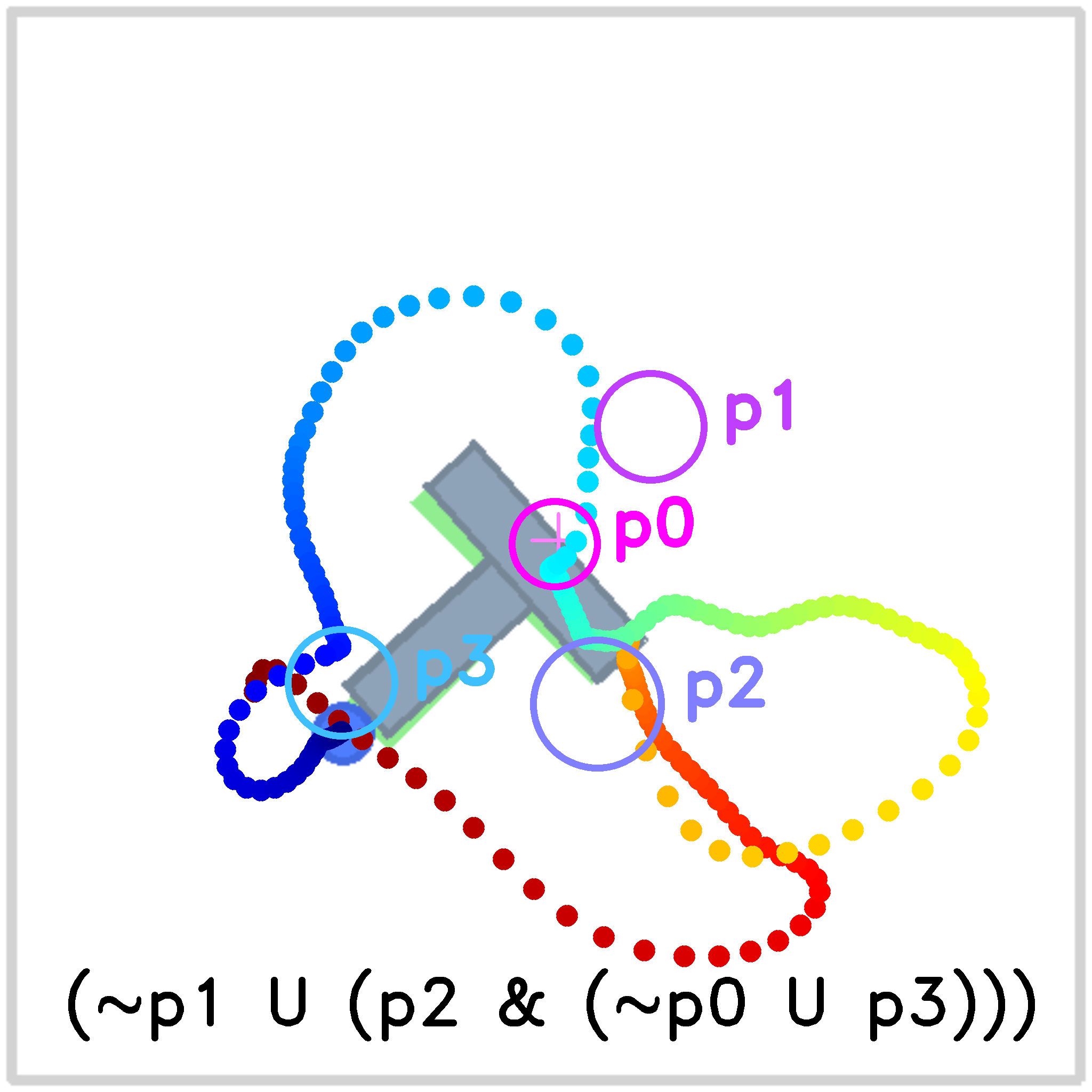}
        \label{fig:rg_frames_ltl35}
    }
    \caption{
        Snapshots of comparisons between the baseline and \method given an $\ltlf$ with intricate \emph{temporal} constraints. 
        The $\ltlf$ ``$\neg p_1 \ltluntil ( p_2 \wedge ( \neg p_0 \ltluntil p_3 ) )$'' demands the mover never enter $p1$ until (or, unless by the time of entering) the two conditions are both met: 1) it has already been in $p2$, and 2) $p0$ is never reached until $p3$ is visited. 
        Clearly, the baseline breaches this complex constraint when proceeding with the original manipulation task, shown from the third to the fifth snapshot (the mover enters $p0$ \emph{without} reaching $p3$ beforehand and then gets into $p0$). 
        In contrast, our method, both \methodps and \methodrg, manage to satisfy the constraint and complete the manipulation task by two different ways (the common ground is to first visit $p3$ and $p2$ during or on the way to maneuvers). 
    }
    \label{fig:pusht_frames_ef_tp_ltl35_app}
\end{figure*}

\figref{fig:illustration_pusht_app}--\figref{fig:pusht_frames_ef_tp_ltl35_app} delineate two more examples of the comparisons between our method \method against the baseline \diffusionpolicy on the obstacle avoidance and temporal behavior tasks. 
These results demonstrate our method's capabilities in guiding diffusion models to follow not only \emph{purely spatial} but also \emph{spatial-temporal} constraints. 

\begin{figure*}
    \centering
    \subfigure[Frames of trajectories from~\diffuser in simulation.]{
        \includegraphics[width=0.19\textwidth]{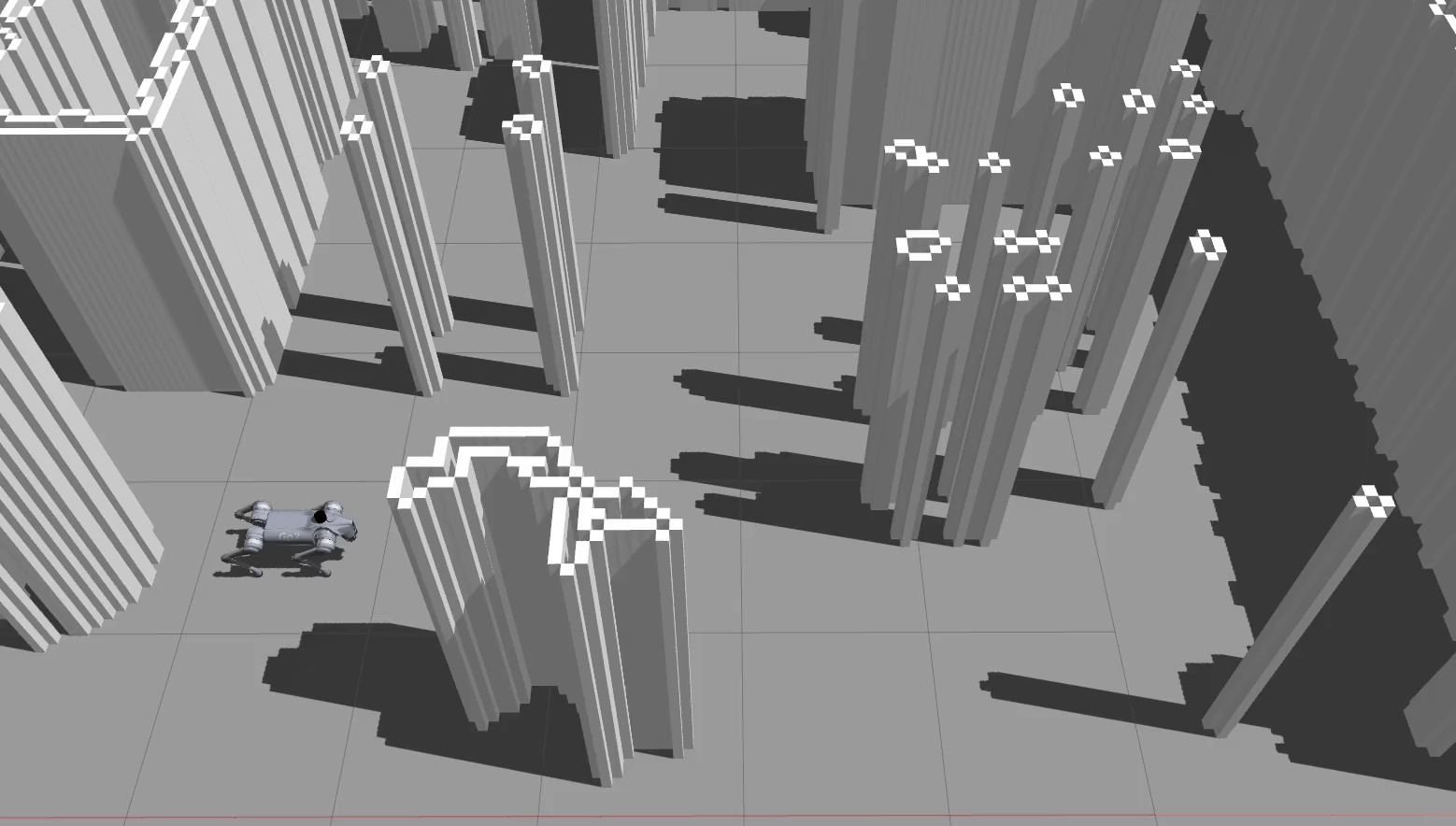}
        \includegraphics[width=0.19\textwidth]{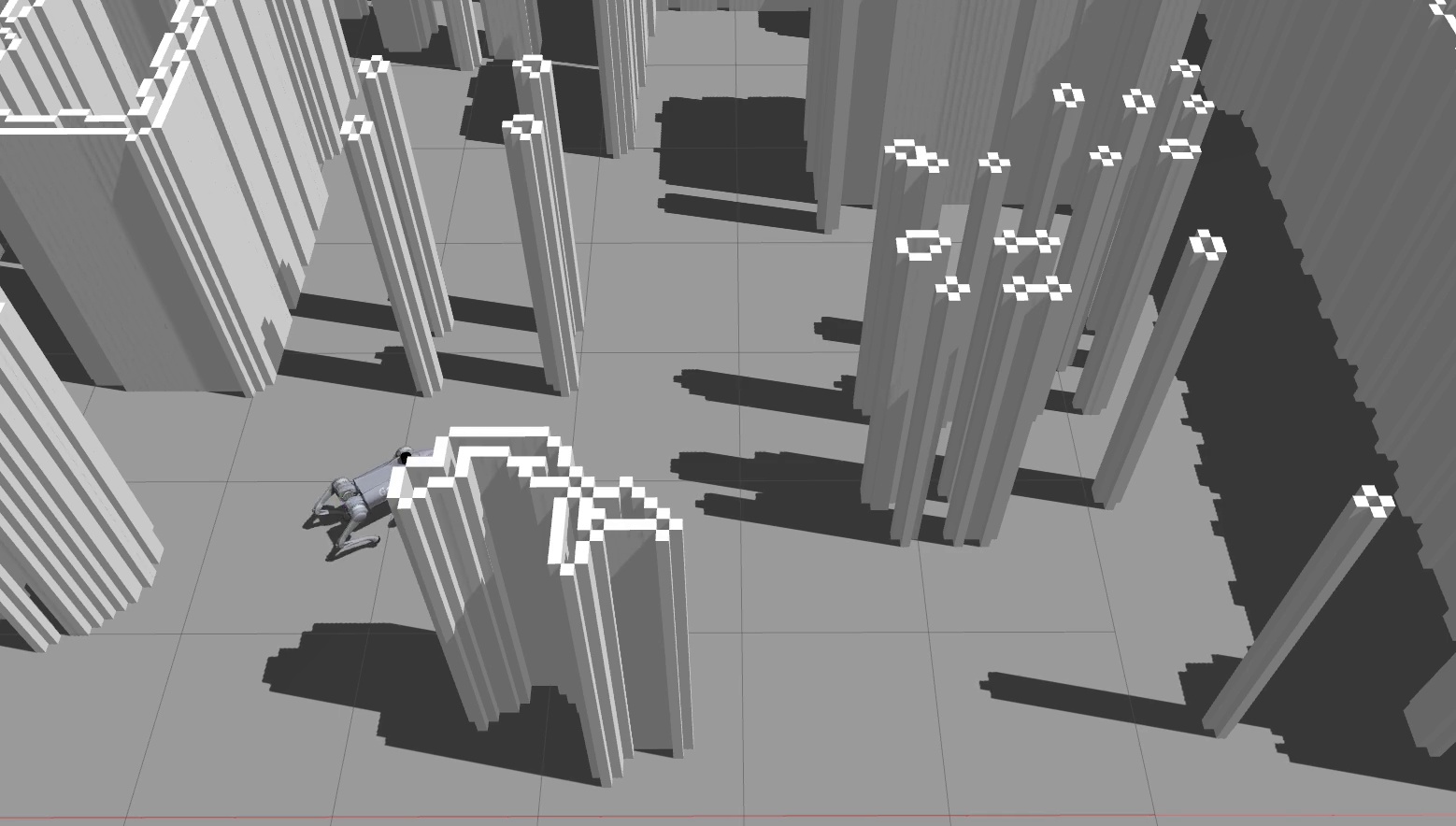}
        \includegraphics[width=0.19\textwidth]{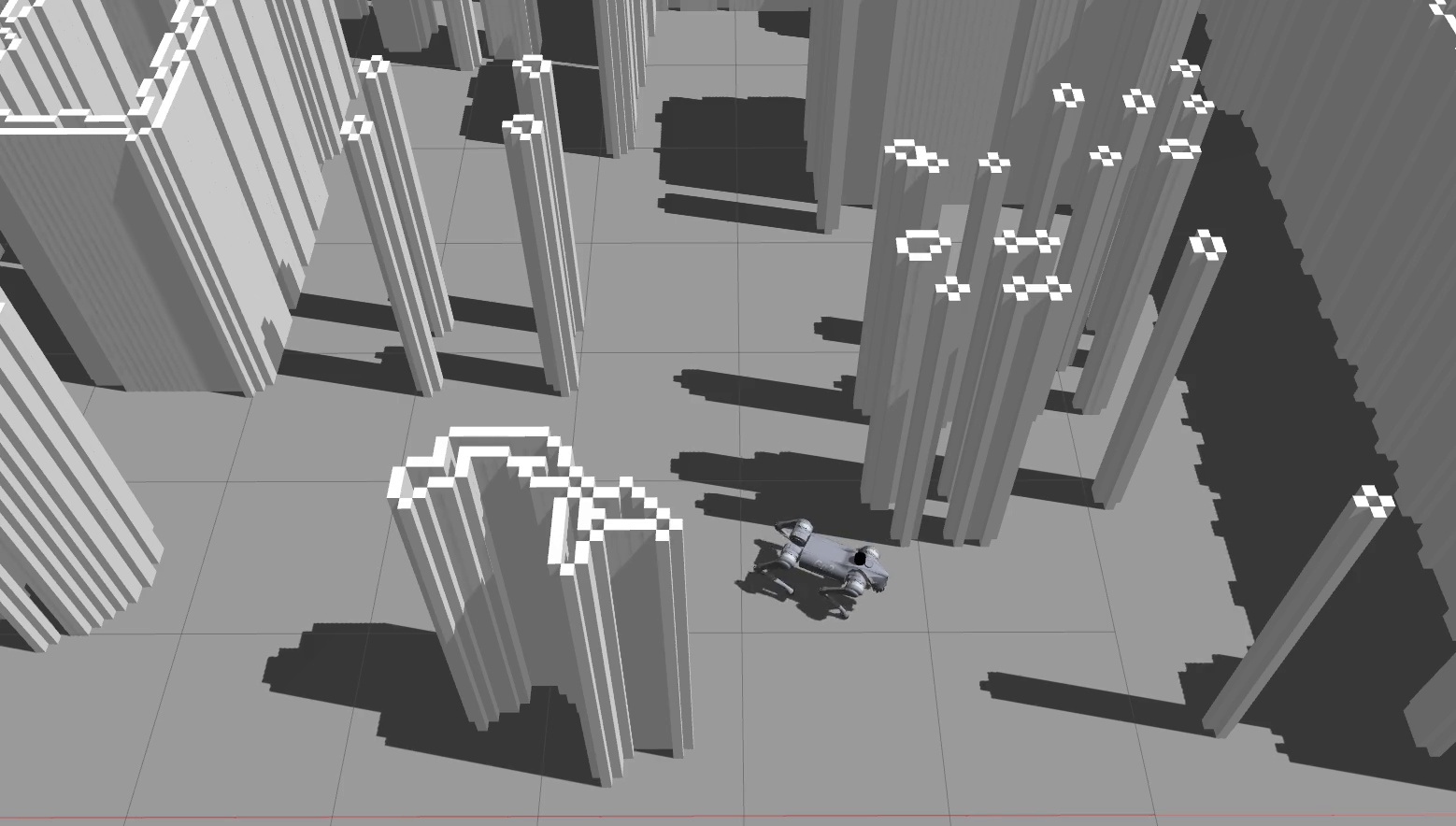}
        \includegraphics[width=0.19\textwidth]{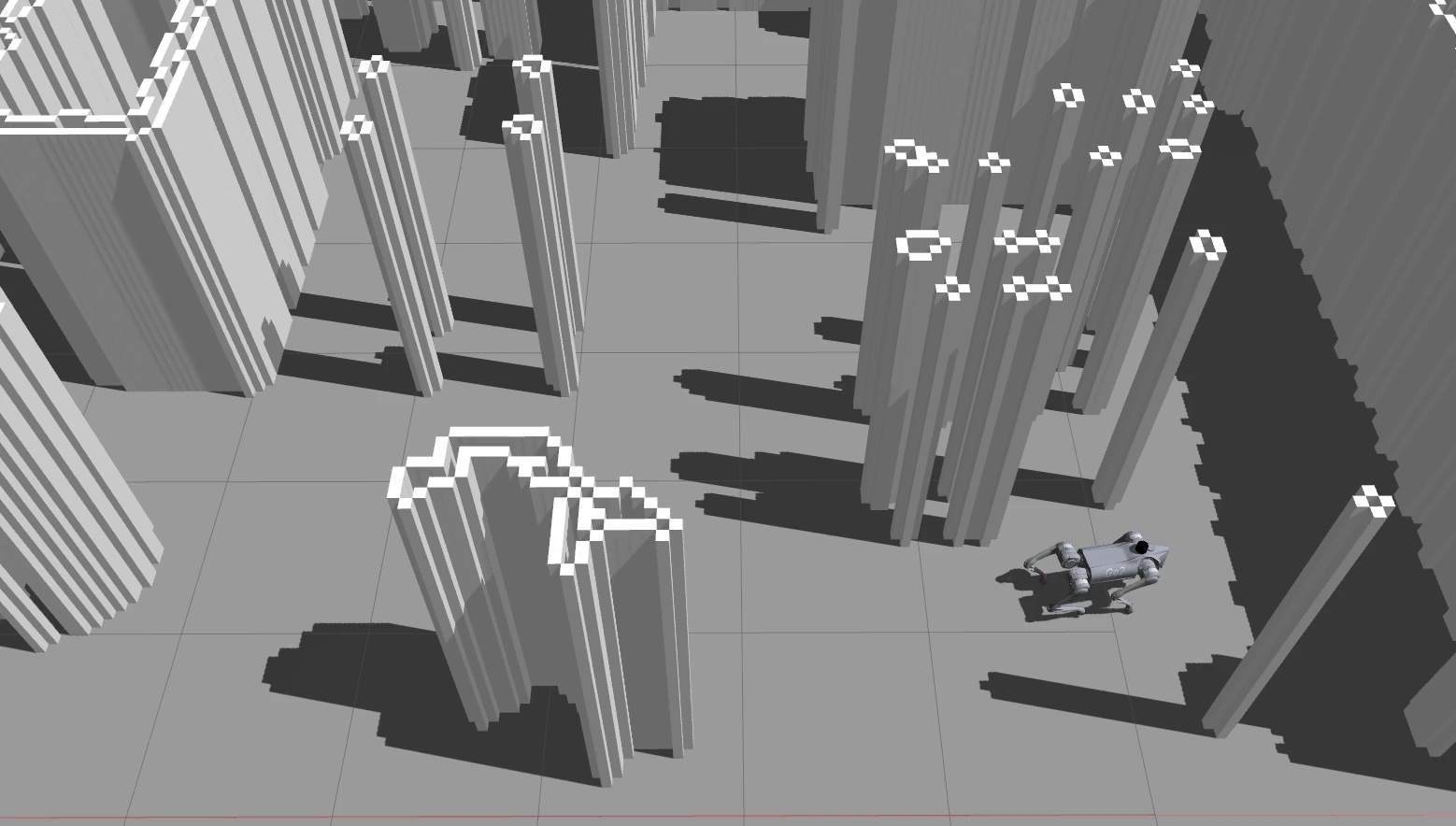}
        \includegraphics[width=0.19\textwidth]{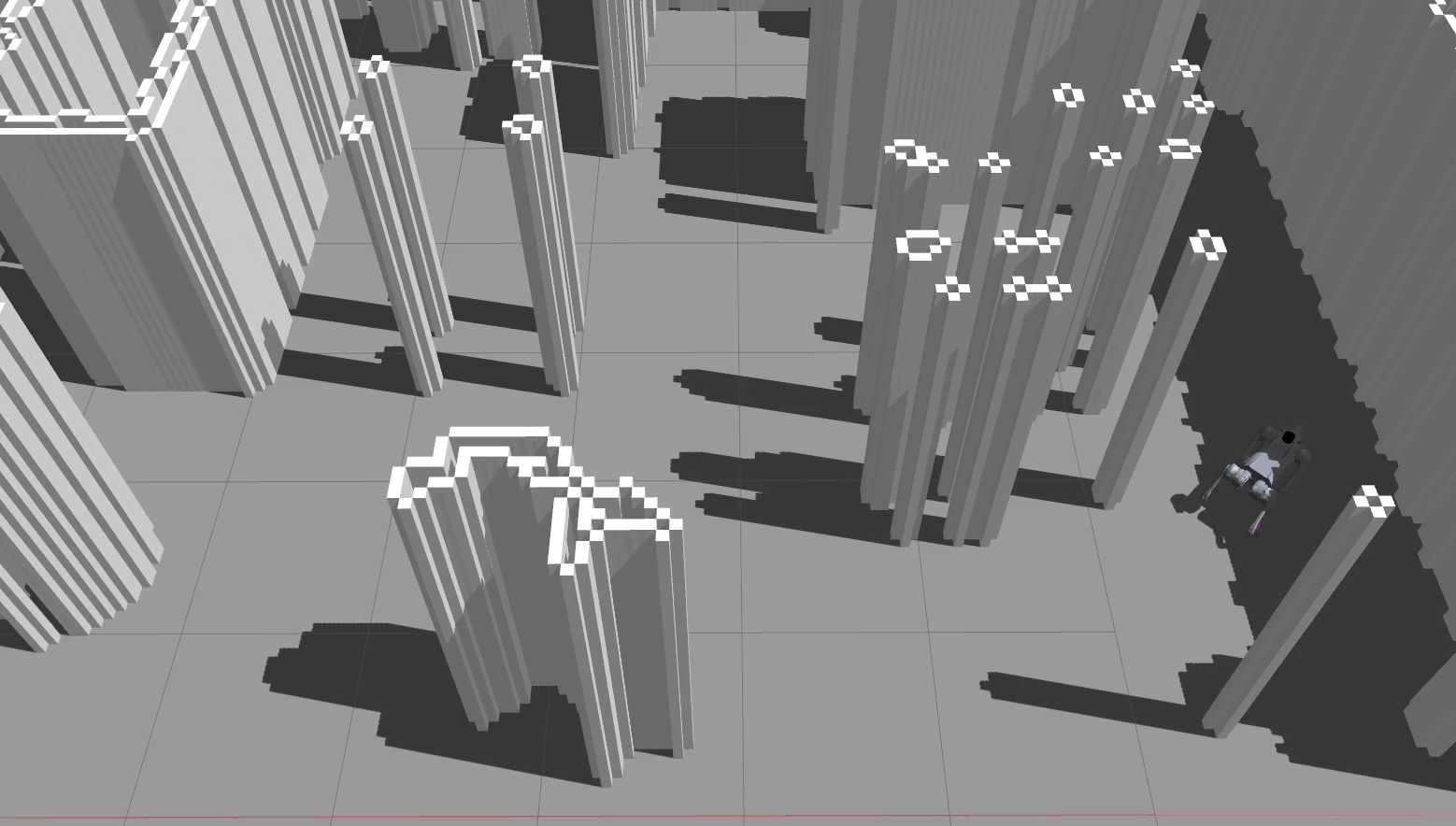}
        \label{fig:result_rls_sim_diffuser}
    }
    \hfill
    \subfigure[Frames of trajectories from~\methodps in simulation.]{
        \includegraphics[width=0.19\textwidth]{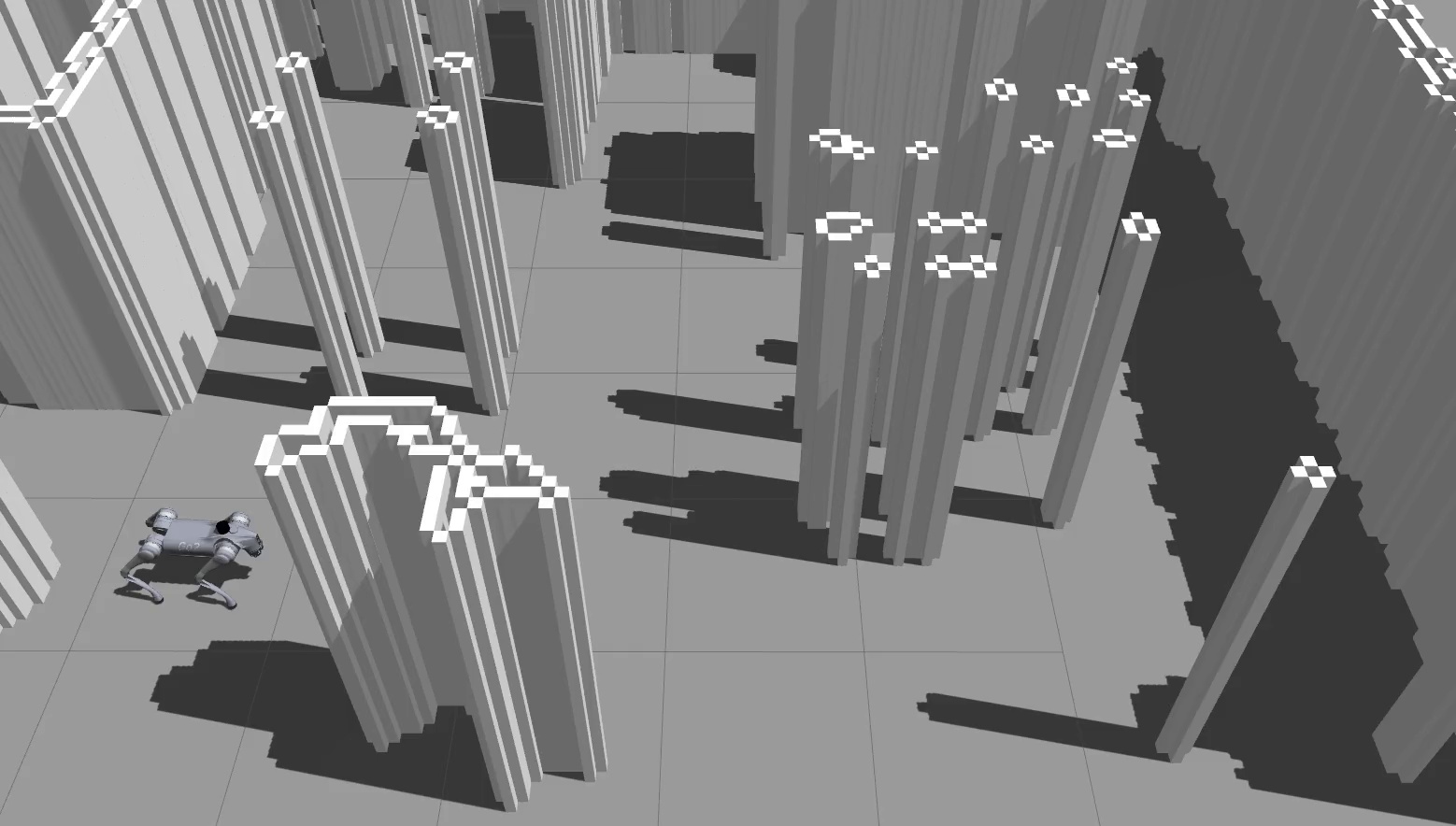}
        \includegraphics[width=0.19\textwidth]{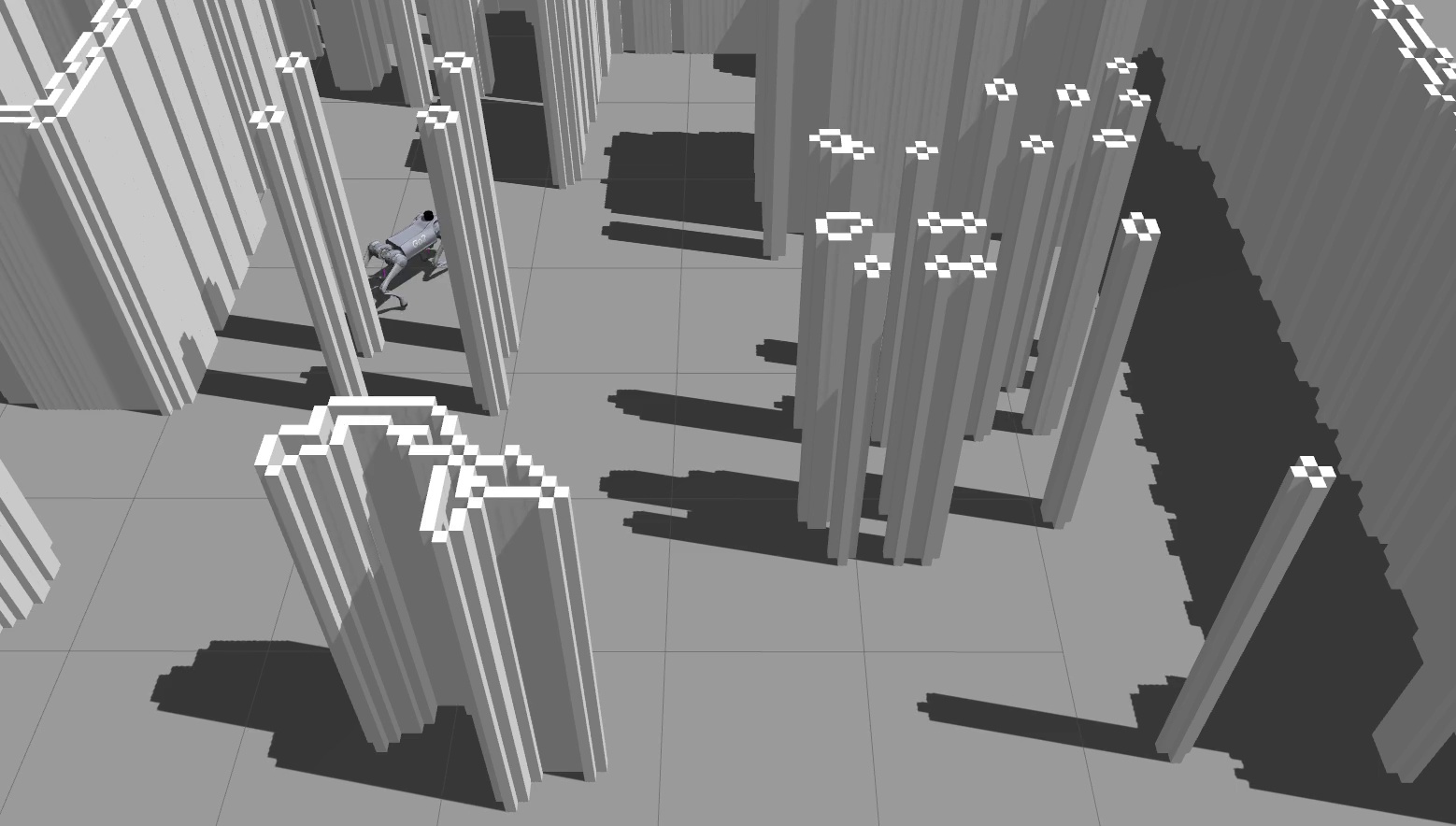}
        \includegraphics[width=0.19\textwidth]{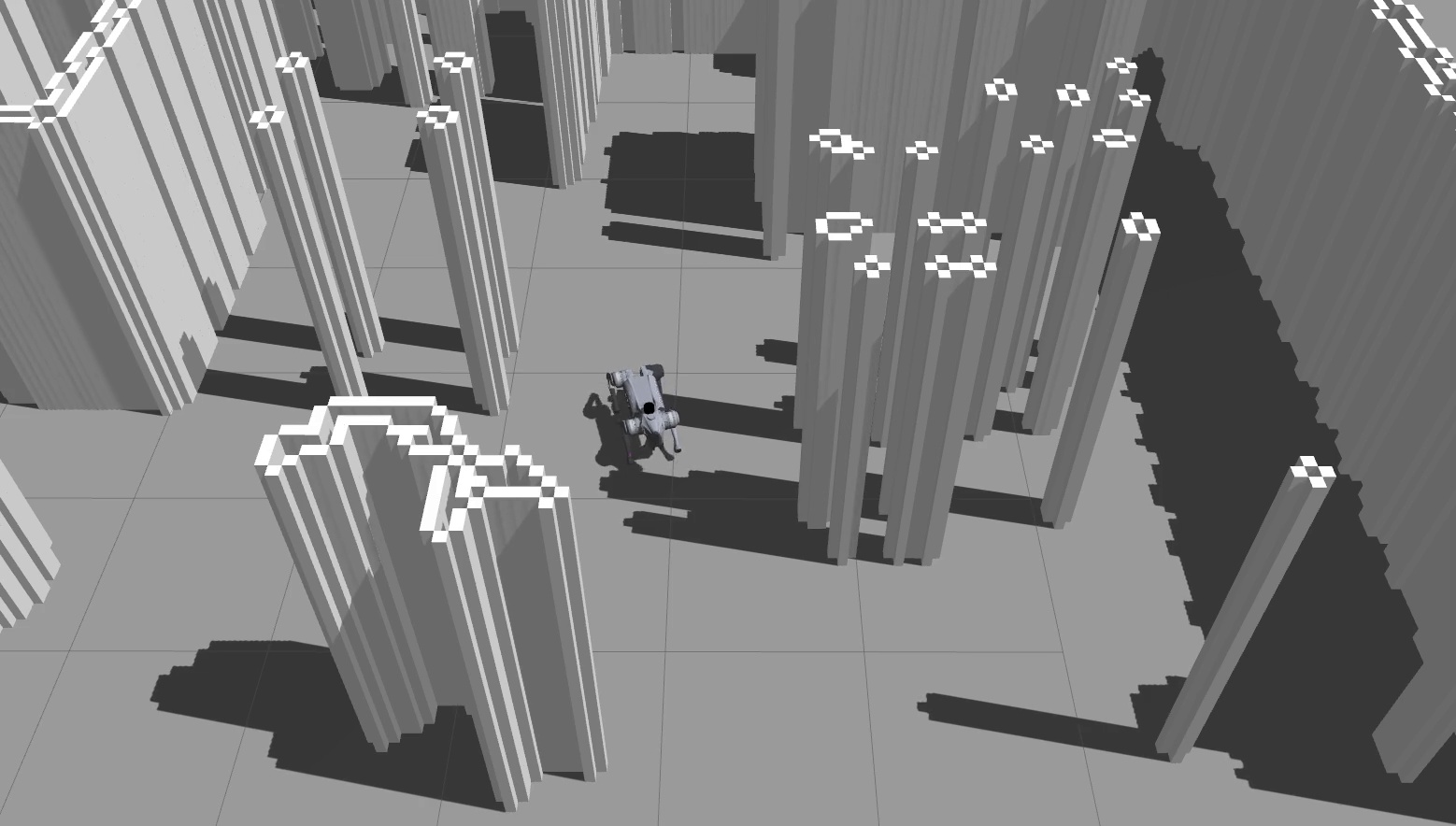}
        \includegraphics[width=0.19\textwidth]{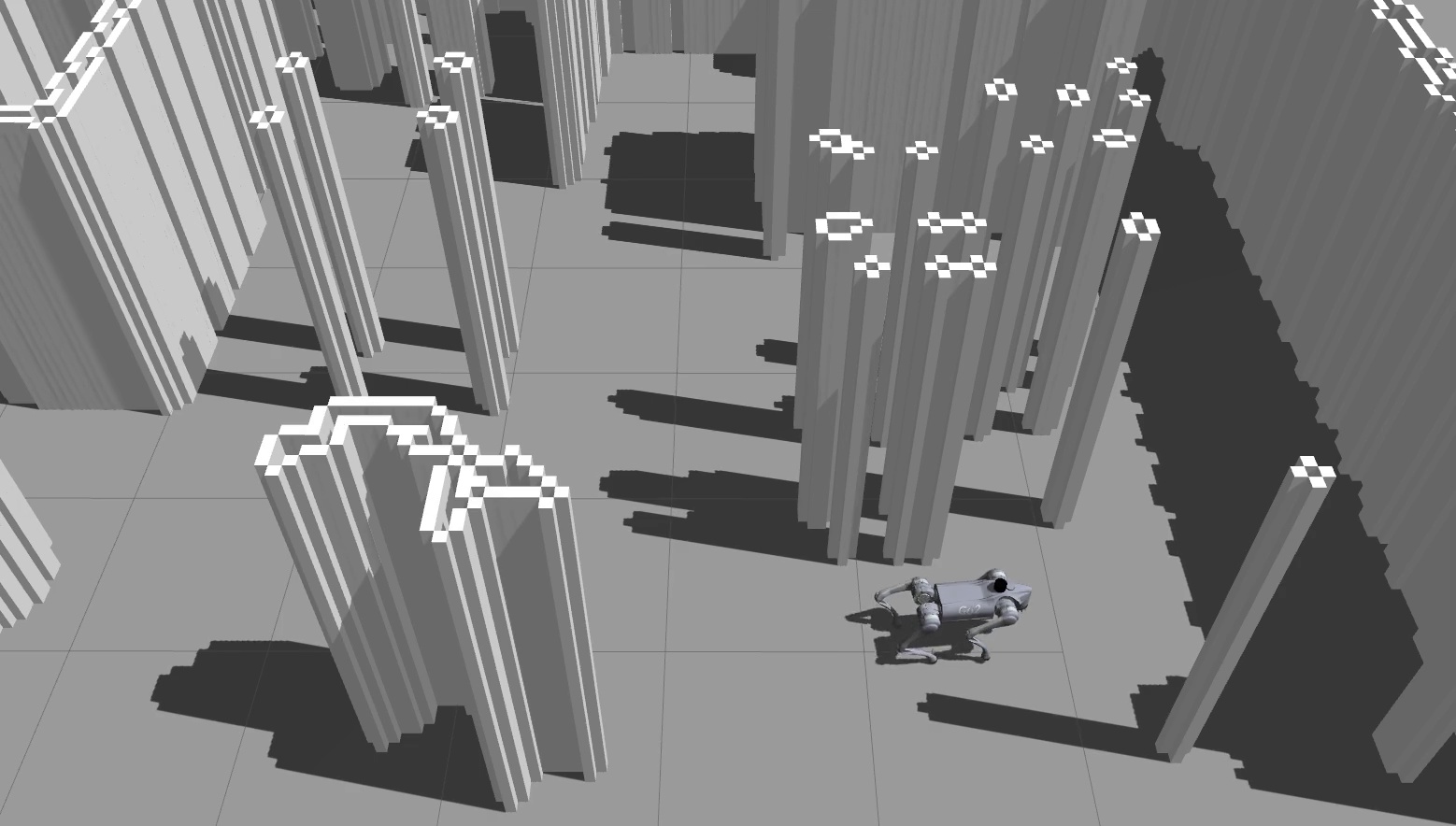}
        \includegraphics[width=0.19\textwidth]{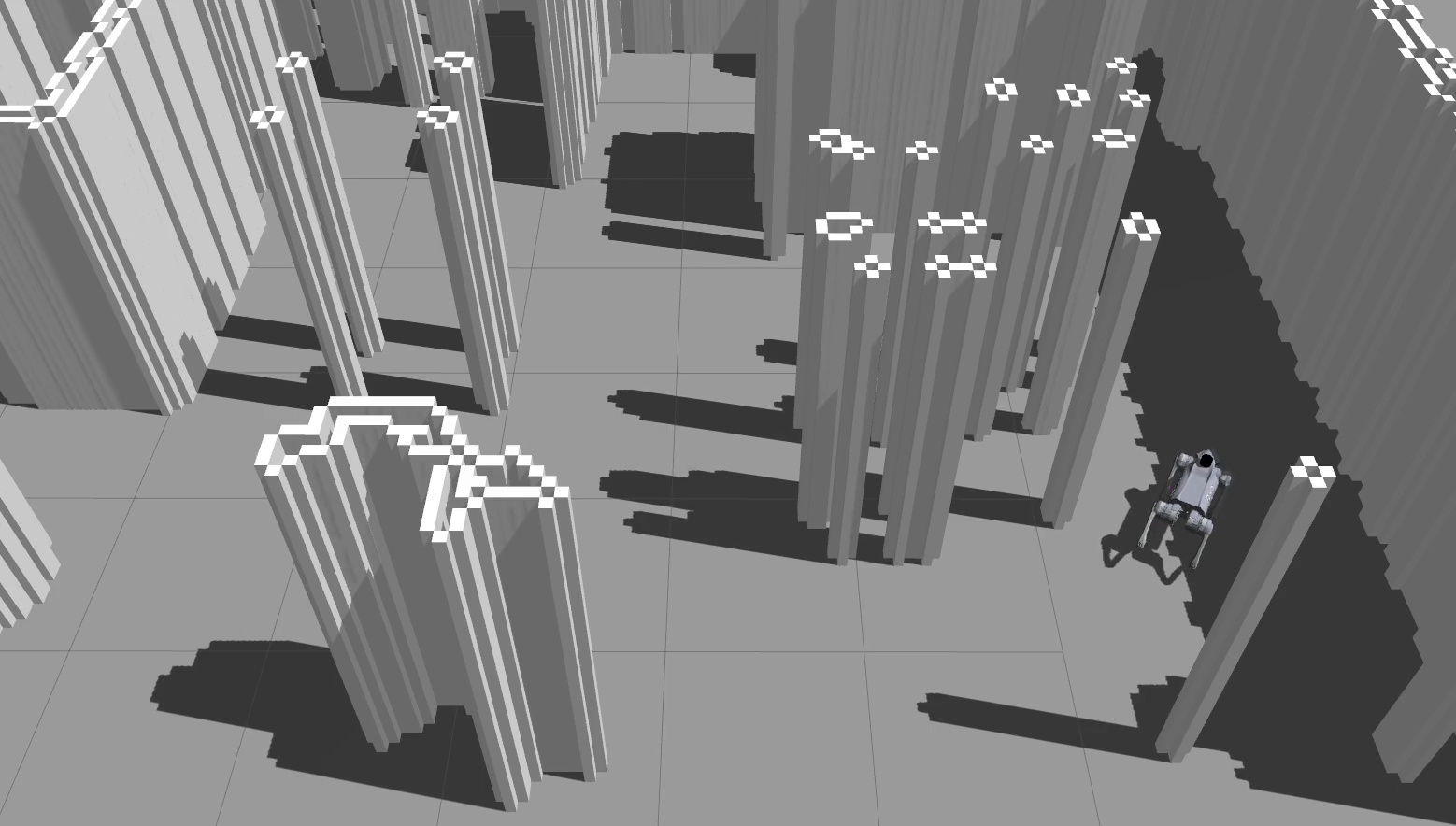}
        \label{fig:result_rls_sim_ps}
    }
    \hfill
    \subfigure[Frames of trajectories from~\methodrg in simulation.]{
        \includegraphics[width=0.19\textwidth]{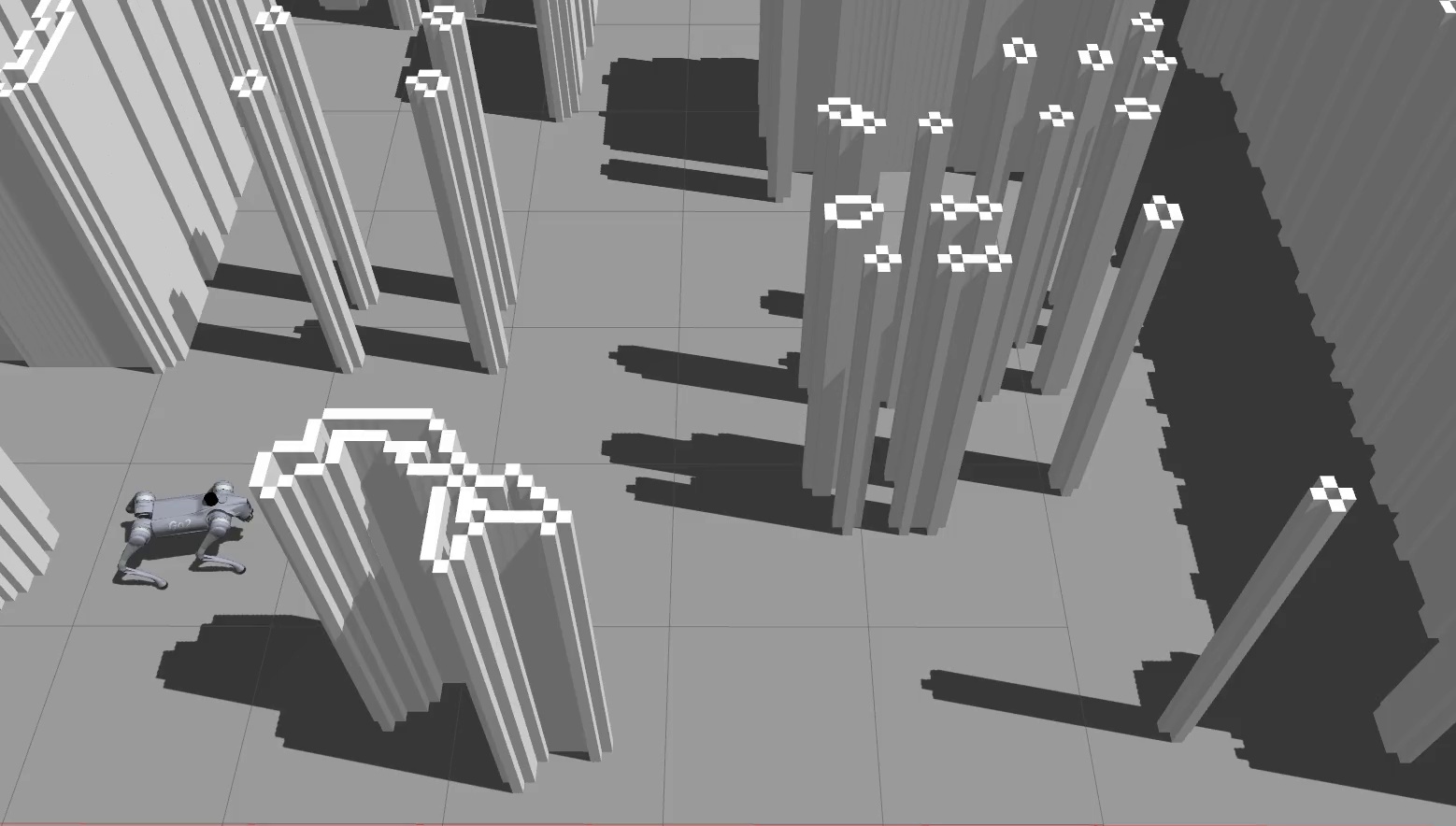}
        \includegraphics[width=0.19\textwidth]{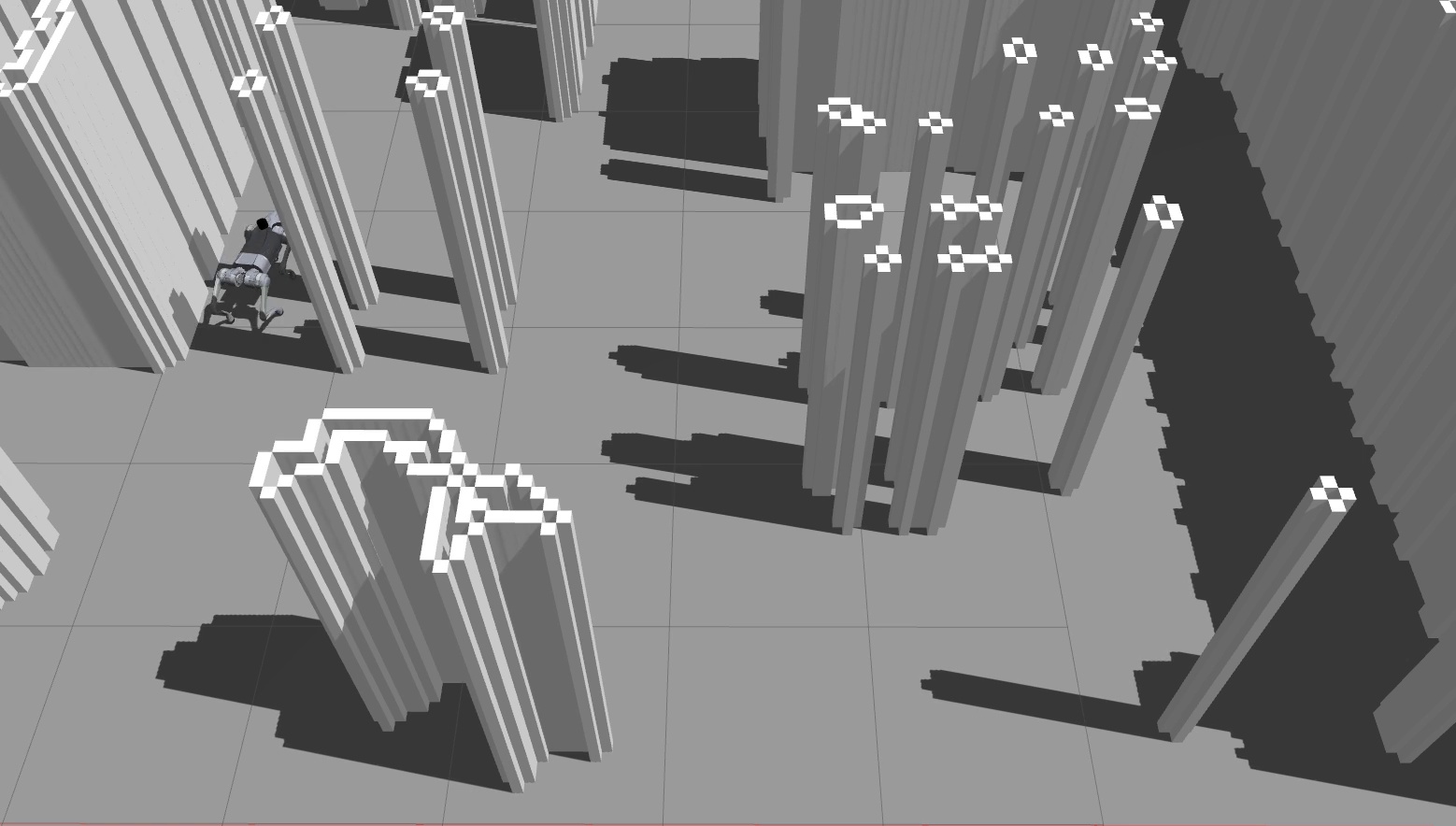}
        \includegraphics[width=0.19\textwidth]{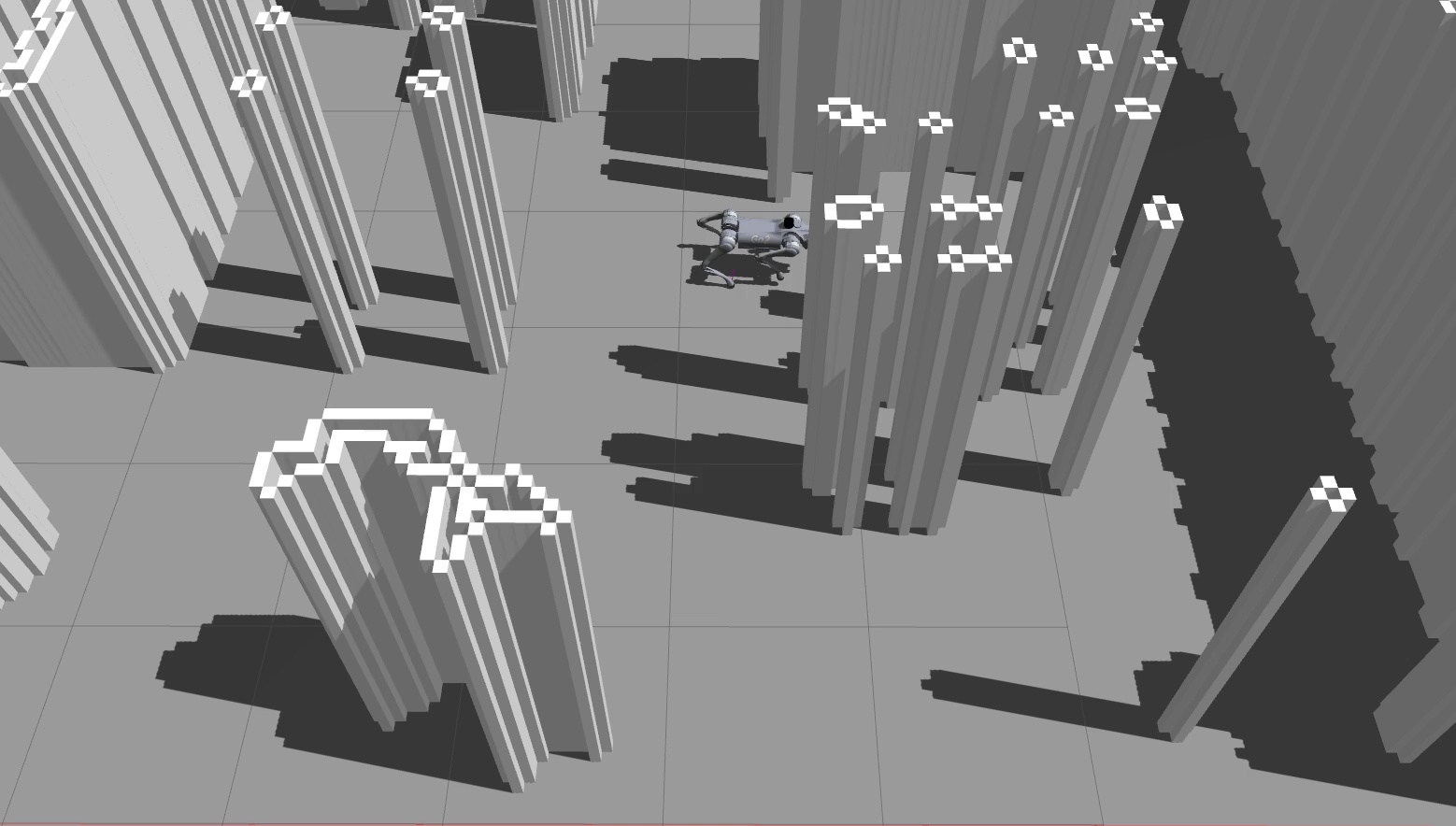}
        \includegraphics[width=0.19\textwidth]{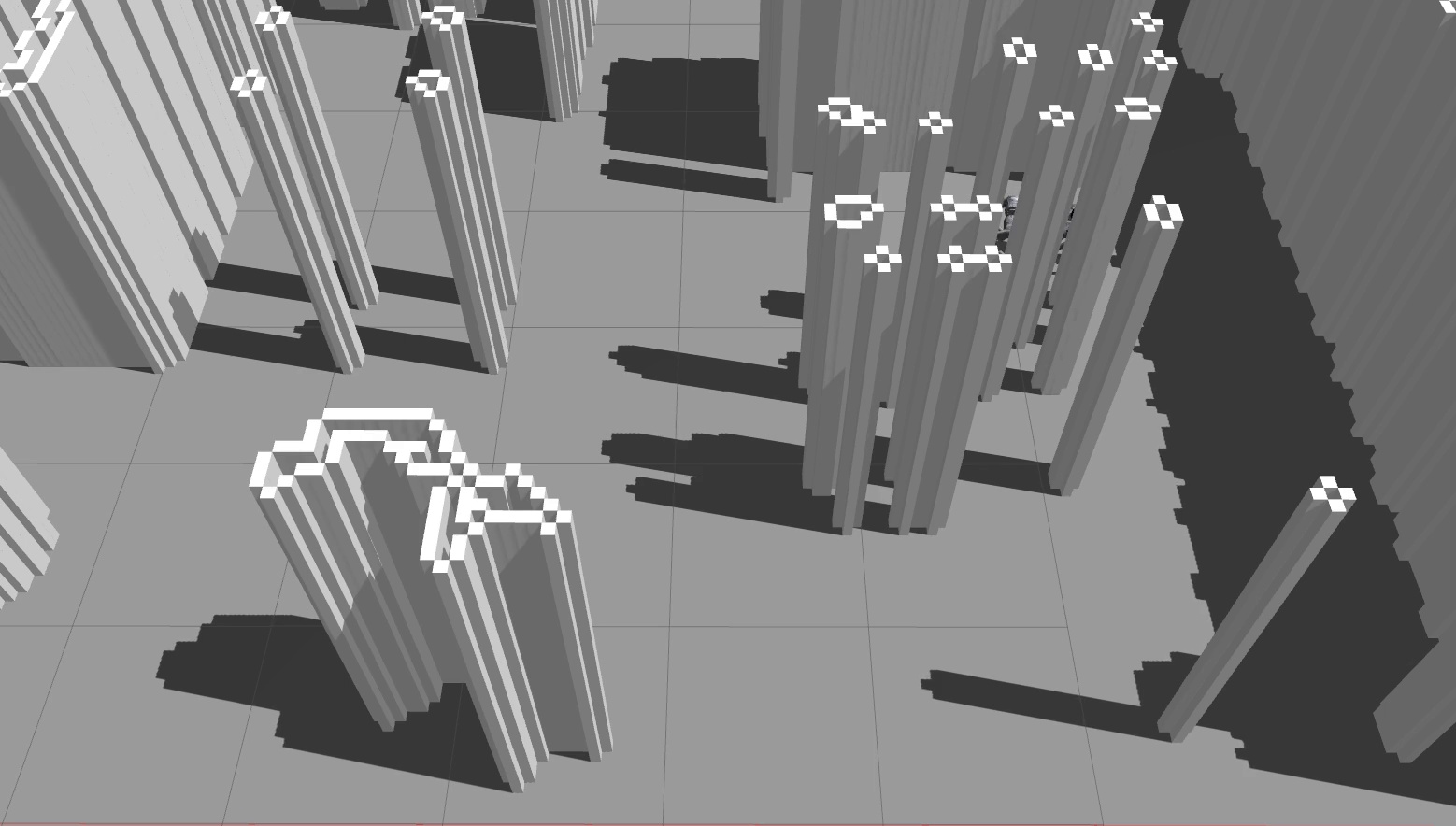}
        \includegraphics[width=0.19\textwidth]{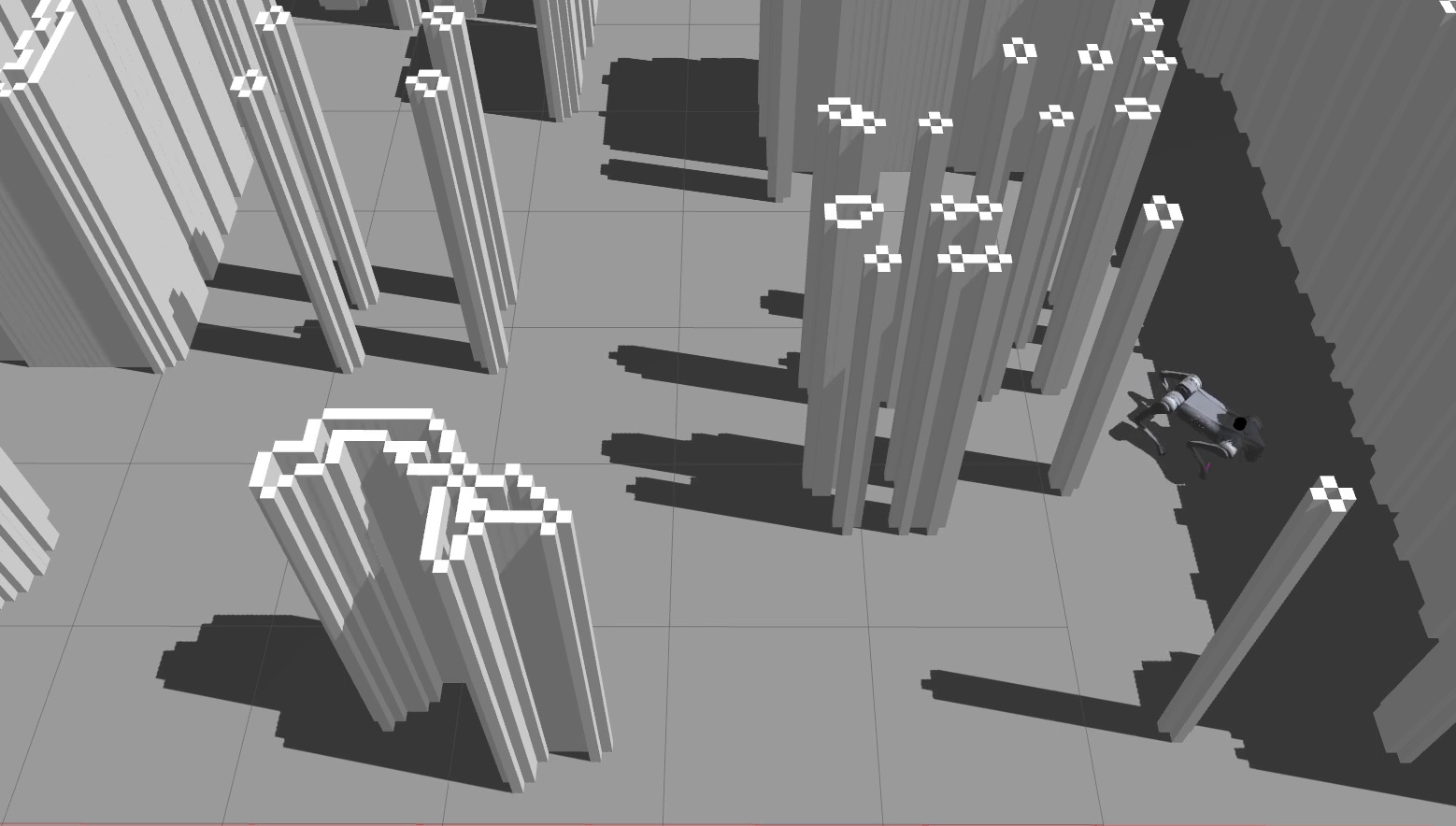}
        \label{fig:result_rls_real_diffuser}
    }
    \caption{Results of safe planning and execution in simulation and indoor room on a quadruped robot.}
    \label{fig:result_rls}
\end{figure*}

We also show in \figref{fig:result_rls} more snapshots of a quadruped robot running in the simulated Lab environment with different policies (baseline against ours). The overall satisfaction rate of all raw generated trajectories in different tasks and environments are summarized in Table~\ref{table:result_rls_ltl}. The overall performance is $85.8\pm14.0\%$ (\emph{c.f.} baseline $2.9\pm5.4\%$) as reported in the main paper.

\subsection{Failure case study}
\label{appsec:failure_case_study}

\figref{fig:result_maze2d_large_unsafe_failure_plan} shows an infeasible trajectory generated from \methodps. Our hypothesis is that posterior sampling in \methodps calculate gradients based on Euclidean distance on the trajectory space due to the LTL evaluation, which may push the trajectory away from the data manifold. On the other hand, the neural network in regressor guidance is trained on noised intermediate samples, which are linked to the data manifold. Gradients ascent on noise space will likely push the noised example to a place where both the satisfaction value and likelihood are high. Visualizations of the difference between two controllers as shown in \figref{fig:result_maze2d_large_unsafe_failure_rollout_c1} \figref{fig:result_maze2d_large_unsafe_failure_rollout_c2} suggest that selecting next locations sequentially without distance checking is a less safe option.

    \parabf{Low-Level Control Policy}
    The success rate during rollout also depends on the specific controller used to visit each location of the generated trajectory. \tabref{table:maze2d_large_unsafe_apdx} shows the comparative results between the two low-level control policies. A more conservative controller (denoted as C2) that visits every location exactly with distance checking can enjoy higher success rate than greedily move (C1, originally used in~\cite{janner2022diffuser}).